\documentclass[10pt,twocolumn,letterpaper]{article}

\usepackage[pagenumbers]{iccv} %

\usepackage{adjustbox}
\usepackage{graphicx}
\usepackage{caption}
\usepackage{subcaption}
\usepackage{printlen}
\usepackage{float}
\usepackage{wrapfig}
\usepackage{multirow}
\usepackage{booktabs}
\usepackage{pifont}
\usepackage{makecell}
\usepackage{nicefrac}
\usepackage{url}
\usepackage{yhmath}
\usepackage{stmaryrd}
\usepackage{verbatim}
\usepackage{amsmath}
\usepackage{algorithm}
\usepackage{algpseudocode}
\usepackage{amsfonts}
\usepackage{microtype}
\usepackage{pgf}
\usepackage{pgfplots}
\usepackage{tabularx}
\usepackage{ifthen}
\usepackage{soul}
\usepackage{tikz}
\usetikzlibrary{angles,quotes,3d,math,arrows.meta,calc,positioning,fit,backgrounds,decorations.pathreplacing,calligraphy,shapes,shapes.multipart}
\usepackage[table]{xcolor}
\usepackage[accsupp]{axessibility}

\newcommand\rurl[1]{%
  \href{https://#1}{\nolinkurl{#1}}%
}

\newcommand{\image}{\mathcal{I}}
\newcommand{\fflow}[1]{\mathbf{f}_{\ifthenelse{\equal{#1}{}}{}{#1}{}}}
\newcommand{\query}{\mathbf{q}}
\newcommand{\queries}{\mathcal{Q}}
\newcommand{\poke}{\mathbf{p}}
\newcommand{\pokes}{\mathcal{P}}

\makeatletter
\newcommand{\xleftrightarrow}[2][]{\ext@arrow 3359\leftrightarrowfill@{#1}{#2}}
\newcommand{\xdashrightarrow}[2][]{\ext@arrow 0359\rightarrowfill@@{#1}{#2}}
\newcommand{\xdashleftarrow}[2][]{\ext@arrow 3095\leftarrowfill@@{#1}{#2}}
\newcommand{\xdashleftrightarrow}[2][]{\ext@arrow 3359\leftrightarrowfill@@{#1}{#2}}
\def\rightarrowfill@@{\arrowfill@@\relax\relbar\rightarrow}
\def\leftarrowfill@@{\arrowfill@@\leftarrow\relbar\relax}
\def\leftrightarrowfill@@{\arrowfill@@\leftarrow\relbar\rightarrow}
\def\arrowfill@@#1#2#3#4{%
  $\m@th\thickmuskip0mu\medmuskip\thickmuskip\thinmuskip\thickmuskip
   \relax#4#1
   \xleaders\hbox{$#4#2$}\hfill
   #3$%
}
\makeatother

\newcommand{\suppvspreprint}[2]{#2}

\definecolor{ourgreen}{RGB}{46, 204, 113}
\definecolor{ourgreenborder}{RGB}{39, 174, 96}
\definecolor{ourblue}{RGB}{52, 152, 219}
\definecolor{ourblueborder}{RGB}{41, 128, 185}
\definecolor{ourorange}{RGB}{230, 126, 34}
\definecolor{ourorangeborder}{RGB}{211, 84, 0}
\definecolor{ourred}{RGB}{231, 76, 60}
\definecolor{ourredborder}{RGB}{192, 57, 43}
\definecolor{ouryellow}{RGB}{241, 196, 15}
\definecolor{ouryellowborder}{RGB}{243, 156, 18}
\definecolor{ourpurple}{RGB}{155, 89, 182}
\definecolor{ourpurpleborder}{RGB}{142, 68, 173}
\definecolor{ourturquoise}{RGB}{26, 188, 156}
\definecolor{ourturquoiseborder}{RGB}{22, 160, 133}
\definecolor{ourturquoise}{RGB}{26, 188, 156}
\definecolor{ourturquoiseborder}{RGB}{22, 160, 133}
\definecolor{ourwhite}{RGB}{236, 240, 241}
\definecolor{ourwhiteborder}{RGB}{189, 195, 199}
\definecolor{ourgray}{RGB}{149, 165, 166}
\definecolor{ourgrayborder}{RGB}{127, 140, 141}

\definecolor{ourwhite2}{RGB}{246, 247, 248}

\definecolor{matplotlibblue}{HTML}{1f77b4}
\definecolor{matplotliborange}{HTML}{ff7f0e}
\definecolor{matplotlibgreen}{HTML}{2ca02c}

\definecolor{ourhighlightcolor}{RGB}{46, 204, 113}

\newcommand{\shorttabular}[1]{\begin{tabular}{c}#1\end{tabular}}

\newcolumntype{H}{>{\setbox0=\hbox\bgroup}c<{\egroup}@{}}

\newcommand{\tikzstylenodedistance}{4mm}
\newcommand{\tikzstyleinnersep}{2mm}
\newcommand{\tikzstyleminimumheight}{8.75mm}
\newcommand{\tikzstyleminimumwidth}{12mm}

\tikzset{
    node distance=\tikzstylenodedistance,
    text centered,
    anchor=center,
}
\tikzset{
    standard node/.style n args={1}{%
        rectangle,
        rounded corners=0.1cm,
        fill=our#1,
        draw=our#1border,
        line width=0.04cm,
        minimum height=\tikzstyleminimumheight,
        minimum width=\tikzstyleminimumwidth,
        inner sep=\tikzstyleinnersep,
        text centered,
        anchor=center,
        align=center,
    }
}
\tikzset{
    standard node module/.style n args={0}{%
        rectangle,
        rounded corners=0.1cm,
        fill=ourturquoise,
        draw=ourturquoiseborder,
        line width=0.04cm,
        minimum height=\tikzstyleminimumheight, %
        minimum width=12mm, %
        inner xsep=\tikzstyleinnersep,
        inner ysep=1mm,
        text centered,
        anchor=center,
        align=center,
    }
}
\tikzset{
    standard node image/.style n args={1}{%
        rectangle,
        fill=our#1,
        draw=our#1border,
        line width=0.04cm,
        minimum height=\tikzstyleminimumheight,
        minimum width=\tikzstyleminimumwidth,
        inner sep=0,
        text centered,
        anchor=center,
        align=center,
    }
}
\tikzset{
    standard node circle/.style n args={1}{%
        fill=our#1,
        draw=our#1border,
        circle,
        inner sep=0.1cm,
        minimum height=0,
        minimum width=0,
    }
}
\tikzset{
    standard node circle/.prefix style = standard node
}

\tikzset{
    standard line/.style n args={0}{%
        line width=0.04cm,
        rounded corners=0.1cm,
    }
}
\tikzset{
    standard arrow/.style n args={0}{%
        -latex,
    }
}
\tikzset{
    standard arrow/.prefix style = standard line
}

\tikzset{
    simple node image/.style n args={0}{%
        rectangle,
        inner sep=0,
        text centered,
        anchor=center,
        align=center,
        node distance=0mm
    }
}

\definecolor{iccvblue}{rgb}{0.21,0.49,0.74}
\usepackage[pagebackref,breaklinks,colorlinks,allcolors=iccvblue]{hyperref}

\def\paperID{6557} %
\def\confName{ICCV}
\def\confYear{2025}

\title{\textit{What If}: Understanding Motion Through Sparse Interactions}

\author{
    Stefan Andreas Baumann\thanks{\small Equal Contribution.} \qquad Nick Stracke\footnotemark[1] \qquad Timy Phan\footnotemark[1] \qquad Bj\"orn Ommer\\[0.5em]
    CompVis @ LMU Munich \\
    Munich Center for Machine Learning (MCML)
}

\begin{document}
\maketitle
\begin{abstract}
Understanding the dynamics of a physical scene involves reasoning about the diverse ways it can potentially change, especially as a result of local interactions.
We present the Flow Poke Transformer (FPT), a novel framework for directly predicting the distribution of local motion, conditioned on sparse interactions termed ``pokes''. Unlike traditional methods that typically only enable dense sampling of a single realization of scene dynamics, FPT provides an interpretable directly accessible representation of multi-modal scene motion, its dependency on physical interactions and the inherent uncertainties of scene dynamics. 

We also evaluate our model on several downstream tasks to enable comparisons with prior methods and highlight the flexibility of our approach. On dense face motion generation, our generic pre-trained model surpasses specialized baselines. FPT can be fine-tuned in strongly out-of-distribution tasks such as synthetic datasets to enable significant improvements over in-domain methods in articulated object motion estimation. Additionally, predicting explicit motion distributions directly enables our method to achieve competitive performance on tasks like moving part segmentation from pokes which further demonstrates the versatility of our FPT.

\noindent Code and models are publicly available at\\\href{https://compvis.github.io/flow-poke-transformer}{compvis.github.io/flow-poke-transformer}.

\end{abstract}
    
\section{Introduction}
\label{sec:intro}

\begin{figure}[H]
    \centering
    \quad\includegraphics[width=.9\linewidth]{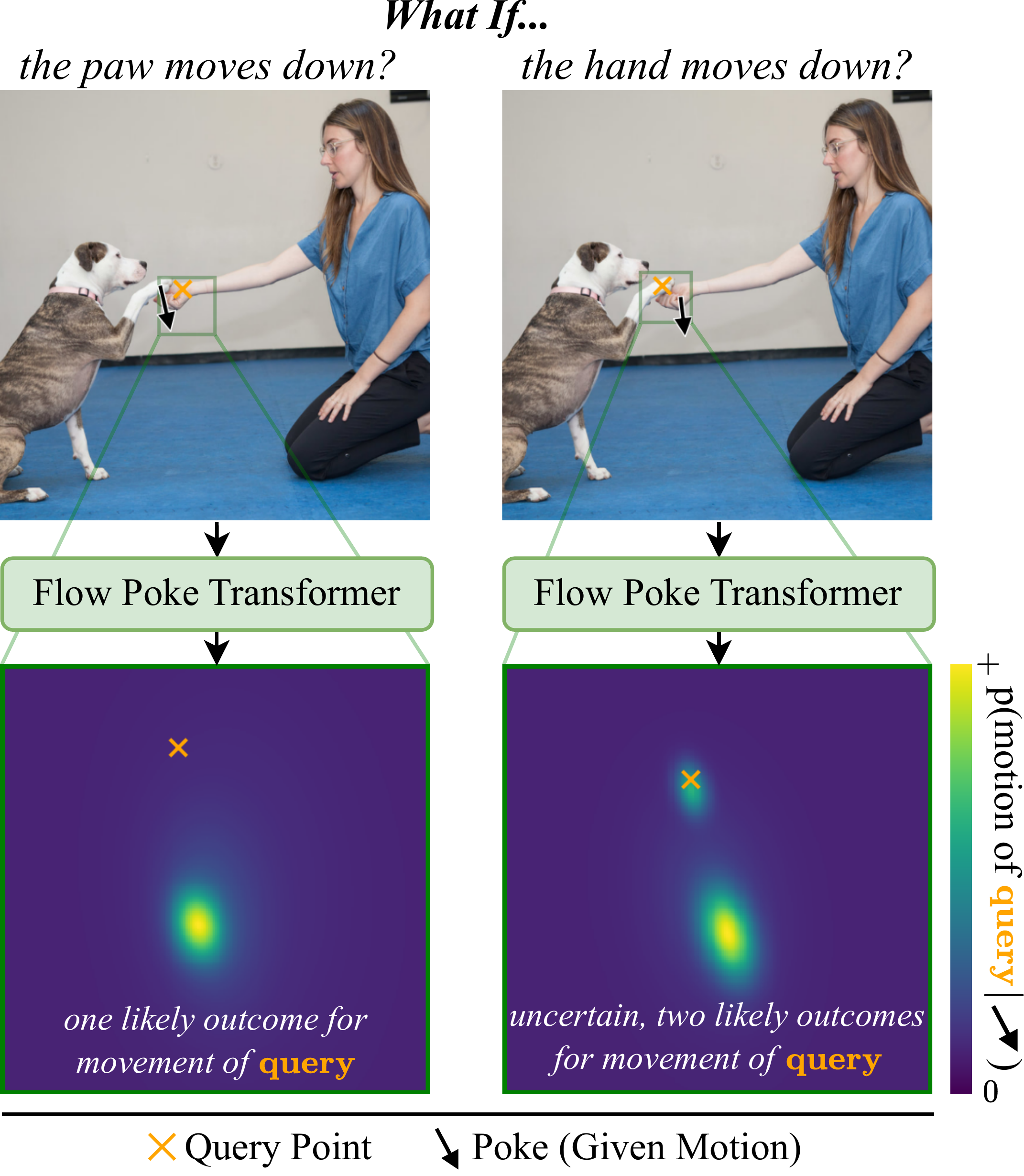}
    \vspace{-2mm}
    \caption{\textit{What If:} Our Flow Poke Transformer directly models the {uncertainty of the world} by predicting \textit{distributions} of how objects ($\color{matplotliborange}\boldsymbol{\times}$) may move conditioned on some input movements (pokes, $\rightarrow$). We see that whether the hand (below paw) or the paw (above hand) moves downwards directly influences the other's movement. Left: the paw pushing the hand down, will force the hand downwards, resulting in a unimodal distribution. Right: the hand moving down results in two modes, the paw following along or staying put. }
    \label{fig:teaser}
\end{figure}

A key feat of human visual intelligence is motion understanding, our ability to understand and predict the various ways the world around us \emph{could potentially} change at a given point in time (see \cref{fig:teaser}). Our cortex is not creating a mental video, focusing on how the colors of individual pixels change. Rather, we are constantly making predictions about the various ways individual objects or parts thereof could potentially move and deform \cite{Palmer_99}. We do not perceive the future as an unambiguously deterministic sequence of events but as a vast space of possibilities. 

It is natural to focus selectively on parts of a scene, infer how they might evolve, and reason about the underlying physical properties and interactions that drive change.
This selective, probabilistic, and multimodal reasoning is rooted in the perceived inherent stochastic nature of the world. It is governed by the stochastic physical properties of complex systems and
further compounded by the presence of agents with a complex, inaccessible internal state, lead by free will or other, from the outside often unapproachable causes.

This inherent uncertainty makes dense, deterministic predictions of future motion both impractical and ill-posed to represent real-world dynamics. The prediction of pixel-perfect and long-term sequences \cite{SoraOpenAI} requires models to commit to one trajectory and, in so doing, ignore the rich multimodality of real-world outcomes. At best, such a prediction biases the scenario towards a single plausible future; at worst, it produces photorealistic frames that show limited understanding of physical processes, interactions, and constraints. In many situations, like autonomous agent systems, robotics, and automated planning, the ability to predict and process multiple possible outcomes of a given situation is more valuable than the naive assumption that events play out according to a single trajectory. 

To address these issues, we propose a framework for representing the distribution of possible motions of parts of a scene. To control the degree of uncertainty, a human observer can interact with or perturb a scene with local ``\textit{pokes}'', by nudging an object or applying a force. {By repeating similar interactions, the multimodal nature of potential outcomes can be observed.} Similarly, we allow conditioning the motion distribution on such sparse, localized pokes.
Compared to traditional dense approaches, our method operates at a higher level of abstraction, predicting 
localized \textit{distributions} of motion rather than committing to a single outcome. This approach aligns more closely with real-world dynamics where uncertainty and multimodality are intrinsic and actionable insights often emerge from reasoning about sparse, local changes rather than exhaustive dense predictions. These include aspects like the inherent interpretability of explicitly predicted distributions, such as identifying modes and quantifying uncertainty directly.

By avoiding dense (video) prediction, our model re-frames motion prediction as a problem of capturing potential dynamics, directly predicting motion distributions. For instance, a poke applied to an unstable stack of blocks might cause it to topple in multiple ways, remain stable, or shift slightly without collapsing. We capture this variability, avoiding the pitfalls of dense video models that have to commit to one specific sample in the set of potential outcomes. This also addresses the impracticality of dense or long-term predictions, where compounding uncertainty renders dense outputs increasingly arbitrary.

Applications of our proposed model include (sparse) interactive simulation, where pokes guide scene exploration and the multimodal distributions of possible motions are directly captured, moving part segmentation,
but also classic sampling of dense motion predictions. {As opposed to optical flow estimation and tracking, where the future is given via a future frame, from which motion is estimated, we predict what future motion might be from only a single frame.}

Overall, we present a step toward a more efficient, flexible, and detailed understanding of scene dynamics. We focus on the vast distribution of what could happen---optionally conditioned on sparse interactions instead of rendering specific futures. Our framework is not only efficient and scalable, but also conceptually aligned with the inherent uncertainties our human perception and reasoning is facing when dealing with our changing environment.

\noindent Our main contributions are as follows:
\begin{itemize}
    \item Multimodal Distribution Prediction: we directly predict full distributions of potential motion instead of just enabling sampling from them, providing increased flexibility in applications over previous approaches, such as directly estimating uncertainties.
    \item Sparse {Kinematics} Modeling: our method reasons about sparse, local motion distributions across the scene. This balances efficiency with expressive power by focusing computational resources where they matter most.
    \item Generalizability: our method can learn a generic motion understanding from {unstructured} web videos, generalizing effectively to diverse, open-world data.
    \item Efficiency: our approach of sparsely modeling interactions enables sparse predictions with our method in 25ms and throughputs of more than 160k parallel predictions per second on a single modern GPU which is promising for real-time applications.
\end{itemize}

\section{Related Work}
Estimation of plausible motion for a given image or scene has been approached in various ways over the years. What makes this task particularly challenging is that it requires the model to have a physical understanding of how objects move in general, how they can be manipulated, and how they relate to each other. Many approaches directly predict a video from still images which makes it harder to access and leverage the underlying motion understanding of the model. Other approaches first predict a dense flow map which they use to later warp the images. However, more complex scenes can have multiple instantiations of realistic motion depending on the given conditioning, which we aim to model directly. In the following, we review various methods which have been studied in the literature.

\paragraph{Motion-based Editing}
A field that has recently gained attention is image editing using diffusion models by providing a set of pokes that indicate how specific parts of the image should move. \cite{pan2023draggan} takes a GAN \cite{goodfellow2014gan} generated image and warps it using motion supervision based on user-provided pokes. ~InstantDrag \cite{shin2024instantdrag} on the other hand first predicts dense optical flow using a GAN and uses that as conditioning for a diffusion model to generate the final warped image.

Similar approaches have been used in video editing \cite{wang2024motionctrl, li2024imageconductor, dai2023motionguidance} that extend base models with ControlNets~\cite{zhang2023control} or LoRAs~\cite{hu2021lora} to condition the model on the desired motion. The goal is then to move entire objects according to a specific poke by selecting the object with a bounding box or entity representation \cite{wu2025draganything}.
Unlike our method, this general direction neglects an understanding of physical and realistic motion in exchange for precise adherence to the poke guidance and realistic inpainting of occluded regions.

\paragraph{Motion Generation}

Various works learn to hallucinate motion for static images \cite{walker2015dense, rosello2016predicting, gao2018im2flow,shi2024motion,liang2024movideo,blattmann2021ipoke}. MoVideo~\cite{liang2024movideo} and Motion-I2V~\cite{shi2024motion} use diffusion models to predict dense flow sequences given a start frame and use them to synthesize videos. Motion-I2V specifically allows conditioning on sparse movement information using ``motion drags'', similar to pokes, which is an improvement upon DragNUWA~\cite{yin2023dragnuwa} that directly synthesizes dense RGB video from drags. The latter makes the actual motion prediction substantially less accessible because it needs to be estimated with an additional model like RAFT~\cite{teed2020raft} or COTR~\cite{jiang2021cotr}, a property also shared by other methods~\cite{li2024puppet,li2024dragapart}.
\cite{walker2015dense} predicts discrete bins of optical flow for static images with a classification loss. This enables them to model multiple flow fields for a single image. Im2Flow \cite{gao2018im2flow} predicts a single realization of continuous optical flow for an image and combines that with the image to boost action classification performance.

Learning how objects move and behave together can also be used as a general pretext task to build physical scene understanding. \cite{blattmann2021ipoke} introduced the concept of pokes as sparse motion conditioning to indicate how the poked object should move and directly synthesize dense RGB videos on limited-domain datasets. This is an unspecified problem as movement information is only available for a small number of poked pixels and the model needs to learn how the remainder of the scene moves. DragAPart~\cite{li2024dragapart} and the follow-up work PuppetMaster~\cite{li2024puppet} focus on modeling the movement of individual parts of objects for a closed-domain, synthetic dataset (part-level motion). While these works focus on building a more fine-grained physical understanding, they directly predict the result of the poke(s) in RGB space. This makes the underlying motion representation harder to access and requires e.g. optical flow estimation between frames. Additionally, they do not provide any uncertainty estimation in the form of an underlying motion distribution, but simply render a single possible sample of the result space.

Other approaches focus on directly predicting a specific physical representation of motion. Generative Image Dynamics~\cite{li2024generative} learns oscillatory dynamics as commonly found in nature using Fourier-based motion representations. PhysDreamer~\cite{zhang2025physdreamer} and PhysGaussian~\cite{xie2024physgaussian} extend the work of \cite{li2024generative} from 2D to 3D scenes. While these approaches work well in their limited domains, they lack the flexibility to model the vast, often non-oscillatory motion space of the real world and are thus limited in the amount of general motion understanding they can obtain though training.

\paragraph{Generative Models in Computer Vision}
Generative models have recently become a cornerstone in computer vision, as they allow modeling tasks through full conditional distributions $p(\mathbf{y}|\mathbf{x})$ instead of reducing predictions to single-point estimates, such as the expectation $\mathbb{E}[\mathbf{y}|\mathbf{x}]$ often used in discriminative models. Major paradigms include GANs~\cite{goodfellow2014gan}, diffusion models~\cite{ho2020denoising,song2021scorebased}, and autoregressive (AR) models~\cite{vaswani2017attention}. GANs and diffusion models enable sampling from the modeled distribution but provide limited direct insight into its structure. Diffusion models, in particular, have demonstrated scalability to general data distributions and billions of parameters~\cite{podell2024sdxl}, whereas GANs are typically constrained to closed-set distributions. AR models, extensively applied in NLP~\cite{vaswani2017attention} and increasingly adopted for vision tasks~\cite{esser2020taming,yu2022parti,cao2021image}, directly model probability mass functions (PMFs) for discrete distributions and scale well to hundreds of billions of parameters~\cite{dubey2024llama3}. However, the discrete nature of PMFs limits their applicability to real-valued problems, which are prevalent in vision tasks. Recent advances such as GIVT~\cite{tschannen2024givt} and \cite{li2024autoregressive} have extended AR transformers to continuous-valued outputs while retaining the scalability of AR transformer models~\cite{fan2024fluid}. Specifically, \cite{li2024autoregressive} employ a diffusion model to sample from autoregressively predicted feature vectors, while GIVT directly parameterizes distributions as Gaussian Mixture Models (GMMs) with diagonal covariances, enabling direct access to the probability density function (PDF) for downstream applications.
Our implementation builds upon the latter, while extending it to non-diagonal covariances to enable accurate modeling of motion distributions.

\section{Method}

\begin{figure*}[ht]
    \centering
    \includegraphics[width=.9\linewidth]{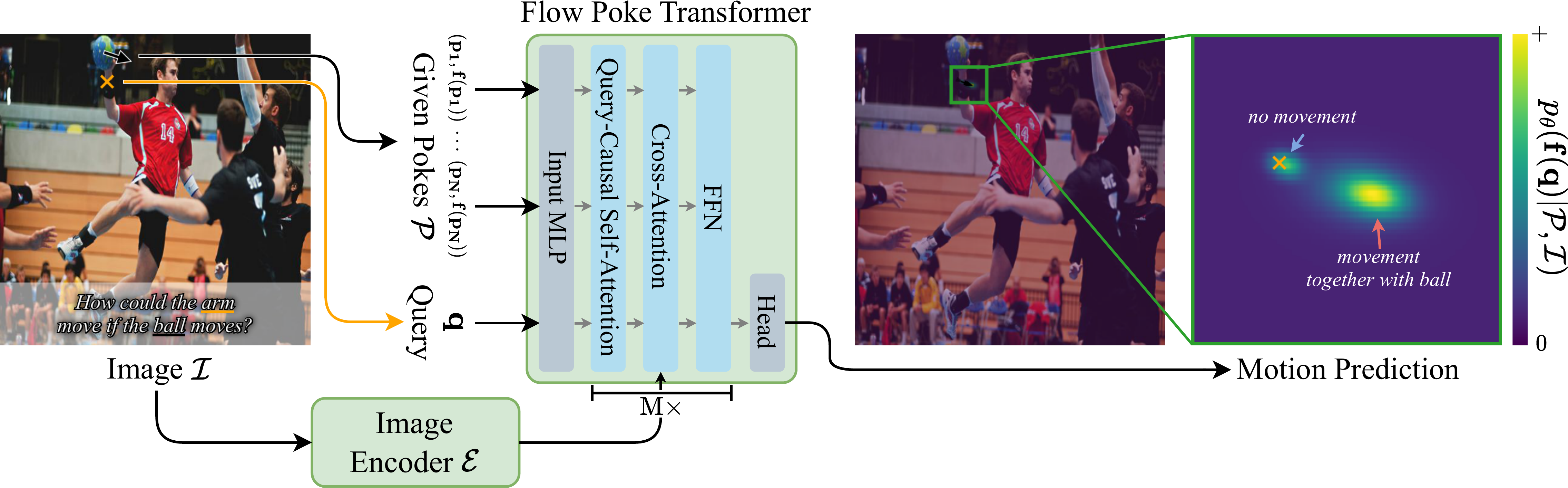}
    \vspace{-2mm}
    \caption{\textbf{High-level Model Architecture Overview}. Given an image $\image$, a set of given pokes $\pokes$ (visualized as arrows $\rightarrow$), and query positions $\query$ ($\!\color{matplotliborange}\times\!$), our model directly predicts an explicit distribution of the movement at each query position. The flow poke transformer cross-attends to features from a jointly trained image encoder to incorporate visual information. Crucially, our architecture represents movement at individual points $\query$ ({enabling} sparse \& off-grid motion processing) and directly predicts continuous, multimodal output distributions.}
    \vspace{-5mm}
    \label{fig:flow_poke_transformer}
\end{figure*}

\subsection{Problem Setting}

Given an image $\image$, we aim to model the movement of all visible points in the image and their interdependencies.
To this end, our goal is to model the conditional distribution $p(\fflow{}(\query)|\pokes,\image)$ of the movement $\fflow{}(\cdot) \in \mathbb{R}^2$ of arbitrary \textit{query} points $\query \in \mathbb{R}^2$ in the image conditioned on a set of $N_p$ \textit{pokes} $\pokes = \{(\poke_i, \fflow{}(\poke_i))\}_{i=1}^{N_p}$, each specifying the movement $\fflow{}(\poke_i)$ at locations $\poke_i \in \mathbb{R}^2$. Explicit conditioning on movement information given at specific points is crucial to enable the exploration of interactions and controlling movement prediction in the scene.
Here, the movement $\fflow{}(\cdot)$ of a point describes its change in position from the current time $t$ to a future time $t + \Delta t$, also referred to as \textit{forward flow}.

\subsection{Flow Poke Transformer}
To model the movement distribution $p_\theta(\fflow{}(\query)|\pokes,\image)$, we use a transformer-based architecture, denoted as $p_\theta$. Transformers~\cite{vaswani2017attention} are especially well-suited for this task, as they are well-capable of working with sparse sequences due to token interactions only being implemented via the attention mechanism. We show a high-level overview in \cref{fig:flow_poke_transformer}. We view each poke $(\poke_i, \fflow{}(\poke_i)) \in \pokes$ and each query point $\query_j \in \queries$ as individual tokens. Each poke's movement $\fflow{}(\poke)$ is encoded at the input using a Fourier embedding, while query tokens are set to a learned embedding. Positional encoding is implemented using relative positional embeddings~\cite{su2021roformer}, allowing positions to be set with arbitrary precision without needing to conform to any grid. This is important, as it enables training the model with high-quality but sparse and off-grid flow obtained via optical tracking. During self-attention, queries only attend to themselves and pokes, not to other queries. This enables evaluating the distribution $p_\theta(\fflow{}(\query)|\pokes,\image)$ for multiple queries $\query_j$ in parallel, which is crucial for efficient dense flow predictions. The image $\image$ is encoded separately using a vision transformer, resulting in a set of spatial encoded image tokens $\mathcal{E}(\image)$. The poke and query tokens then cross-attend to the image tokens, with spatial information again encoded using relative positional embeddings~\cite{su2021roformer}.

To obtain the movement distribution $p_\theta(\fflow{}(\query)|\pokes,\image)$, a projection head at the transformer's output directly predicts a Gaussian Mixture Model (GMM), enabling real-valued distributional predictions following GIVT~\cite{tschannen2024givt}.
The distribution being directly accessible in this manner enables a range of additional capabilities, such as directly capturing multi-modal distributions in a single forward pass or enabling the fine-grained quantification of uncertainty.
Unlike \cite{tschannen2024givt}, we parametrize each component $n$ using a full covariance matrix $\boldsymbol{\Sigma}^{(n)} \in \mathbb{R}^{2 \times 2}$ instead of a purely diagonal one, greatly increasing the prediction's degrees of freedom. The positive semi-definiteness of the covariance matrix is ensured by the model predicting a lower triangular matrix $\mathbf{L}^{(n)} \in \mathbb{R}^{2 \times 2}$ with a positive diagonal (by soft-clipping to a lower threshold), from which the covariance matrix is computed as $\boldsymbol{\Sigma}^{(n)} = \mathbf{L}^{(n)} (\mathbf{L}^{(n)})^\top$. Overall, this results in the predicted $N$-component GMM
\begin{equation}
    p_\theta = \textstyle \sum_{n=1}^{N} \pi^{(n)}\cdot\mathcal{N}(\boldsymbol{\mu}^{(n)}, \boldsymbol{\Sigma}^{(n)}),
\end{equation}
with component mixture coefficients $\pi^{(n)}$ and means $\boldsymbol{\mu}^{(n)}$.

A specific challenge with learning motion understanding from open-world web videos is that camera movement can dominate the overall motion distribution of a frame.
Only training on videos with static cameras is not viable, as it would limit potential training data too much. We address this by replacing the typical normalization layers in the transformer with adaptive normalization layers~\cite{huang2017arbitrary}, using which we condition the model on whether the camera is static, which we detect by whether a significant fraction of the scene's content is static.
This allows us to learn motion prediction on general videos.

\paragraph{Training Objective}
\label{par:training_object}
We directly train our model to minimize the negative log-likelihood (NLL) of a ground truth flow $\fflow{}(\query)$ of the random query point $\query$, conditioned on a random set of flow pokes $\pokes$
\begin{multline}
    \!\mathcal{L}(\fflow{}(\query), \pokes, \image; \theta) = -\log p_\theta(\fflow{}(\query)|\pokes,\image) \\
    \!= \textstyle \!-\!\log \Bigl(\sum_{n=1}^{N} \!\pi^{(n)} \!\mathcal{N}(\fflow{}(\query) | \boldsymbol{\mu}^{(n)}_\theta\!(\pokes, \image), \boldsymbol{\Sigma}^{(n)}_\theta\!(\pokes, \image))\!\Bigr).
\end{multline}
Specifically, we compute the loss for image $\image$ conditioned on random sets of pokes $\pokes^{(i)}$ of length $|\pokes^{(0)}| = 0, \ldots, |\pokes^{(N_p)}| = N_p$.
Predicting the flow distribution at $N_q$ different random query positions $\query$ per set of pokes. To enable efficient training, we introduce a variation of teacher forcing~\cite{sutskever2014seq2seq}, typically used to train autoregressive transformers~\cite{vaswani2017attention}. We select the set of random pokes such that $|\pokes^{(0)}| \subset |\pokes^{(1)}| \subset \ldots \subset |\pokes^{(N_p)}|$. We then use a causal attention mask on the poke tokens and let the queries for each set of pokes individually attend to all the pokes in their respective set. We call the resulting attention pattern \textit{query-causal attention} (see \cref{fig:query_causal_attention_pattern} for visualizations). As opposed to training with independent sets of pokes and full self-attention for all sets of pokes and queries, this reduces the computational complexity from {$\mathcal{O}({N_p^2 \cdot N_q^2})$ to $\mathcal{O}(N_p^2 + N_p \cdot N_q)$} for the same number of trained predictions. Since $N_p$ can be large during training, this substantially improves performance and enables efficient training.

\subsection{Downstream Applications}\label{subsec:downstream_apps}
Our method's primary goal is to enable efficient and interpretable modeling of multimodal movement distributions of different parts in scenes.
We achieve this by modeling the conditional distribution of the movement of query points $\query$ in the image $\image$ given the movement of any number of pokes $\pokes$ to condition on.
As our model directly makes probability density functions for each query's movement accessible and captures its multi-modality (c.f., \cref{fig:distributions_qualitative}), the distribution of potential movements can be directly interpreted. Besides modeling this relation of movement of different parts in a scene, this also enables other direct and indirect downstream tasks and applications, which we describe in this section.

\paragraph{Dense Motion Prediction} The conditional distribution our method learns to model can also be used to predict a dense grid of queries $\queries$. This prediction can be done both purely conditioned on the reference image or given reference pokes $\pokes$. This motion can be obtained by predicting the pointwise flow distributions in parallel or via autoregressive sampling.
To sample from the joint distribution $p(\fflow{}(\queries)|\pokes,\image)$ that models the precise interactions between all points in the image, we employ autoregressive sampling.
Iteratively, we predict the flow distribution for a random query $\query_i \in \queries$, sample a flow instance from the conditional distribution $\fflow{}(\query_i) \sim p_\theta(\fflow{}(\query_i)|\pokes,\image)$, and add the query to the set of pokes $\pokes \leftarrow \pokes \cup \{\query_i\}$. This results in individual coherent samples $\mathbf{F}_\text{sample} \sim p_\theta(\fflow{}(\queries)|\pokes,\image)$ from the distribution of possible dense flows. We show qualitative examples of sampled dense flow in \cref{fig:motion-prior-sampling}.
For parallel sampling, we compute the mean dense flow
$\mathbf{F}_\text{mean} = \mathbb{E}\left[\fflow{}(\queries)|\pokes,\image;\theta\right]$
in a pointwise manner for all queries $\query_i \in \queries$ in parallel. This provides efficient, high-quality flow predictions (see, e.g., appendix \cref{fig:app_dense}), but also results in mode averaging. 

\paragraph{Segmenting Moving Parts} Segmenting parts that move together is a task introduced in \cite{li2024dragapart} and is useful for various applications such as predicting affordances. Given a poke $(\poke, \fflow{}(\poke))$, the aim is to segment the image into parts that would move in response to it.
Unlike \cite{li2024dragapart}, we do not need to rely on involved methods for extracting and comparing internal feature activations of our model for this task.
Instead, our model enables direct quantification of the effect a movement $\fflow{}(\poke)$ of a point $\poke$ has on another point $\query$ by measuring the relative entropy between the conditional and unconditional distribution, i.e., how much conditioning on $\poke$ changes the movement distribution of $\query$. This is done using the Kullback-Leibler (KL) divergence
\begin{equation}\label{eq:moving_dependency_kl}
    D_\mathrm{KL}(p_\theta(\fflow{}(\query)|(\poke, \fflow{}(\poke)),\image) \parallel p_\theta(\fflow{}(\query)|\image)).
\end{equation}
Specifically, if the movements of $\query$ and $\poke$ are independent, the conditional distribution $p(\fflow{}(\query)|(\poke, \fflow{}(\poke)),\image)$ is equal to the marginal distribution $p(\fflow{}(\query)|\image))$, and thus, the KL divergence in \cref{eq:moving_dependency_kl} is zero. Otherwise, it quantifies the change in movement distribution, and, thus, the motion interdependencies of different parts of the scene.
We efficiently approximate the KL divergence using the matched bound approximation \cite{goldberger2003efficient}.
This can then be computed over all points $\query$ in the image $\image$ in parallel and directly quantifies the effect the given movement of $\poke$ has on each point $\query$.

\section{Experiments}

\subsection{Dataset and Implementation Details}
\label{sec:imp_details}
For general pretraining, we train on a random 3.8M video clip subset of WebVid~\cite{bain2021frozenintime}. The wide variety of concepts present in WebVid enables our model to learn a general representation for motion instead of being limited to a specific domain, such as face-only videos. We train our model with flow from optical tracks using CoTracker3~\cite{karaev2024cotracker3} for a random 48-frame interval from each clip using a uniform $48^2$ grid from the respective start frames.

The image encoder and the poke transformer are ViT-Base transformers \cite{dosovitskiy2021an} for a total parameter count of 220M. We use RoPE~\cite{su2021roformer,crowson2024hourglass} both for self-attention between flow tokens and for cross-attention to image tokens. We initialize the image encoder with DINOv2-R \cite{oquab2023dinov2,darcet2023vitneedreg} to make training more efficient but keep the weights unlocked. Jointly training the full model is essential, as DINOv2 does not have good instance segmentation capabilities (see \cref{sec:ablations} for additional details), which are essential for our task.
We pass images to the vision encoder at a resolution of $448^2$ to obtain a $32^2$ grid of visual embedding tokens.
The model is trained in bfloat16 precision for 800k steps using AdamW~\cite{loshchilov2018decoupled} with a learning rate of 5e-5, at a global batch size of 32 images, which is increased to 128 after 250k steps. Per image, we sample sets of random pokes of sizes $0, 1, \ldots, 128$, and compute losses on $N_q = 15$ random query points per set of pokes. This results in a global batch size of 61,440 (245,760) queries. Overall, training this model took 7 days on 2 Nvidia H200s.
{We also train a second model on a dataset of 5M open-set video clips we collected, with optical tracks obtained using TAPNext~\cite{zholus2025tapnext}. Here, we simplify the training setup to use a batch size of 128 across the whole training, and add a cosine decay~\cite{loshchilov2016sgdr} for the learning rate after an initial warmup. A further optimized training setup allows us to train this model to 1M steps in 24h on 8 Nvidia H200s.}
We provide additional details and ablations in appendix \cref{sec:app_imp_details,sec:ablations}, and explore the potential for extension to 3D motion in \cref{sec:3d-extension}.

Without inference optimizations such as quantization or 8-bit inference, a single conditional movement distribution prediction for query in an image can be obtained in less than 25ms of delay on a single H200. This makes our model applicable for real-time applications. Throughput (with parallel predictions) is about 160k predictions per second per image, thanks to our query-causal attention implementation.

We generally evaluate predicted motion at a resolution of $64^2$ unless specified otherwise. We primarily rely on endpoint error $\mathrm{EPE} = \|\hat{\fflow{}}(\query) - {\mathbf{f}_\mathrm{GT}}(\query)\|_2$, which measures the difference between the true motion ${\mathbf{f}_\mathrm{GT}}(\query)$ and the predicted motion $\hat{\fflow{}}(\query)$. Additionally, we also compute the percentage of correct keypoints $\mathrm{PCK} = \mathbb{E}[\|\hat{\fflow{}}(\query) - {\mathbf{f}_\mathrm{GT}}(\query)\|_2 < \alpha]$, with $\alpha = 1\mathrm{px}$ unless specified otherwise.

{
\setlength{\tabcolsep}{0.01\linewidth}
\begin{figure}[t]
    \centering

    \includegraphics[width=\linewidth]{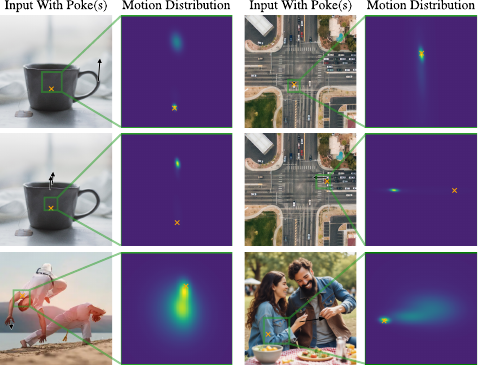}
    \vspace{-4mm}
    \caption{\textbf{Multimodal Motion Distribution Prediction}. We condition on one or multiple pokes ($\rightarrow$) and then query the motion distribution of specific points ({\color{ourorange}$\boldsymbol{\times}$}). Our model's predictions capture the multi-modal nature of motion and exhibit understanding of interactions, such as only lifting the cup by its handle not necessarily causing the whole cup to move upwards, while grabbing it at stable points does. It also demonstrates prior understanding from scenes, such as a car in an intersection being more likely to move forwards than backward and cars in traffic likely moving together.}
    \vspace{-4mm}
    \label{fig:distributions_qualitative}
\end{figure}
}

\subsection{{Evaluation of FPT's Key Abilities}}
\paragraph{{FPT's Ability to Predict Movement Distributions}}
{We observe our model's predicted distributions $p_\theta(\fflow{}(\query)|\pokes,\image)$ given an image $\image$ of a scene and conditioned on a sparse set of pokes $\pokes$ in \cref{fig:distributions_qualitative} ({see appendix \cref{fig:app_distribution} for additional samples}). Qualitatively, our model exhibits an understanding of physical phenomena and interactions, predicting realistic movement distributions for the given pokes in the context of the respective scenes. Most importantly, it captures the multi-modality of potential movements in different circumstances and their variability/uncertainty.} {Our approach is trained in an open-world setting, being not limited to individual object categories, but, nevertheless, captures fine-grained details of specific objects' potential motion.}

One can also draw samples from the joint distribution of movement of the whole scene $p_\theta(\fflow{}(\queries)|\pokes,\image)$ by autoregressive sampling.
We show examples of such unconditional motion generations in \cref{fig:motion-prior-sampling}. They successfully show diverse but realistic global scene motion.

\vspace{-3mm}

\begin{figure}[h]
    \centering
    \includegraphics[width=\linewidth]{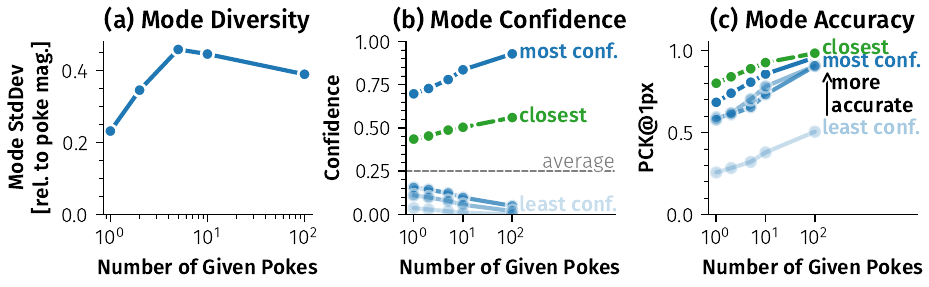}
    \vspace{-6mm}
    \caption{\textbf{Predicted Mode Analysis.} Values are computed at a resolution of $64^2$. \textbf{(a)} Diversity of predicted modes is high, with mode variation covering a large fraction of poke magnitude. \textbf{(b)} One mode typically has a substantially higher confidence than others, which increases with given poke count. The mode {\color{matplotlibgreen}\textbf{closest}} to the ground truth consistently has a higher-than-average confidence. \textbf{(c)} More confident modes are more accurate as measured by PCK.}
    \label{fig:mode_calibration}
    \vspace{-6mm}
\end{figure}

\paragraph{{FPT predicts meaningful multimodal motion distributions.}}
One important property that differentiates FPT from common motion modeling approaches is that it \textit{directly} predicts the multimodality of possible future motions. For these multimodal predictions to be valuable, they should cover the diverse modes of possible motion and have meaningful predicted {confidences} $\pi^{(n)}$. We analyze these properties in \cref{fig:mode_calibration}. Generally, we find the modes to be highly diverse (\cref{fig:mode_calibration}a, cf. \cref{fig:distributions_qualitative,fig:app_distribution}), with them covering substantially different movements.
As expected, the multimodal predictions reduce to primarily unimodal predictions when enough conditioning information is available to reduce the stochastic uncertainty of the future and discern one clear correct mode (\cref{fig:mode_calibration}b). Importantly, the confidence of the mode closest to the ground truth motion is consistently substantially higher than the average. This indicates that the model's confidence predictions {are meaningful}. Similarly, analyzing the modes' accuracy (\cref{fig:mode_calibration}c) shows that the model assigns higher confidences to modes more likely to be correct. Still, secondary and tertiary predicted modes are also meaningful, as indicated by the accuracy of the mode closest to the ground truth exceeding that of the most confident one.

\begin{table*}[ht]
    \centering
    \newcommand{\inte}{\color{ourgray}}
    {
    \adjustbox{max width=\linewidth}{
    \begin{tabular}{lcc@{\hskip 1.5mm}c@{\hskip 1.5mm}cc@{\hskip 1.5mm}c@{\hskip 1.5mm}cc@{\hskip 1.5mm}c@{\hskip 1.5mm}cc@{\hskip 1.5mm}c@{\hskip 1.5mm}cc@{\hskip 1.5mm}c@{\hskip 1.5mm}c}
        \toprule
        \multirow{2}{*}[-3pt]{Method} & \multirow{2}{*}[-3pt]{\shortstack{Trained On}} & \multicolumn{3}{c}{1 Poke} & \multicolumn{3}{c}{2 Pokes} & \multicolumn{3}{c}{5 Pokes} & \multicolumn{3}{c}{10 Pokes} & \multicolumn{3}{c}{100 Pokes} \\
        \cmidrule(lr){3-5} \cmidrule(lr){6-8} \cmidrule(lr){9-11} \cmidrule(lr){12-14} \cmidrule(lr){15-17}
        & & {\footnotesize EPE $\downarrow$} & {\footnotesize PCK $\uparrow$} & {\footnotesize LPIPS $\downarrow$} & {\footnotesize EPE $\downarrow$} & {\footnotesize PCK $\uparrow$} & {\footnotesize LPIPS $\downarrow$} & {\footnotesize EPE $\downarrow$} & {\footnotesize PCK $\uparrow$} & {\footnotesize LPIPS $\downarrow$} & {\footnotesize EPE $\downarrow$} & {\footnotesize PCK $\uparrow$} & {\footnotesize LPIPS $\downarrow$} & {\footnotesize EPE $\downarrow$} & {\footnotesize PCK $\uparrow$} & {\footnotesize LPIPS $\downarrow$} \\
        \midrule
        {InstantDrag~\cite{shin2024instantdrag} } & Faces & \underline{9.24} & \textbf{0.193} & \underline{0.18} & \underline{9.12} & \textbf{0.196} & \underline{0.17} & \underline{8.82} & \textbf{0.197} & \underline{0.17} & \underline{8.39} & \textbf{0.198} & \underline{0.16} & \underline{7.29} & \underline{0.212} & \underline{0.15} \\
        Motion-I2V~\cite{shi2024motion} & \textbf{Generic (WebVid-10M, Zero-Shot)} & 29.08 & 0.029 & 0.35 & 27.40 & 0.031 & 0.34 & 24.22 & 0.030 & 0.32 & 20.90 & 0.048 & 0.30 & n/a & n/a & n/a \\
        \multirow{2}{*}{\textbf{Ours}}  & \textbf{Generic (WebVid-3.8M, Zero-Shot)} & \textbf{7.64} & \underline{0.150} & \textbf{0.16} & \textbf{6.87} & \underline{0.154} & \textbf{0.15} & \textbf{5.32} & \underline{0.167} & \textbf{0.13} & \textbf{4.20} & \underline{0.183} & \textbf{0.12} & \textbf{2.51} & \textbf{0.264} & \textbf{0.10} \\
        & \inte\textbf{Generic (Open Set-5M, Zero-Shot)} & \inte 7.99 & \inte 0.150 & \inte 0.17 & \inte 7.02 & \inte 0.154 & \inte 0.16 & \inte 5.50 & \inte 0.158 & \inte 0.14 & \inte 4.17 & \inte 0.182 & \inte 0.12 & \inte 2.44 & \inte 0.285 & \inte 0.10 \\
        \bottomrule
    \end{tabular}
    }}
    \vspace{-2mm}
    \caption{\textbf{Face Motion Generation Evaluation}.
    We evaluate the accuracy of predicted motion on TalkingHead-1KH~\cite{wang2021facevid2vid} given a starting frame and one or more pokes (partially) defining the head movement. Our method performs substantially better in a zero-shot comparison to Motion-I2V, which was also trained in a generic setting. Compared to InstantDrag, which was trained for specifically this setting, our method achieves a substantially better endpoint error (EPE) but slightly worse PCK for low poke counts, highlighting our model's capability to perform competitively with purpose-trained methods while being generic. It can also make more efficient use of the available information, achieving greater accuracy gains from additional pokes compared to other methods. When using the predicted motions to warp the source image, our method consistently outperforms others.
    }
    \label{tab:face_flow_quantitative}
    \vspace{-4mm}
\end{table*}

\vspace{-2mm}

\begin{figure}[t]
    \centering
    \includegraphics[width=\linewidth]{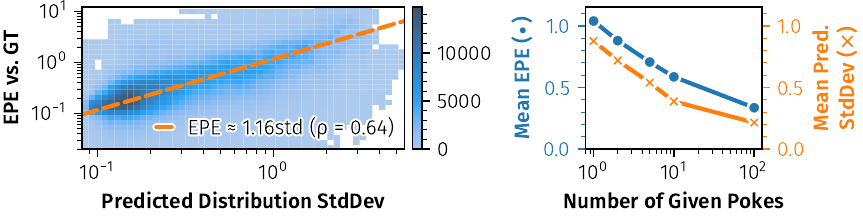}
    \vspace{-6mm}
    \caption{\textbf{Uncertainty Calibration.}
        We find that the motion prediction error measured by EPE strongly correlates with the predicted uncertainty (Pearson $\rho \!=\! {0.64}$). This relationship holds for low \& high numbers of given pokes.
    }
    \label{fig:uncertainty_calibration}
    \vspace{-6mm}
\end{figure}

\begin{figure}[H]
    \vspace{-2mm}
    \centering
    {
    \setlength{\tabcolsep}{0.01\linewidth}
    \newcommand{\imgwidth}{0.2\linewidth}
    \begin{tabular}{@{}c@{\hskip .5em}c@{\hskip .2em}c@{\hskip .2em}c@{}}
        {\footnotesize Input Image} & \multicolumn{3}{c}{\footnotesize Unconditional Motion Samples} \\
        {\includegraphics[width=\imgwidth]{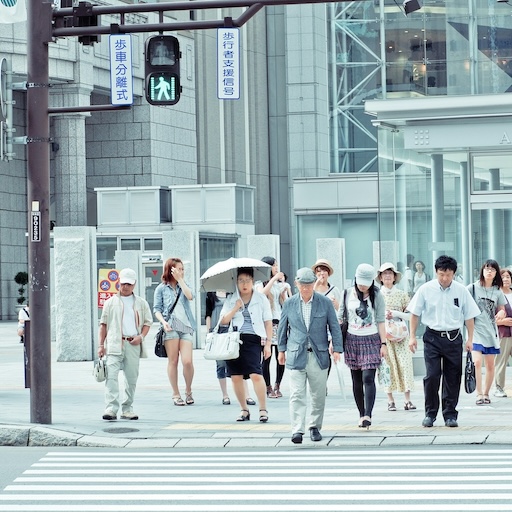}} &
        \includegraphics[width=\imgwidth]{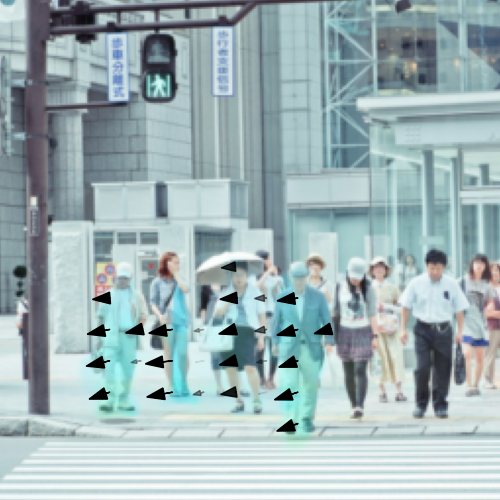} &
        \includegraphics[width=\imgwidth]{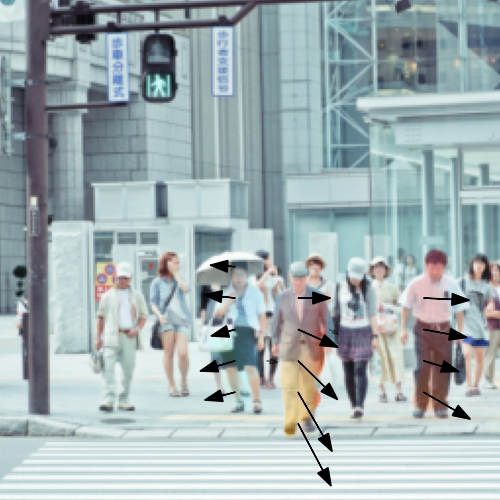} &
        \includegraphics[width=\imgwidth]{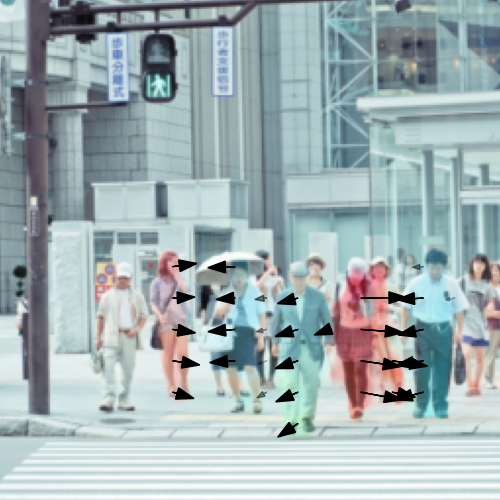} \\[-0.2em]

        {\includegraphics[width=\imgwidth]{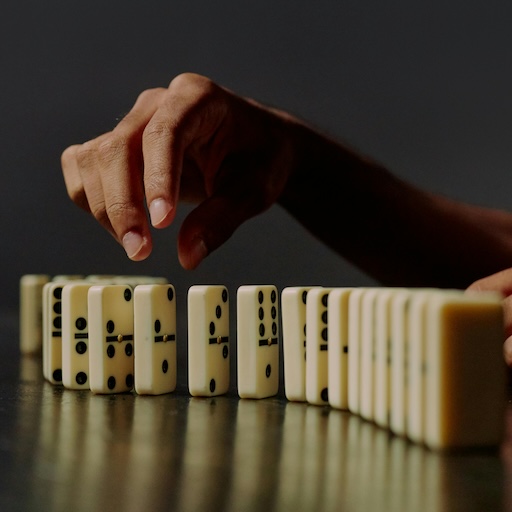}} &
        \includegraphics[width=\imgwidth]{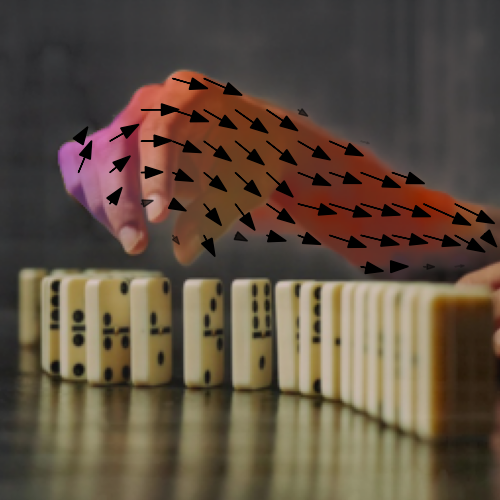} &
        \includegraphics[width=\imgwidth]{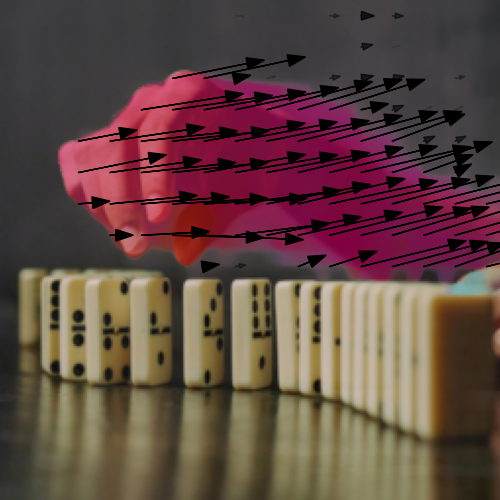} &
        \includegraphics[width=\imgwidth]{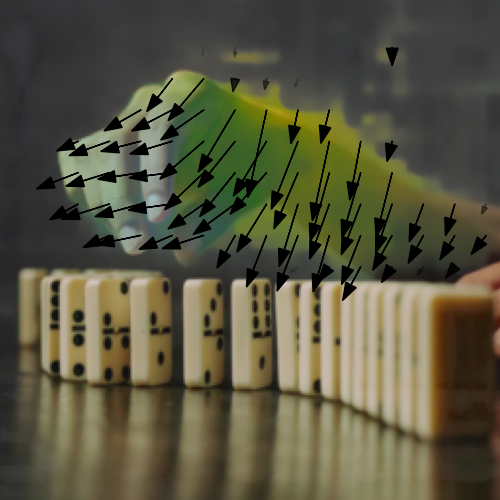} \\[-0.2em]

        {\includegraphics[width=\imgwidth]{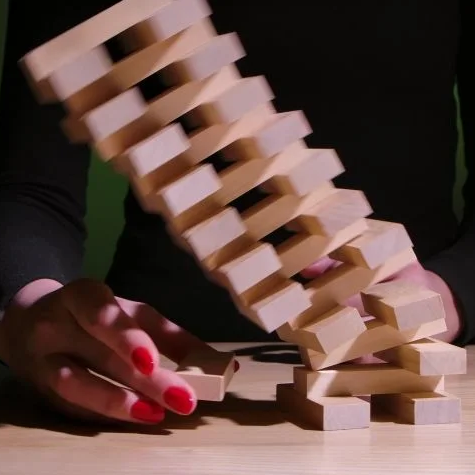}} &
        \includegraphics[width=\imgwidth]{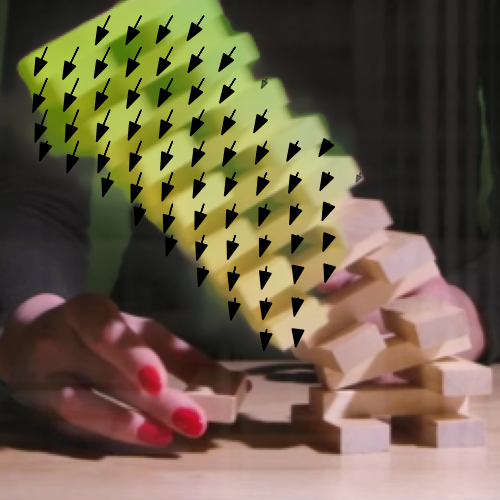} &
        \includegraphics[width=\imgwidth]{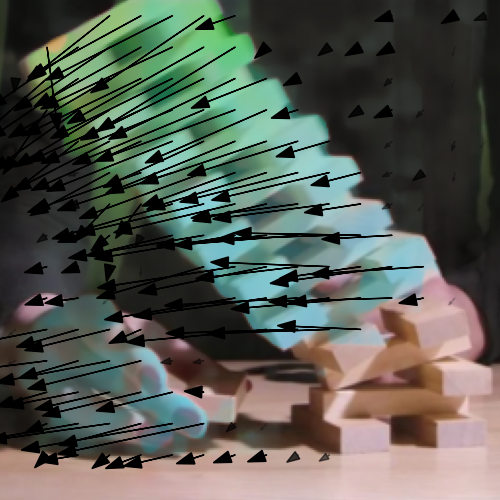} &
        \includegraphics[width=\imgwidth]{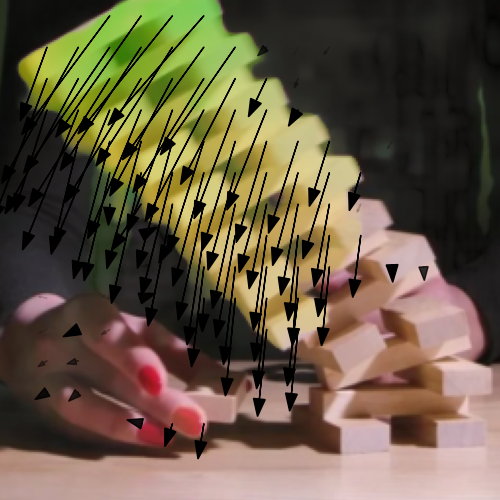} \\
    \end{tabular}
    }
    \vspace{-3mm}
    \caption{\textbf{Unconditional AR Motion Sampling}. We show samples of generated flow without prior motion conditioning on pokes. Our model can generate a wide variety of realistic motions.}
    \vspace{-3mm}
    \label{fig:motion-prior-sampling}
\end{figure}

\paragraph{{FPT's predicted distributions accurately model uncertainty.}}
We evaluate the predictive quality of our model's predicted uncertainty w.r.t. true prediction error in \cref{fig:uncertainty_calibration}. Specifically, we investigate the relation between the predicted distribution's standard deviation
    $\mathrm{Std}[\fflow{}(\query)|\pokes,\image; \theta]$
and the motion estimation error as measured with the endpoint error (EPE). We find that the predicted motion's error strongly correlates to the predicted uncertainty. This capability is independent of the approach to derive the single motion prediction from the predicted distribution.
Sampling from the predicted distribution (Pearson $\rho \!=\! {0.66}$), using its mean ($\rho \!=\! {0.64}$), or using the most confident mode ($\rho \!=\! {0.62}$) all lead to high predictive accuracy of the true prediction error compared to the ground truth.

\subsection{Comparisons on Motion Prediction}
{To enable quantitative comparisons of our model's motion understanding with existing methods, we evaluate on predicting dense flow from sparse flow pokes, given a starting frame.} We compare against baseline methods in the setting they have been trained on to enable fair comparisons. To evaluate against DragAPart~\cite{li2024dragapart} and PuppetMaster~\cite{li2024puppet}, which only generate images/videos based on drags, we extract flow from the source to the generated image using RAFT~\cite{teed2020raft}.

\begin{figure}[t]
    \centering
    \vspace{-6mm}
    {
    \setlength{\tabcolsep}{0.01\linewidth}
    \setlength{\aboverulesep}{0pt}\setlength{\belowrulesep}{0pt}
    \begin{tabular}{@{}c@{\hskip .35em}c@{\hskip .15em}c@{\hskip .5em}c@{\hskip .15em}c@{}}
        \multirow{2}{*}[-3pt]{\footnotesize\shorttabular{Input\\with Pokes}}  &  \multicolumn{2}{c}{\footnotesize Ours} & \multicolumn{2}{c}{\footnotesize InstantDrag~\cite{shin2024instantdrag}} \\
        \cmidrule(lr){2-3} \cmidrule(lr){4-5}
         & \footnotesize Flow & \footnotesize Warped & \footnotesize Flow & \footnotesize Warped \\
        \includegraphics[width=0.15\linewidth]{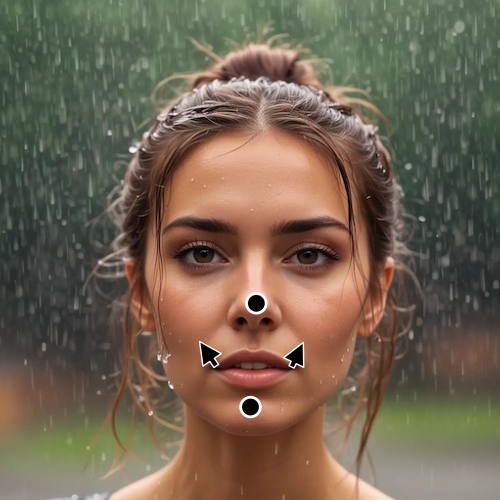} &
        \includegraphics[width=0.15\linewidth]{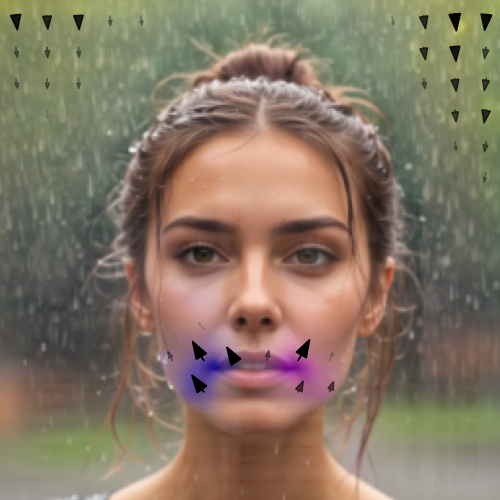} &
        \includegraphics[width=0.15\linewidth]{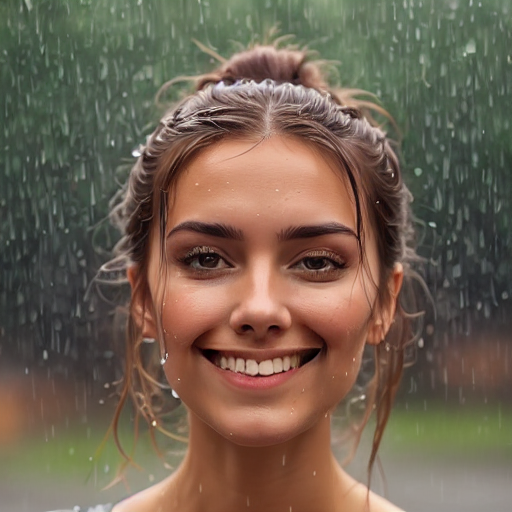} &
        \includegraphics[width=0.15\linewidth]{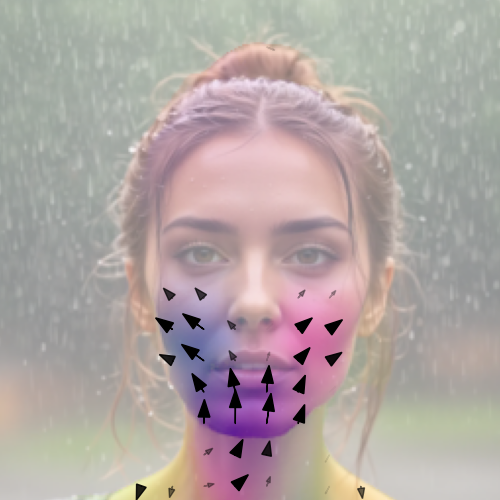} &
        \includegraphics[width=0.15\linewidth]{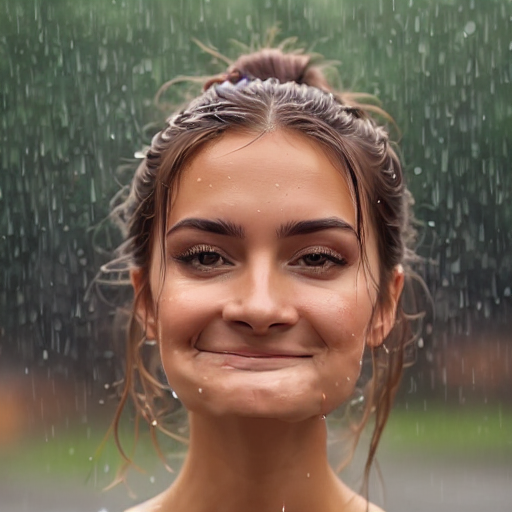} \\[-0.2em]
        
        \includegraphics[width=0.15\linewidth]{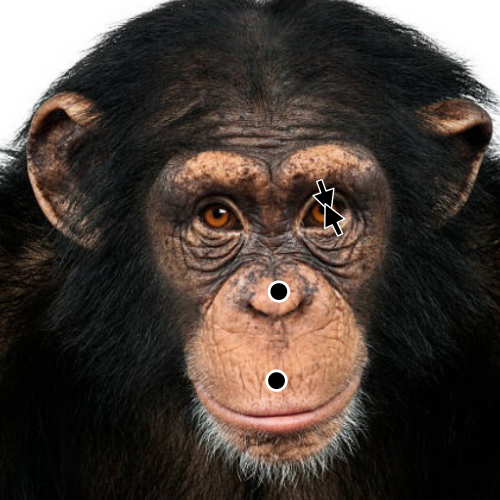} &
        \includegraphics[width=0.15\linewidth]{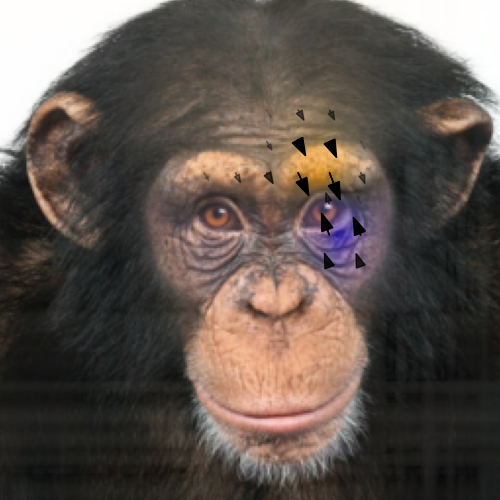} &
        \includegraphics[width=0.15\linewidth]{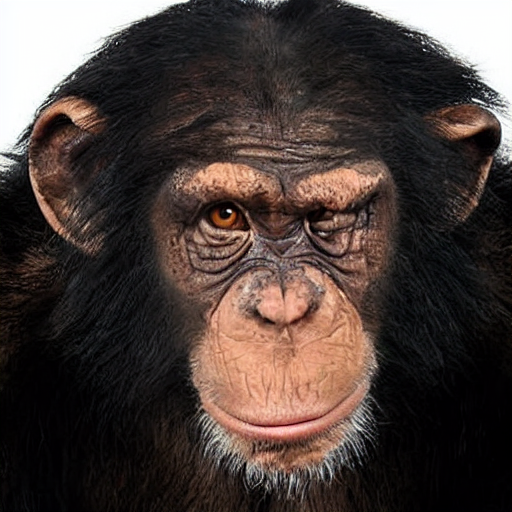} &
        \includegraphics[width=0.15\linewidth]{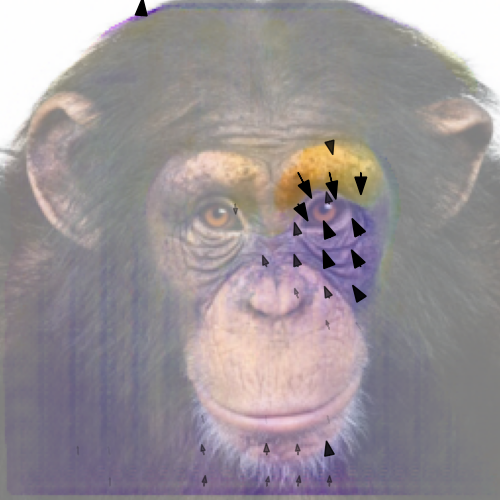} &
        \includegraphics[width=0.15\linewidth]{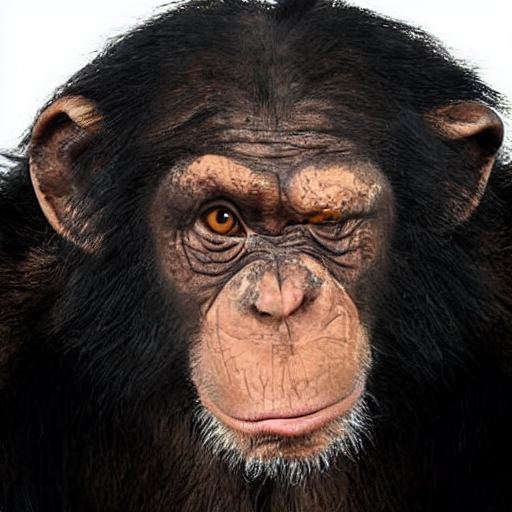} \\[-0.2em]

        \includegraphics[width=0.15\linewidth]{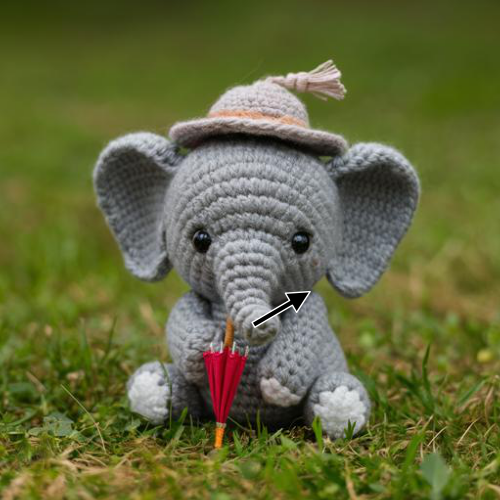} &
        \includegraphics[width=0.15\linewidth]{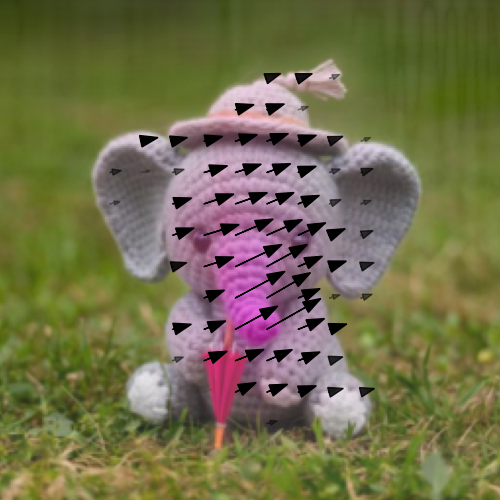} &
        \includegraphics[width=0.15\linewidth]{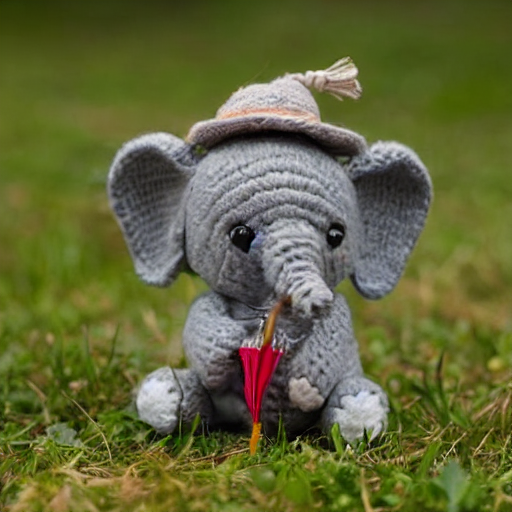} &
        \includegraphics[width=0.15\linewidth]{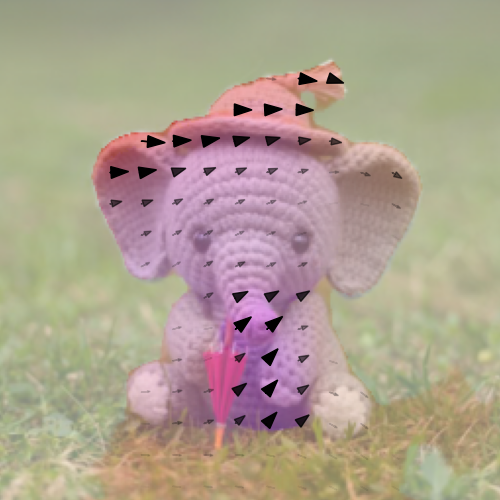} &
        \includegraphics[width=0.15\linewidth]{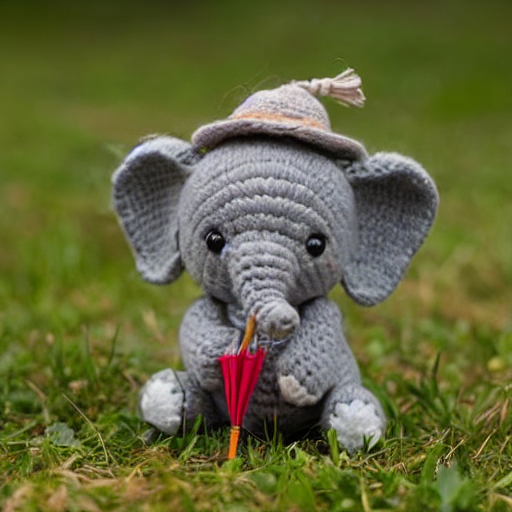} \\
    \end{tabular}
    }
    \vspace{-3mm}
    \caption{\textbf{Fine-grained Face Motion Control}. We show fine-grained zero-shot poking results on faces and compare against InstantDrag~\cite{shin2024instantdrag}, which was trained for this task. We further visualize the predicted motion as warps using I-D's face warping model.}
    \label{fig:face_distributions_qualitative}
    \vspace{-6mm}
\end{figure}

\paragraph{Face Motion Estimation}
We compare against InstantDrag~\cite{shin2024instantdrag}, which was trained on CelebV-Text~\cite{yu2023celebv} in their evaluation setting -- poke-conditioned motion prediction on aligned faces on TalkingHead-1KH~\cite{wang2021facevid2vid}.
The qualitative results (see \cref{fig:face_distributions_qualitative}) show that our model tends to predict more accurate and localized motion. This can also be observed when visualizing the motion by warping the image. For quantitative evaluations, we extract chunks of length 0.8s (following \cite{shin2024instantdrag}) and use CoTracker3~\cite{karaev2024cotracker3} at a grid size of $128^2$ to obtain the target motion from the start to the end frame. Then, we condition on $N \in \{1, 2, 5, 10, 100\}$ pokes $\pokes$, where the first poke is chosen to be the one with the largest flow magnitude, and the others are sampled randomly, and compare the dense predicted motion to the target downsampled by a factor of two. We also compare quantitatively with \cite{shi2024motion}, which was trained generically. We compare favorably to both methods in EPE independent of the number of given pokes, indicating further that our predicted flow is more precise (\cref{tab:face_flow_quantitative}). The PCK is slightly worse than the non-generically trained InstantDrag for low poke counts but catches up with more conditioning. When using the respective method's generated motion to perform image warping using InstantDrag's warping stage, our method consistently outperforms both others, as measured by the LPIPS~\cite{zhang2018unreasonable} distance to the ground truth images.

\vspace{-2mm}
\paragraph{Articulated Objects}
DragAPart~\cite{li2024dragapart} and PuppetMaster~\cite{li2024puppet} were explicitly trained on part-level object motion on a synthetic dataset, Drag-A-Move~\cite{li2024dragapart} (DAM). To enable comparisons with them, we evaluate on the test set of DAM, which contains synthetic images of furniture with one or more pokes $\mathcal{P}$ that define the movement of one or more articulated parts of that object. It also provides ground truth dense flow at a resolution of $512^2$, which we compare against. As DAM is significantly dissimilar from our train set, we evaluate our model both in a zero-shot and fine-tuned setting. For the fine-tuned setting, we fine-tune our model on DAM for 30k steps at a batch size of 128 with an exponentially decaying learning rate of 5e-7 that halves every 10k steps. Our quantitative evaluation (see \cref{tab:articulated_objects_quantitative}, for additional qualitative samples see appendix \cref{fig:dragamove_qualitative}) shows that, in a zero-shot setting, our model's predicted motion is substantially more accurate than Motion-I2V~\cite{shi2024motion} but less accurate than the specifically trained DragAPart and PuppetMaster. This demonstrates that our model generalizes to out-of-distribution data comparatively well, but, expectedly, falls short of models explicitly trained for this OOD domain. Once fine-tuned, our model outperforms even the purpose-made DragAPart and PuppetMaster by a wide margin. We also show qualitative examples in \cref{fig:dragamove_qualitative}. These find Motion-I2V to hallucinate substantial motion independently from the given pokes and struggle with matching the precise flow direction and magnitude for the given pokes. The extracted flow from DragAPart also seems to face struggles in complex multi-poke situations, which our model handles better.

\begin{table}[t]
    \centering
    \adjustbox{max width=\columnwidth}{
    \begin{tabular}{lcc@{\hskip 1.5mm}c@{}c}
        \toprule
        \multirow{2}{*}[-3pt]{Method} & \multirow{2}{*}[-3pt]{\shortstack{Trained On}} & \multicolumn{2}{c}{\textbf{(a)} Motion Est.} & \textbf{(b)} Moving Part Segm.\\
        \cmidrule(lr){3-4} \cmidrule(lr){5-5}
        & & { EPE $\downarrow$} & { PCK $\uparrow$} & {mIoU $\uparrow$} \\
        \midrule
        {Motion-I2V~\cite{shi2024motion}} & \textbf{Generic (Zero-Shot)} & 33.27 & 0.043 & 0.073 \\
        DragAPart~\cite{li2024dragapart} & Objects (DAM) & {9.69} & \underline{0.514} & {\footnotesize\ \ }{0.273$^\dagger$} \\
        PuppetMaster~\cite{li2024puppet} & Objects (DAM + OAHQ) & \underline{9.62} & 0.472 & 0.112 \\
        \textbf{Ours}  & \textbf{Generic (WV, Zero-Shot)} & 12.74 & 0.191 & \underline{0.287} \\
        \textbf{Ours} (fine-tuned) & Generic $\rightarrow$ DAM & \textbf{3.57} & \textbf{0.834} & \textbf{0.572} \\
        \bottomrule
        \multicolumn{5}{c}{{\small $^\dagger$taken from original publication, our evaluation yields 0.228.}}
    \end{tabular}
    }
    \vspace{-3mm}
    \caption{\textbf{Articulated Object Motion Estimation.} We compare motion (flow) estimation (a) and moving part segmentation performance (b) on Drag-A-Move~\cite{li2024dragapart} (DAM). On zero-shot motion estimation, our model substantially outperforms the other zero-shot method M-I2V, while not being much worse than specifically trained methods. When adapted, our method significantly outperforms previous approaches. In moving part segmentation, even our generic model outperforms other, in-domain models.}
    \label{tab:articulated_objects_quantitative}
    \vspace{-3mm}
\end{table}

\subsection{Segmenting Moving Parts}
\begin{figure}[t]
    \centering
    \begin{subfigure}[t]{\columnwidth}
        \centering
        \adjustbox{max width=\columnwidth}{
            \begin{minipage}{1.18\columnwidth}
                \hspace{-4mm}
                \begin{tabularx}{\columnwidth}{X@{}X@{}X@{}X@{}X@{}X}
                     \multirow{2}{*}[-3pt]{\ {\footnotesize\shorttabular{Input\\with Poke}}} & \multicolumn{2}{c}{\footnotesize Moving Part Segmentation}
                     & \multirow{2}{*}[-3pt]{\ {\footnotesize\shorttabular{Input\\with Poke}}} & \multicolumn{2}{c}{\footnotesize Moving Part Segmentation} \\
                     \cmidrule{2-3} \cmidrule{5-6}
                     & \makecell[c]{\footnotesize \ \ \ DragAPart} & \makecell[c]{\footnotesize \ \ \ \ \ {Ours}} & & \makecell[c]{\footnotesize \ \ \ DragAPart} & \makecell[c]{\footnotesize \ \ \ \ \ {Ours}} \\[0.15em]
                \end{tabularx}
            \end{minipage}
        }
        \adjustbox{max width=\columnwidth}{
            \includegraphics[width=\linewidth]{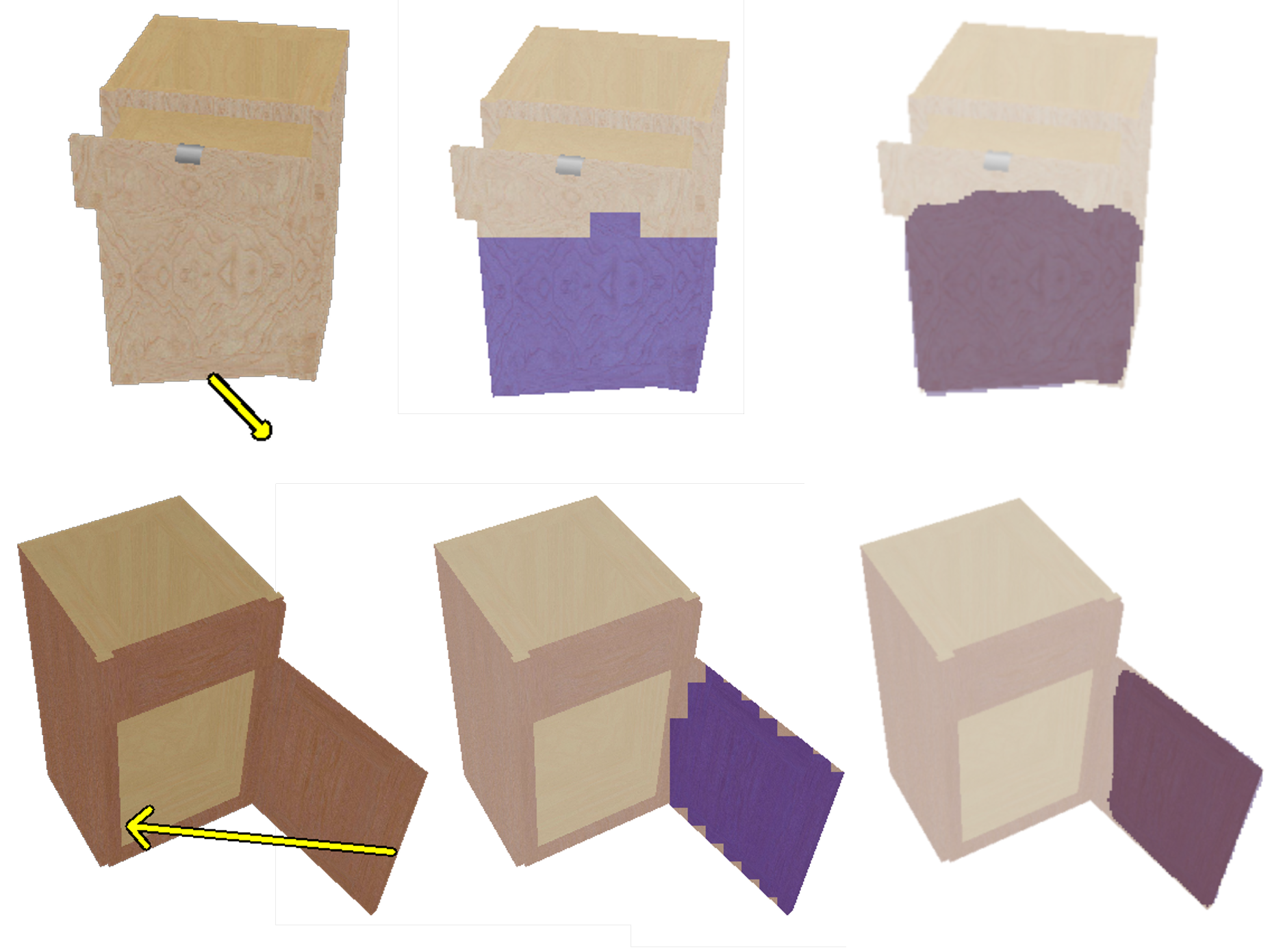}
            \includegraphics[width=\linewidth]{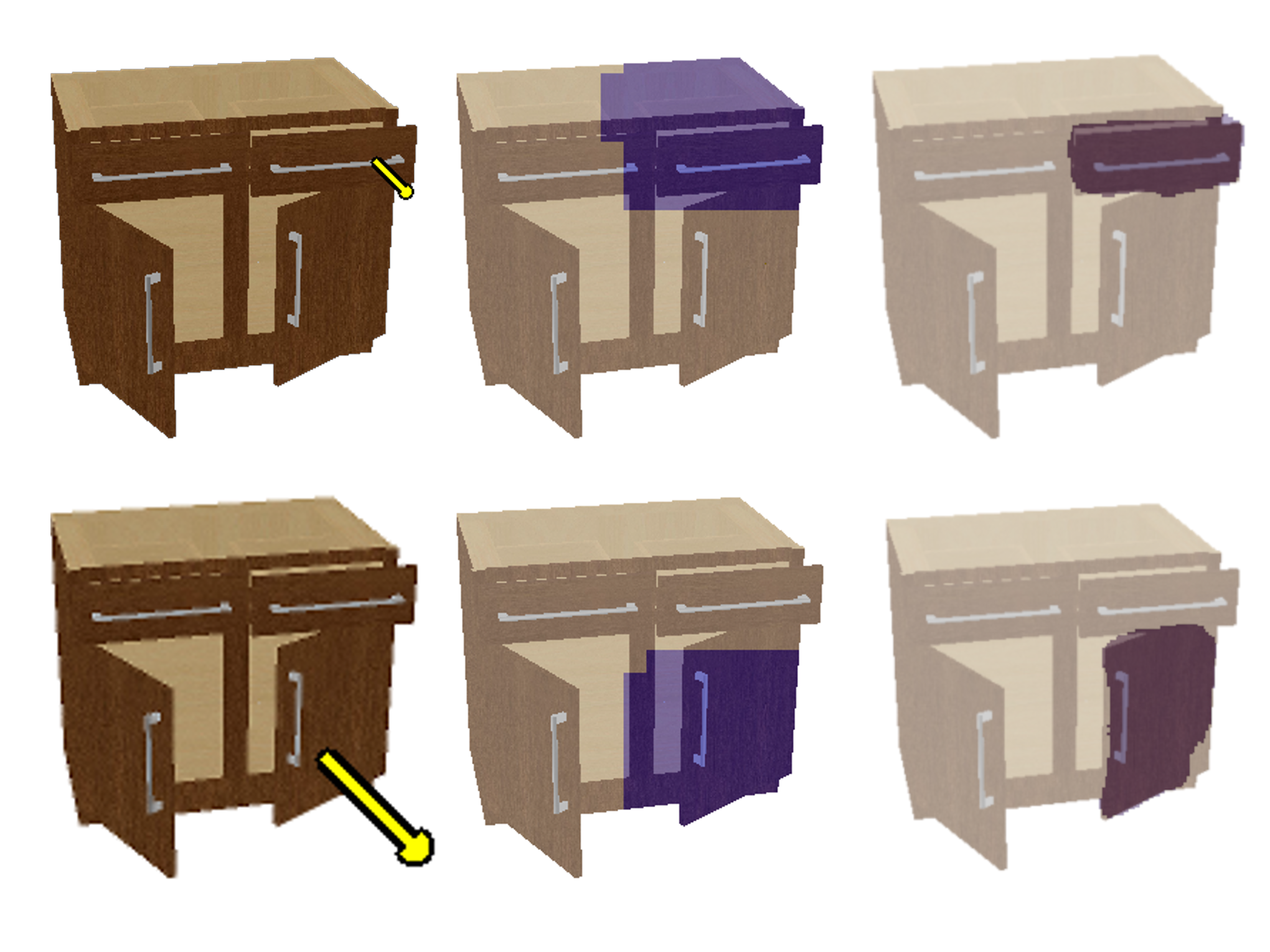}
        }
        \caption{We directly replicate Fig.\ 7 from DragAPart~\cite{li2024dragapart} with our method. Our method provides spatially continuous predictions and makes fewer critical mistakes like segmenting the furniture body with the drawer (top right).}
    \end{subfigure}
    ~
    \begin{subfigure}[t]{\columnwidth}
        \centering
        \adjustbox{max width=\columnwidth}{
        \begin{tabular}{@{}c@{\hskip .15em}c@{\hskip .5em}c@{\hskip .15em}c@{\hskip .5em}c@{\hskip .15em}c@{}}
            \scriptsize Input & \scriptsize Prediction & \scriptsize Input & \scriptsize Prediction & \scriptsize Input & \scriptsize Prediction \\
            \includegraphics[width=.16\columnwidth]{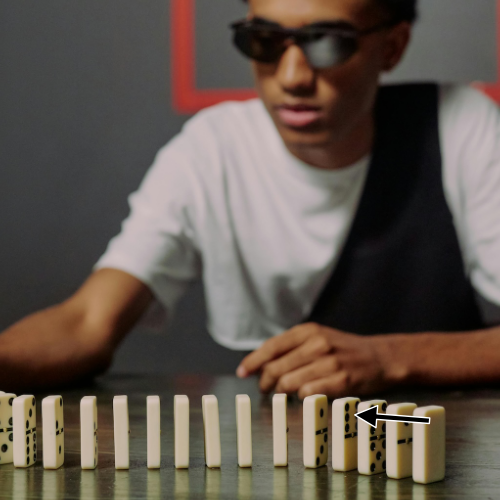} & \includegraphics[width=.16\columnwidth]{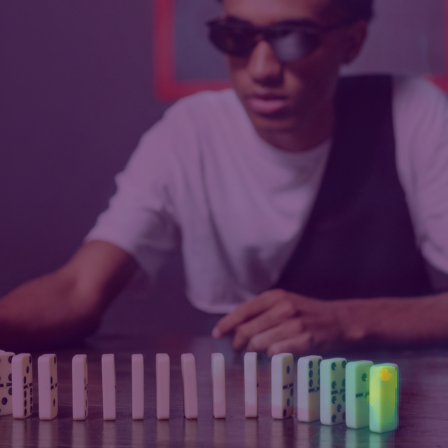}
            & \includegraphics[width=.16\columnwidth]{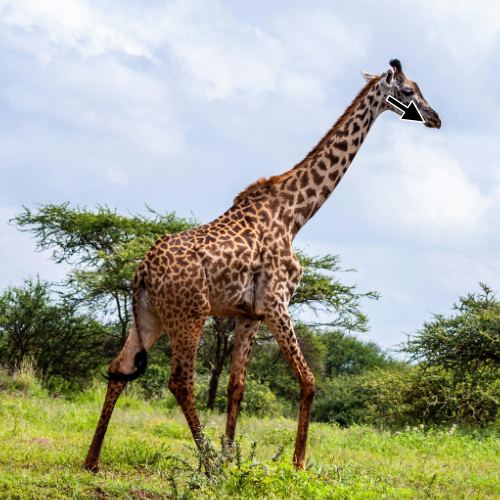} & \includegraphics[width=.16\columnwidth]{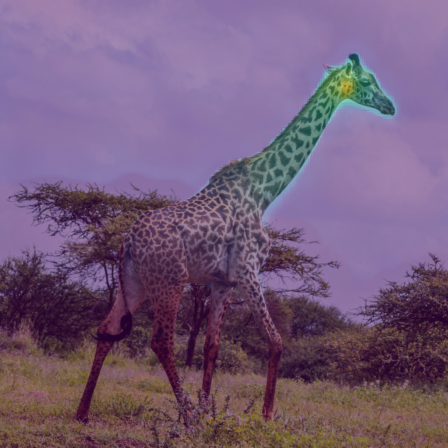}
            & \includegraphics[width=.16\columnwidth]{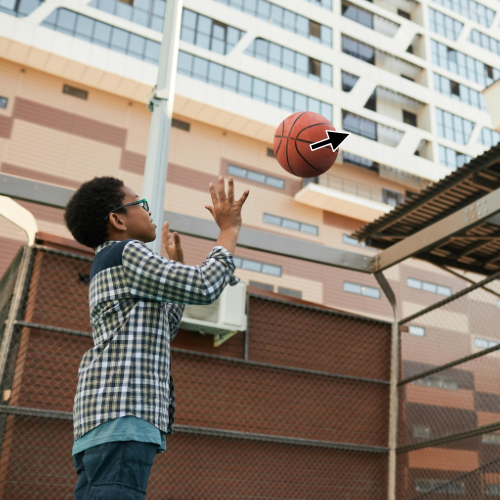} & \includegraphics[width=.16\columnwidth]{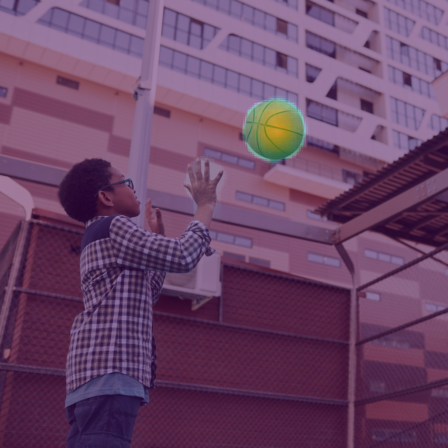}
        \end{tabular}
        }
        \caption{Open-set moving part dependency visualization. The degree to which the movement of each part is influenced by the poke ($\rightarrow$) is visualized as a heatmap, where brighter color means a higher degree of influence.}
    \end{subfigure}
    \vspace{-1.5mm}
    \caption{\textbf{Segmenting Moving Parts}. We show qualitative results for moving part segmentation, as introduced in \cite{li2024dragapart}, both on the Drag-A-Move dataset (a) and in a generic, open-set setting (b).}
    \label{fig:moving_part_segmentation}
    \vspace{-3mm}
\end{figure}

We perform moving part segmentation with our method by thresholding the KL divergence between the pointwise unconditional motion distribution and the pointwise motion distribution conditioned on a specific poke (\cref{eq:moving_dependency_kl}).
We show qualitative results in \cref{fig:moving_part_segmentation} and compare quantitatively against other methods, similarly thresholding the flow magnitude in \cref{tab:articulated_objects_quantitative}b. Here, we find that our method, especially when finetuned in-domain, outperforms DragAPart~\cite{li2024dragapart}, which introduced this benchmark, and the other methods by a wide margin. A large part of this gain can be attributed to our divergence-based score, which leverages FPT's unique property of directly predicting distributions. Without it, our method achieves an mIoU of 0.415 -- still outperforming the previous state-of-the-art by a wide margin, but less so. Our method is also robust to threshold choice -- halving/doubling it leads to mIoUs of 0.54/0.52 respectively.

\section{Conclusion}
We introduce the Flow Poke Transformer (FPT), a novel framework for motion understanding that captures real-world dynamics' multi-modal and stochastic nature through interpretable distributions of local motion, conditioned on targeted interactions (\textit{pokes}). Contrary to previous motion prediction approaches, FPT directly models the probabilistic distribution of possible outcomes, providing insights into the effects of physical interactions and inherent uncertainties.

\begin{figure}[t]
    \centering
    \begin{tabular}{c@{\hskip 0.5em}cc@{\hskip 0.5em}c}
        {\footnotesize Input} & {\footnotesize Prediction} & {\footnotesize Input} & {\footnotesize Prediction} \\
        \includegraphics[width=0.2\linewidth]{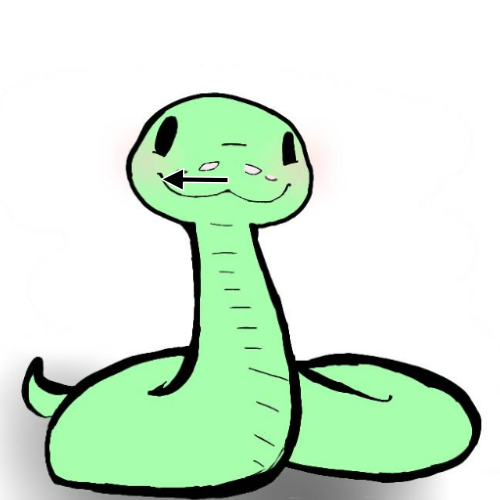} & \includegraphics[width=0.2\linewidth]{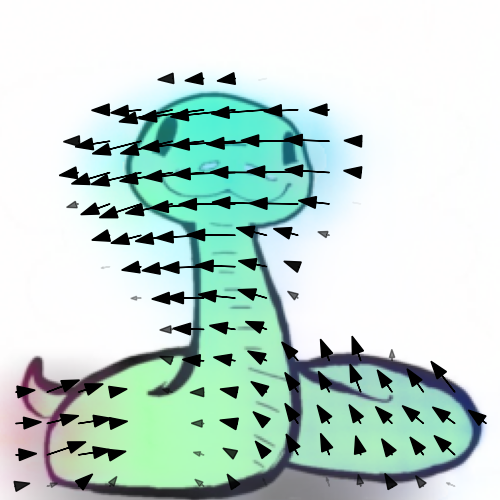} &
        \includegraphics[width=0.2\linewidth]{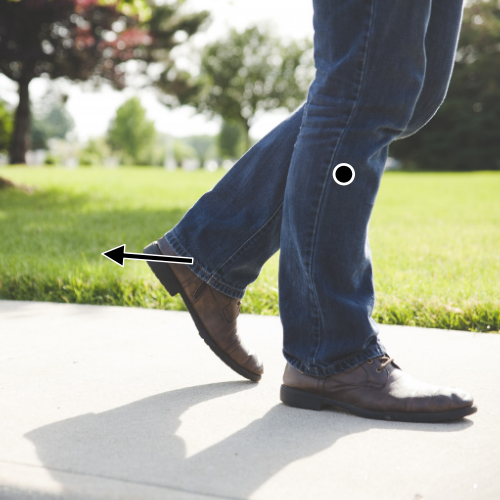} & \includegraphics[width=0.2\linewidth]{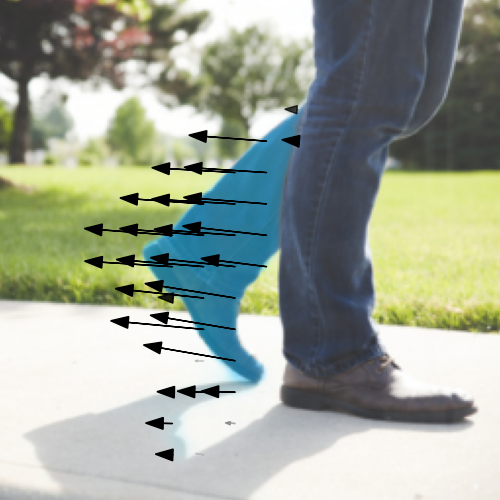}
    \end{tabular}
    \vspace{-2mm}
    \caption{\textbf{Common Failure Cases}. Our model, generically pretrained on primarily realistic videos, does not generalize well to cartoons, causing parts of the background to be moved together with objects. Additionally, our model sometimes, but not consistently, jointly predicts the movement of shadows together with objects, which can be problematic for downstream use cases.}
    \label{fig:failure_cases}
\end{figure}

Our evaluations demonstrate FPT’s versatility across different domains and its generalization capabilities. Despite being designed and trained for sparse, general-purpose motion understanding, it also offers competitive performance in established tasks such as dense motion generation on faces or articulated objects. Importantly, while valuable for comparison, these evaluations do not fully reflect our method's primary strength -- its ability to provide directly interpretable and useable predictions of motion distributions in interactive environments, bridging the gap between physical plausibility and efficiency. Furthermore, capabilities such as moving part segmentation directly emerge from our method's design. Overall, these results show our method's strength as a versatile, interpretable, general-purpose motion model. 
We envision this work as a foundation for more probabilistic, generalizable, and actionable approaches to motion understanding, paving the way for deeper insights into complex physical dynamics and future advancements in handling ambiguous and extreme out-of-distribution scenarios (\cref{fig:failure_cases}).

\section*{Acknowledgment}
We would like to thank Mahdi M. Kalayeh for helpful discussions and feedback. We would further like to thank Jannik Wiese, Kim-Louis Barwig, Enrico Shippole, Ming Gui, Thomas Ressler-Antal, Olga Grebenkova, Paul Hofman, Owen Vincent, and the anonymous reviewers. This work was supported in part by a research grant from Netflix. The authors thank Netflix for its support. This project was also supported by the Federal Ministry for Economic Affairs and Energy within the project ``NXT GEN AI METHODS - Generative Methoden für Perzeption, Prädiktion und Planung'', the project ``GeniusRobot'' (01IS24083) funded by the Federal Ministry of Research, Technology and Space (BMFTR), the bidt project KLIMA-MEMES, and the Horizon Europe project ELLIOT (GA No. 101214398). The authors gratefully acknowledge the Gauss Center for Supercomputing for providing compute through the NIC on JUWELS/JUPITER at JSC and the HPC resources supplied by the NHR @FAU Erlangen.

{
    \small
    \bibliographystyle{ieeenat_fullname}
    \bibliography{main}

\begin{thebibliography}{57}
\providecommand{\natexlab}[1]{#1}
\providecommand{\url}[1]{\texttt{#1}}
\expandafter\ifx\csname urlstyle\endcsname\relax
  \providecommand{\doi}[1]{doi: #1}\else
  \providecommand{\doi}{doi: \begingroup \urlstyle{rm}\Url}\fi

\bibitem[Bain et~al.(2021)Bain, Nagrani, Varol, and Zisserman]{bain2021frozenintime}
Max Bain, Arsha Nagrani, G{\"u}l Varol, and Andrew Zisserman.
\newblock Frozen in time: A joint video and image encoder for end-to-end retrieval.
\newblock In \emph{IEEE International Conference on Computer Vision}, 2021.

\bibitem[Blattmann et~al.(2021)Blattmann, Milbich, Dorkenwald, and Ommer]{blattmann2021ipoke}
Andreas Blattmann, Timo Milbich, Michael Dorkenwald, and Bj{\"o}rn Ommer.
\newblock ipoke: Poking a still image for controlled stochastic video synthesis.
\newblock In \emph{Proceedings of the IEEE/CVF International Conference on Computer Vision}, pages 14707--14717, 2021.

\bibitem[Cao et~al.(2021)Cao, Hong, Li, Wang, Xu, Fu, and Xue]{cao2021image}
Chenjie Cao, Yuxin Hong, Xiang Li, Chengrong Wang, Chengming Xu, Yanwei Fu, and Xiangyang Xue.
\newblock The image local autoregressive transformer.
\newblock \emph{Advances in Neural Information Processing Systems}, 34:\penalty0 18433--18445, 2021.

\bibitem[Crowson et~al.(2024)Crowson, Baumann, Birch, Abraham, Kaplan, and Shippole]{crowson2024hourglass}
Katherine Crowson, Stefan~Andreas Baumann, Alex Birch, Tanishq~Mathew Abraham, Daniel~Z Kaplan, and Enrico Shippole.
\newblock Scalable high-resolution pixel-space image synthesis with hourglass diffusion transformers.
\newblock In \emph{Proceedings of the 41st International Conference on Machine Learning}, pages 9550--9575. PMLR, 2024.

\bibitem[Dai et~al.(2023)Dai, Zhang, Yao, Qiu, Zhu, Qin, and Wang]{dai2023motionguidance}
Zuozhuo Dai, Zhenghao Zhang, Yao Yao, Bingxue Qiu, Siyu Zhu, Long Qin, and Weizhi Wang.
\newblock Fine-grained open domain image animation with motion guidance.
\newblock \emph{arXiv preprint arXiv:2311.12886}, 2023.

\bibitem[Darcet et~al.(2023)Darcet, Oquab, Mairal, and Bojanowski]{darcet2023vitneedreg}
Timothée Darcet, Maxime Oquab, Julien Mairal, and Piotr Bojanowski.
\newblock Vision transformers need registers, 2023.

\bibitem[Dosovitskiy et~al.(2021)Dosovitskiy, Beyer, Kolesnikov, Weissenborn, Zhai, Unterthiner, Dehghani, Minderer, Heigold, Gelly, Uszkoreit, and Houlsby]{dosovitskiy2021an}
Alexey Dosovitskiy, Lucas Beyer, Alexander Kolesnikov, Dirk Weissenborn, Xiaohua Zhai, Thomas Unterthiner, Mostafa Dehghani, Matthias Minderer, Georg Heigold, Sylvain Gelly, Jakob Uszkoreit, and Neil Houlsby.
\newblock An image is worth 16x16 words: Transformers for image recognition at scale.
\newblock In \emph{International Conference on Learning Representations}, 2021.

\bibitem[Dubey et~al.(2024)Dubey, Jauhri, Pandey, Kadian, Al-Dahle, Letman, Mathur, Schelten, Yang, Fan, et~al.]{dubey2024llama3}
Abhimanyu Dubey, Abhinav Jauhri, Abhinav Pandey, Abhishek Kadian, Ahmad Al-Dahle, Aiesha Letman, Akhil Mathur, Alan Schelten, Amy Yang, Angela Fan, et~al.
\newblock The llama 3 herd of models.
\newblock \emph{arXiv preprint arXiv:2407.21783}, 2024.

\bibitem[Esser et~al.(2020)Esser, Rombach, and Ommer]{esser2020taming}
Patrick Esser, Robin Rombach, and Bj{\"o}rn Ommer.
\newblock Taming transformers for high-resolution image synthesis. 2021 ieee.
\newblock In \emph{CVF Conference on Computer Vision and Pattern Recognition (CVPR)}, 2020.

\bibitem[Fan et~al.(2024)Fan, Li, Qin, Li, Sun, Rubinstein, Sun, He, and Tian]{fan2024fluid}
Lijie Fan, Tianhong Li, Siyang Qin, Yuanzhen Li, Chen Sun, Michael Rubinstein, Deqing Sun, Kaiming He, and Yonglong Tian.
\newblock Fluid: Scaling autoregressive text-to-image generative models with continuous tokens.
\newblock \emph{arXiv preprint arXiv:2410.13863}, 2024.

\bibitem[Gao et~al.(2018)Gao, Xiong, and Grauman]{gao2018im2flow}
Ruohan Gao, Bo Xiong, and Kristen Grauman.
\newblock Im2flow: Motion hallucination from static images for action recognition.
\newblock In \emph{Proceedings of the IEEE conference on computer vision and pattern recognition}, pages 5937--5947, 2018.

\bibitem[Goldberger and Greenspan(2003)]{goldberger2003efficient}
Goldberger and Greenspan.
\newblock An efficient image similarity measure based on approximations of kl-divergence between two gaussian mixtures.
\newblock In \emph{Proceedings Ninth IEEE International conference on computer vision}, pages 487--493. IEEE, 2003.

\bibitem[Goodfellow et~al.(2014)Goodfellow, Pouget-Abadie, Mirza, Xu, Warde-Farley, Ozair, Courville, and Bengio]{goodfellow2014gan}
Ian Goodfellow, Jean Pouget-Abadie, Mehdi Mirza, Bing Xu, David Warde-Farley, Sherjil Ozair, Aaron Courville, and Yoshua Bengio.
\newblock Generative adversarial nets.
\newblock In \emph{Advances in Neural Information Processing Systems}. Curran Associates, Inc., 2014.

\bibitem[Ho et~al.(2020)Ho, Jain, and Abbeel]{ho2020denoising}
Jonathan Ho, Ajay Jain, and Pieter Abbeel.
\newblock Denoising diffusion probabilistic models.
\newblock \emph{Advances in neural information processing systems}, 33:\penalty0 6840--6851, 2020.

\bibitem[Hu et~al.(2021)Hu, Shen, Wallis, Allen-Zhu, Li, Wang, Wang, and Chen]{hu2021lora}
Edward~J Hu, Yelong Shen, Phillip Wallis, Zeyuan Allen-Zhu, Yuanzhi Li, Shean Wang, Lu Wang, and Weizhu Chen.
\newblock Lora: Low-rank adaptation of large language models.
\newblock \emph{arXiv preprint arXiv:2106.09685}, 2021.

\bibitem[Huang and Belongie(2017)]{huang2017arbitrary}
Xun Huang and Serge Belongie.
\newblock Arbitrary style transfer in real-time with adaptive instance normalization.
\newblock In \emph{Proceedings of the IEEE international conference on computer vision}, pages 1501--1510, 2017.

\bibitem[Jiang et~al.(2021)Jiang, Trulls, Hosang, Tagliasacchi, and Yi]{jiang2021cotr}
Wei Jiang, Eduard Trulls, Jan Hosang, Andrea Tagliasacchi, and Kwang~Moo Yi.
\newblock Cotr: Correspondence transformer for matching across images.
\newblock In \emph{Proceedings of the IEEE/CVF international conference on computer vision}, pages 6207--6217, 2021.

\bibitem[Karaev et~al.(2024)Karaev, Makarov, Wang, Neverova, Vedaldi, and Rupprecht]{karaev2024cotracker3}
Nikita Karaev, Iurii Makarov, Jianyuan Wang, Natalia Neverova, Andrea Vedaldi, and Christian Rupprecht.
\newblock {CoTracker3}: Simpler and better point tracking by pseudo-labelling real videos.
\newblock 2024.

\bibitem[Li et~al.(2024{\natexlab{a}})Li, Zheng, Rupprecht, and Vedaldi]{li2024puppet}
Ruining Li, Chuanxia Zheng, Christian Rupprecht, and Andrea Vedaldi.
\newblock Puppet-master: Scaling interactive video generation as a motion prior for part-level dynamics.
\newblock \emph{arXiv preprint arXiv:2408.04631}, 2024{\natexlab{a}}.

\bibitem[Li et~al.(2025)Li, Zheng, Rupprecht, and Vedaldi]{li2024dragapart}
Ruining Li, Chuanxia Zheng, Christian Rupprecht, and Andrea Vedaldi.
\newblock Dragapart: Learning a part-level motion prior for articulated objects.
\newblock In \emph{Computer Vision -- ECCV 2024}, pages 165--183, Cham, 2025. Springer Nature Switzerland.

\bibitem[Li et~al.(2024{\natexlab{b}})Li, Tian, Li, Deng, and He]{li2024autoregressive}
Tianhong Li, Yonglong Tian, He Li, Mingyang Deng, and Kaiming He.
\newblock Autoregressive image generation without vector quantization.
\newblock \emph{NeurIPS 2024}, 2024{\natexlab{b}}.

\bibitem[Li et~al.(2024{\natexlab{c}})Li, Wang, Zhang, Wang, Yuan, Xie, Zou, and Shan]{li2024imageconductor}
Yaowei Li, Xintao Wang, Zhaoyang Zhang, Zhouxia Wang, Ziyang Yuan, Liangbin Xie, Yuexian Zou, and Ying Shan.
\newblock Image conductor: Precision control for interactive video synthesis.
\newblock \emph{arXiv preprint arXiv:2406.15339}, 2024{\natexlab{c}}.

\bibitem[Li et~al.(2024{\natexlab{d}})Li, Tucker, Snavely, and Holynski]{li2024generative}
Zhengqi Li, Richard Tucker, Noah Snavely, and Aleksander Holynski.
\newblock Generative image dynamics.
\newblock In \emph{Proceedings of the IEEE/CVF Conference on Computer Vision and Pattern Recognition}, pages 24142--24153, 2024{\natexlab{d}}.

\bibitem[Liang et~al.(2024)Liang, Fan, Zhang, Timofte, Van~Gool, and Ranjan]{liang2024movideo}
Jingyun Liang, Yuchen Fan, Kai Zhang, Radu Timofte, Luc Van~Gool, and Rakesh Ranjan.
\newblock Movideo: Motion-aware video generation with diffusion models.
\newblock In \emph{European Conference on Computer Vision}, pages 0000--0000, 2024.

\bibitem[Loshchilov and Hutter(2016)]{loshchilov2016sgdr}
Ilya Loshchilov and Frank Hutter.
\newblock Sgdr: Stochastic gradient descent with warm restarts.
\newblock \emph{arXiv preprint arXiv:1608.03983}, 2016.

\bibitem[Loshchilov and Hutter(2019)]{loshchilov2018decoupled}
Ilya Loshchilov and Frank Hutter.
\newblock Decoupled weight decay regularization.
\newblock In \emph{International Conference on Learning Representations}, 2019.

\bibitem[Nan et~al.(2024)Nan, Xie, Zhou, Fan, Yang, Chen, Li, Yang, and Tai]{nan2024openvid}
Kepan Nan, Rui Xie, Penghao Zhou, Tiehan Fan, Zhenheng Yang, Zhijie Chen, Xiang Li, Jian Yang, and Ying Tai.
\newblock Openvid-1m: A large-scale high-quality dataset for text-to-video generation.
\newblock \emph{arXiv preprint arXiv:2407.02371}, 2024.

\bibitem[OpenAI(2024)]{SoraOpenAI}
OpenAI.
\newblock Sora, 2024.

\bibitem[Oquab et~al.(2023)Oquab, Darcet, Moutakanni, Vo, Szafraniec, Khalidov, Fernandez, Haziza, Massa, El-Nouby, Howes, Huang, Xu, Sharma, Li, Galuba, Rabbat, Assran, Ballas, Synnaeve, Misra, Jegou, Mairal, Labatut, Joulin, and Bojanowski]{oquab2023dinov2}
Maxime Oquab, Timothée Darcet, Theo Moutakanni, Huy~V. Vo, Marc Szafraniec, Vasil Khalidov, Pierre Fernandez, Daniel Haziza, Francisco Massa, Alaaeldin El-Nouby, Russell Howes, Po-Yao Huang, Hu Xu, Vasu Sharma, Shang-Wen Li, Wojciech Galuba, Mike Rabbat, Mido Assran, Nicolas Ballas, Gabriel Synnaeve, Ishan Misra, Herve Jegou, Julien Mairal, Patrick Labatut, Armand Joulin, and Piotr Bojanowski.
\newblock Dinov2: Learning robust visual features without supervision, 2023.

\bibitem[Palmer(1999)]{Palmer_99}
Stephen~E. Palmer.
\newblock \emph{Vision Science: {Photons} to Phenomenology}.
\newblock The MIT Press, 1999.

\bibitem[Pan et~al.(2023)Pan, Tewari, Leimk{\"u}hler, Liu, Meka, and Theobalt]{pan2023draggan}
Xingang Pan, Ayush Tewari, Thomas Leimk{\"u}hler, Lingjie Liu, Abhimitra Meka, and Christian Theobalt.
\newblock Drag your gan: Interactive point-based manipulation on the generative image manifold.
\newblock In \emph{ACM SIGGRAPH 2023 Conference Proceedings}, pages 1--11, 2023.

\bibitem[Podell et~al.(2024)Podell, English, Lacey, Blattmann, Dockhorn, M{\"u}ller, Penna, and Rombach]{podell2024sdxl}
Dustin Podell, Zion English, Kyle Lacey, Andreas Blattmann, Tim Dockhorn, Jonas M{\"u}ller, Joe Penna, and Robin Rombach.
\newblock Sdxl: Improving latent diffusion models for high-resolution image synthesis.
\newblock In \emph{The Twelfth International Conference on Learning Representations}, 2024.

\bibitem[Rosello(2016)]{rosello2016predicting}
Pol Rosello.
\newblock Predicting future optical flow from static video frames.
\newblock \emph{Retrieved on: Jul}, 18:\penalty0 2, 2016.

\bibitem[Shazeer(2020)]{shazeer2020glu}
Noam Shazeer.
\newblock Glu variants improve transformer.
\newblock \emph{arXiv preprint arXiv:2002.05202}, 2020.

\bibitem[Shi et~al.(2024)Shi, Huang, Wang, Bian, Li, Zhang, Zhang, Cheung, See, Qin, et~al.]{shi2024motion}
Xiaoyu Shi, Zhaoyang Huang, Fu-Yun Wang, Weikang Bian, Dasong Li, Yi Zhang, Manyuan Zhang, Ka~Chun Cheung, Simon See, Hongwei Qin, et~al.
\newblock Motion-i2v: Consistent and controllable image-to-video generation with explicit motion modeling.
\newblock \emph{SIGGRAPH 2024}, 2024.

\bibitem[Shin et~al.(2024)Shin, Choi, and Park]{shin2024instantdrag}
Joonghyuk Shin, Daehyeon Choi, and Jaesik Park.
\newblock Instantdrag: Improving interactivity in drag-based image editing.
\newblock \emph{arXiv preprint arXiv:2409.08857}, 2024.

\bibitem[Song et~al.(2021)Song, Sohl-Dickstein, Kingma, Kumar, Ermon, and Poole]{song2021scorebased}
Yang Song, Jascha Sohl-Dickstein, Diederik~P Kingma, Abhishek Kumar, Stefano Ermon, and Ben Poole.
\newblock Score-based generative modeling through stochastic differential equations.
\newblock In \emph{International Conference on Learning Representations}, 2021.

\bibitem[Su et~al.(2021)Su, Lu, Pan, Wen, and Liu]{su2021roformer}
Jianlin Su, Yu Lu, Shengfeng Pan, Bo Wen, and Yunfeng Liu.
\newblock Roformer: enhanced transformer with rotary position embedding. corr abs/2104.09864 (2021).
\newblock \emph{arXiv preprint arXiv:2104.09864}, 2021.

\bibitem[Sutskever et~al.(2014)Sutskever, Vinyals, and Le]{sutskever2014seq2seq}
Ilya Sutskever, Oriol Vinyals, and Quoc~V Le.
\newblock Sequence to sequence learning with neural networks.
\newblock In \emph{Advances in Neural Information Processing Systems}. Curran Associates, Inc., 2014.

\bibitem[Teed and Deng(2020)]{teed2020raft}
Zachary Teed and Jia Deng.
\newblock Raft: Recurrent all-pairs field transforms for optical flow.
\newblock In \emph{Computer Vision--ECCV 2020: 16th European Conference, Glasgow, UK, August 23--28, 2020, Proceedings, Part II 16}, pages 402--419. Springer, 2020.

\bibitem[Tschannen et~al.(2024)Tschannen, Eastwood, and Mentzer]{tschannen2024givt}
Michael Tschannen, Cian Eastwood, and Fabian Mentzer.
\newblock Givt: Generative infinite-vocabulary transformers.
\newblock In \emph{European Conference on Computer Vision}, pages 292--309. Springer, 2024.

\bibitem[Vaswani(2017)]{vaswani2017attention}
A Vaswani.
\newblock Attention is all you need.
\newblock \emph{Advances in Neural Information Processing Systems}, 2017.

\bibitem[Walker et~al.(2015)Walker, Gupta, and Hebert]{walker2015dense}
Jacob Walker, Abhinav Gupta, and Martial Hebert.
\newblock Dense optical flow prediction from a static image.
\newblock In \emph{Proceedings of the IEEE international conference on computer vision}, pages 2443--2451, 2015.

\bibitem[Wang et~al.(2021)Wang, Mallya, and Liu]{wang2021facevid2vid}
Ting-Chun Wang, Arun Mallya, and Ming-Yu Liu.
\newblock One-shot free-view neural talking-head synthesis for video conferencing.
\newblock In \emph{CVPR}, 2021.

\bibitem[Wang et~al.(2024)Wang, Yuan, Wang, Li, Chen, Xia, Luo, and Shan]{wang2024motionctrl}
Zhouxia Wang, Ziyang Yuan, Xintao Wang, Yaowei Li, Tianshui Chen, Menghan Xia, Ping Luo, and Ying Shan.
\newblock Motionctrl: A unified and flexible motion controller for video generation.
\newblock In \emph{ACM SIGGRAPH 2024 Conference Papers}, pages 1--11, 2024.

\bibitem[Wu et~al.(2025)Wu, Li, Gu, Zhao, He, Zhang, Shou, Li, Gao, and Zhang]{wu2025draganything}
Weijia Wu, Zhuang Li, Yuchao Gu, Rui Zhao, Yefei He, David~Junhao Zhang, Mike~Zheng Shou, Yan Li, Tingting Gao, and Di Zhang.
\newblock Draganything: Motion control for anything using entity representation.
\newblock In \emph{European Conference on Computer Vision}, pages 331--348. Springer, 2025.

\bibitem[Xiao et~al.(2025)Xiao, Wang, Xue, Karaev, Makarov, Kang, Zhu, Bao, Shen, and Zhou]{xiao2025spatialtracker}
Yuxi Xiao, Jianyuan Wang, Nan Xue, Nikita Karaev, Iurii Makarov, Bingyi Kang, Xin Zhu, Hujun Bao, Yujun Shen, and Xiaowei Zhou.
\newblock Spatialtrackerv2: 3d point tracking made easy.
\newblock In \emph{ICCV}, 2025.

\bibitem[Xiao~et al.(2024)]{xiao2024spatialtracker}
Yuxi Xiao~et al.
\newblock Spatialtracker: Tracking any 2d pixels in 3d space.
\newblock In \emph{CVPR}, 2024.

\bibitem[Xie et~al.(2024)Xie, Zong, Qiu, Li, Feng, Yang, and Jiang]{xie2024physgaussian}
Tianyi Xie, Zeshun Zong, Yuxing Qiu, Xuan Li, Yutao Feng, Yin Yang, and Chenfanfu Jiang.
\newblock Physgaussian: Physics-integrated 3d gaussians for generative dynamics.
\newblock In \emph{Proceedings of the IEEE/CVF Conference on Computer Vision and Pattern Recognition}, pages 4389--4398, 2024.

\bibitem[Yin et~al.(2023)Yin, Wu, Liang, Shi, Li, Ming, and Duan]{yin2023dragnuwa}
Shengming Yin, Chenfei Wu, Jian Liang, Jie Shi, Houqiang Li, Gong Ming, and Nan Duan.
\newblock Dragnuwa: Fine-grained control in video generation by integrating text, image, and trajectory.
\newblock \emph{arXiv preprint arXiv:2308.08089}, 2023.

\bibitem[Yu et~al.(2022)Yu, Xu, Koh, Luong, Baid, Wang, Vasudevan, Ku, Yang, Ayan, Hutchinson, Han, Parekh, Li, Zhang, Baldridge, and Wu]{yu2022parti}
Jiahui Yu, Yuanzhong Xu, Jing~Yu Koh, Thang Luong, Gunjan Baid, Zirui Wang, Vijay Vasudevan, Alexander Ku, Yinfei Yang, Burcu~Karagol Ayan, Ben Hutchinson, Wei Han, Zarana Parekh, Xin Li, Han Zhang, Jason Baldridge, and Yonghui Wu.
\newblock Scaling autoregressive models for content-rich text-to-image generation.
\newblock \emph{Transactions on Machine Learning Research}, 2022.
\newblock Featured Certification.

\bibitem[Yu et~al.(2023)Yu, Zhu, Jiang, Loy, Cai, and Wu]{yu2023celebv}
Jianhui Yu, Hao Zhu, Liming Jiang, Chen~Change Loy, Weidong Cai, and Wayne Wu.
\newblock Celebv-text: A large-scale facial text-video dataset.
\newblock In \emph{Proceedings of the IEEE/CVF Conference on Computer Vision and Pattern Recognition}, pages 14805--14814, 2023.

\bibitem[Zhang and Sennrich(2019)]{zhang2019root}
Biao Zhang and Rico Sennrich.
\newblock Root mean square layer normalization.
\newblock \emph{Advances in Neural Information Processing Systems}, 32, 2019.

\bibitem[Zhang et~al.(2023)Zhang, Rao, and Agrawala]{zhang2023control}
Lvmin Zhang, Anyi Rao, and Maneesh Agrawala.
\newblock Adding conditional control to text-to-image diffusion models.
\newblock In \emph{Proceedings of the IEEE/CVF International Conference on Computer Vision}, pages 3836--3847, 2023.

\bibitem[Zhang et~al.(2018)Zhang, Isola, Efros, Shechtman, and Wang]{zhang2018unreasonable}
Richard Zhang, Phillip Isola, Alexei~A Efros, Eli Shechtman, and Oliver Wang.
\newblock The unreasonable effectiveness of deep features as a perceptual metric.
\newblock In \emph{Proceedings of the IEEE conference on computer vision and pattern recognition}, pages 586--595, 2018.

\bibitem[Zhang et~al.(2025)Zhang, Yu, Wu, Feng, Zheng, Snavely, Wu, and Freeman]{zhang2025physdreamer}
Tianyuan Zhang, Hong-Xing Yu, Rundi Wu, Brandon~Y Feng, Changxi Zheng, Noah Snavely, Jiajun Wu, and William~T Freeman.
\newblock Physdreamer: Physics-based interaction with 3d objects via video generation.
\newblock In \emph{European Conference on Computer Vision}, pages 388--406. Springer, 2025.

\bibitem[Zholus et~al.(2025)Zholus, Doersch, Yang, Koppula, Patraucean, He, Rocco, Sajjadi, Chandar, and Goroshin]{zholus2025tapnext}
Artem Zholus, Carl Doersch, Yi Yang, Skanda Koppula, Viorica Patraucean, Xu~Owen He, Ignacio Rocco, Mehdi~SM Sajjadi, Sarath Chandar, and Ross Goroshin.
\newblock Tapnext: Tracking any point (tap) as next token prediction.
\newblock \emph{arXiv preprint arXiv:2504.05579}, 2025.

\end{thebibliography}
}

\clearpage

\setcounter{figure}{0}
\setcounter{table}{0}
\setcounter{equation}{0}
\setcounter{section}{0}
\renewcommand\thefigure{\Alph{figure}}  
\renewcommand\thetable{\Alph{table}}  
\renewcommand\thesection{\Alph{section}}

\section{Implementation Details}\label{sec:app_imp_details}
The hyperparameters for the base model used for all evaluation and qualitative examples are reported in \cref{tab:imp_details}. We train the WebVid model for a total of 800k steps with a learning rate of $5.0 \times 10^{-5}$ using the AdamW~\cite{loshchilov2018decoupled} optimizer and a linear warmup of 5000 steps. The first 250k steps are trained with a batch size of 32; for the remainder, we set the batch size to 128. For our second model, keep the batch size constant and significantly increase warmup time. We also add a cosine LR decay and increase training time to 1M steps. Training time differences are due to an improved trainer setup from the first to the second model.

As described in \suppvspreprint{Sec. 4.1}{\cref{sec:imp_details}}, we randomly sample flow pokes and their corresponding positions from the given trajectories. We enforce that all flow values are in $[-1, 1]$ by applying a \texttt{tanh} mapping and obtain a sinusoidal embedding for the $x$ and $y$ components of the flow. We then find the corresponding image features using the pokes positions and concatenate them with the flow features and local image features extracted by the image feature extractor $\mathcal{E}(\mathcal{I})$. Finally, we project them using a mapping network. These embeddings are then combined with query tokens. The query tokens represent the locations in the image for which we want to predict a flow distribution and are realized by a learnable token and the corresponding positional encoding.

\begin{figure}[t]
    \includegraphics[width=\linewidth]{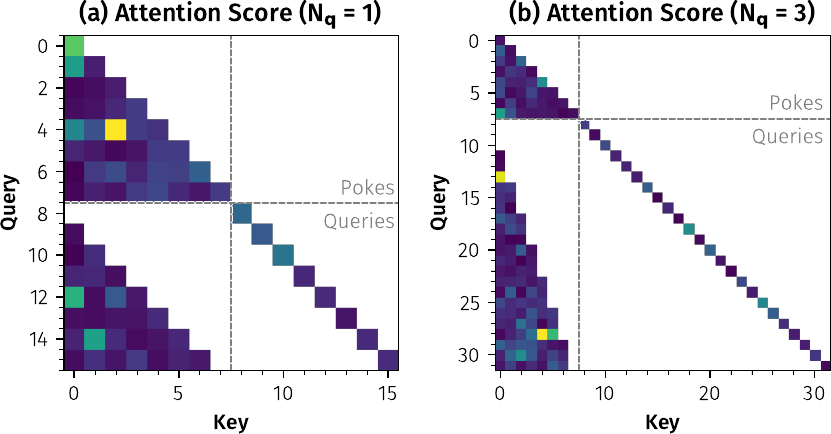}
    \caption{\textbf{Query-Causal Attention Pattern Visualization}. We show the resulting attention patterns for our query-causal attention for different numbers of queries per poke count. We put poke tokens first, followed by query tokens. \textbf{(a)} In the simplest setting, with one query per set of pokes, there is one query token per set of pokes, with each query token attending to one more poke token than its predecessor. \textbf{(b)} For $N_q > 1$, the poke attention does not change, but there are multiple query tokens per poke set. Here, even query tokens for the same poke set do not attend to each other to enable parallel evaluation during inference.}
    \label{fig:query_causal_attention_pattern}
\end{figure}

The flow pokes and query tokens are fed to the transformer, which is 12 blocks deep and has a width of 768. The self-attention uses our \textit{query-causal attention} mask as introduced in \suppvspreprint{Sec. 3.2}{\cref{par:training_object}}. We visualize it in \cref{fig:query_causal_attention_pattern}. In the cross-attention, the pokes and queries attend to the image features, enabling them to learn a global understanding of the scene. In both attention mechanisms, we use 2D Axial RoPE~\cite{su2021roformer,crowson2024hourglass} to model the spatial relationships of tokens. The FNNs expand the internal feature dimension by a factor of three, use SwiGLU~\cite{shazeer2020glu} as an activation function, and are conditioned on whether or not the camera is static using AdaRMSNorms. On the output, we observe that directly using the output of the transformer for GMM parameter prediction sometimes produces unreasonable distributions and thus bad performance. Therefore, we use a simple MLP to project the transformer's output, alleviating that problem.
For further details, we refer to our reference implementation\footnote{\href{https://github.com/CompVis/flow-poke-transformer}{https://github.com/CompVis/flow-poke-transformer}}, which contains extensive further comments in context.

Whether the camera is static or not is detected using a simple heuristic: we consider it to be static if a significant fraction of the scene's content is static. This information can be directly derived from the training tracks. Specifically, we find that considering a camera static once 40\% of the frame move by at most 3px (at our training resolution of $448^2$) works well on our training data. For the second training, we use 1\% of the frame side length as the threshold instead.

\begin{table*}[ht]
    \centering
    \adjustbox{max width=\linewidth}{
    \begin{tabular}{lccHcHcHcHcH}
        \toprule
        {Method} & {\shortstack{Trained On}} & {EPE@1 $\downarrow$} & & {EPE@2 $\downarrow$} & & {EPE@5 $\downarrow$} & & {EPE@10 $\downarrow$} & & {EPE@100 $\downarrow$} \\
        \midrule
        {InstantDrag~\cite{shin2024instantdrag} } & Faces & \underline{9.24} & \textbf{0.193} & \underline{9.12} & \textbf{0.196} & \underline{8.82} & \textbf{0.197} & \underline{8.39} & 0.198 & 7.29 & 0.212 \\
        Motion-I2V~\cite{shi2024motion} & \textbf{Generic (Zero-Shot)} & 29.08 & 0.029 & 27.40 & 0.031 & 24.22 & 0.030 & 20.90 & 0.048 & n/a\  & n/a\  \\
        \textbf{Ours} (full training)  & \textbf{Generic (Zero-Shot)} & \textbf{7.64} & \underline{0.150} & \textbf{6.87} & \underline{0.154} & \textbf{5.32} & \underline{0.167} & \textbf{4.20} & \underline{0.183} & \textbf{2.51} & \textbf{0.264} \\
        \midrule
        \midrule
        \multicolumn{3}{l}{\textit{Vision Feature Extractor Initialization}}\\
        Jointly Trained Pre-Trained (\textbf{Ours}) & \textbf{Generic (Zero-Shot)}     & \textbf{8.08} & 0.129 & \textbf{6.96} & 0.135 & \textbf{5.38} & 0.158 & \underline{3.99} & 0.177 & \underline{2.33} & 0.255 \\
        Frozen Pre-Trained & \textbf{Generic (Zero-Shot)}                              & {8.30} & 0.191 & \underline{7.22} & 0.194 & \underline{5.44} & 0.208 & \textbf{3.78} & 0.233 & \textbf{2.22} & 0.334 \\
        Trained from Scratch & \textbf{Generic (Zero-Shot)}                            & \underline{8.15} & 0.148 & 7.51 & 0.153 & 6.14 & 0.158 & 4.73 & 0.188 & 2.57 & 0.266 \\
        \midrule
        \multicolumn{3}{l}{\textit{GMM Component Count}}\\
        1 Component & \textbf{Generic (Zero-Shot)}                                     & \textbf{7.60} & 0.084 & \textbf{6.87} & 0.084 & \underline{5.42} & 0.101 & 4.18 & 0.109 & 2.59 & 0.150 \\
        2 Components & \textbf{Generic (Zero-Shot)}                                    & 8.98 & 0.135 & 7.87 & 0.136 & 5.83 & 0.155 & 4.08 & 0.174 & 2.34 & 0.261 \\
        4 Components (\textbf{Ours}) & \textbf{Generic (Zero-Shot)}                    & \underline{8.08} & 0.129 & \underline{6.96} & 0.135 & \textbf{5.38} & 0.158 & \underline{3.99} & 0.177 & \underline{2.33} & 0.255 \\
        8 Components & \textbf{Generic (Zero-Shot)}                                    & 8.23 & \underline{0.154} & 7.19 & \underline{0.157} & 5.57 & \underline{0.171} & 4.01 & \underline{0.192} & \textbf{2.29} & \underline{0.278} \\
        16 Components & \textbf{Generic (Zero-Shot)}                                   & 8.41 & 0.128 & 7.26 & 0.128 & 5.44 & 0.144 & \textbf{3.90} & 0.165 & 2.34 & 0.248 \\
        \midrule
        \multicolumn{3}{l}{\textit{GMM Covariance Parametrization}}\\
        Full Covariance, 4 Components (\textbf{Ours}) & \textbf{Generic (Zero-Shot)}   & \textbf{8.08} & \underline{0.129} & \textbf{6.96} & \underline{0.135} & \textbf{5.38} & \underline{0.158} & \underline{3.99} & \underline{0.177} & \underline{2.33} & \underline{0.255} \\
        Diagonal, 4 Components & \textbf{Generic (Zero-Shot)}                          & \underline{8.13} & \textbf{0.166} & \underline{7.09} & \textbf{0.168} & \underline{5.40} & \textbf{0.187} & \textbf{3.98} & \textbf{0.213} & \textbf{2.24} & \textbf{0.304} \\
        \bottomrule
    \end{tabular}
    }
    \vspace{-2mm}
    \caption{Extension of \suppvspreprint{Tab. 1}{\cref{tab:face_flow_quantitative}} including our ablations. The experiment is identical to the original one. The ablation models have been trained for 250k steps compared to 800k for the full training. EPE@N refers to the endpoint error given N pokes.}
    \label{tab:ablations_quantitative}
    \vspace{-3mm}
\end{table*}

\begin{table}[H]
    \centering
    \adjustbox{max width=\columnwidth}{
    \begin{tabular}{l@{}cc}
        \toprule
        Parameter & \multicolumn{2}{c}{Value} \\
        \midrule
        Dataset & WebVid~\cite{bain2021frozenintime} Subset & Open-Set Videos \\
        Number of clips & 3.8M & 5M \\
        \midrule
        Batch size & 32$\rightarrow$128 & 128  \\
        Optimizer & AdamW~\cite{loshchilov2018decoupled} & AdamW~\cite{loshchilov2018decoupled} \\
        Peak learning rate & $5.0 \times 10^{-5}$ & $5.0 \times 10^{-5}$ \\
        Learning rate schedule & constant & cosine decay to $10^{-8}$ \\
        Betas & $(0.9, 0.99)$ & $(0.9, 0.99)$ \\
        Warm-up steps & 5k & 100k \\
        Total Steps & 800k & 1M \\
        Precision & bfloat16 & bfloat16 \\
        Total Parameters & 230M & 230M \\
        \midrule
        GPUs & 2 Nvidia H200 & 8 Nvidia H200 \\
        Training Time & 7 days & 1 day \\
        \midrule
        Tracker & CoTracker3~\cite{karaev2024cotracker3} & TAPNext~\cite{zholus2025tapnext} \\
        Tracker position seeding & $48\times48$ grid & 1024 random \\
        Flow scale & $[-1, 1]$ & $[-1, 1]$ \\
        Image size & $448 \times 448$ & $448 \times 448$ \\
        Mixtures & 4 & 4 \\
        Covariance & Full & Full \\
        Given pokes & 128 & 128 \\
        Query factor & 15 & 15 \\
        \midrule
        Depth & 12 & 12 \\
        SA width & 768 & 768 \\
        CA width & 768 & 768 \\
        Normalization & RMSNorm~\cite{zhang2019root} & RMSNorm~\cite{zhang2019root} \\
        FFN expand factor & 3 & 3\\
        Activation & SwiGLU~\cite{shazeer2020glu} & SwiGLU~\cite{shazeer2020glu} \\
        Positional encoding & 2D Axial RoPE ~\cite{su2021roformer,crowson2024hourglass} & 2D Axial RoPE ~\cite{su2021roformer,crowson2024hourglass} \\
        Static scene conditioning & Adaptive Norm~\cite{huang2017arbitrary} & Adaptive Norm~\cite{huang2017arbitrary} \\
        \bottomrule
    \end{tabular}
    }
    \caption{Hyperparameters for our main model across both datasets. Ablation models use the same parameters as the first one, but only train for 250k steps.}
    \label{tab:imp_details}
    \vspace{2cm}
\end{table}

\section{Ablations}\label{sec:ablations}
We show an extended version of \suppvspreprint{Tab. 1}{\cref{tab:face_flow_quantitative}} that includes quantitative results for hyperparameter ablations, which were used to motivate the choices mentioned in the main paper, in \cref{tab:ablations_quantitative}. All comparison models follow the original training recipe but are only trained until 250k steps due to compute constraints.

\begin{figure}
    \centering
        \begin{tabular}{c@{\hskip .5em}c@{\hskip .5em}c}
            \footnotesize Input & \footnotesize \makecell[c]{Prediction\\(Frozen Vis. Enc.)} & \footnotesize \makecell[c]{Prediction\\(Jointly Trained)}\\
            \includegraphics[width=.3\columnwidth]{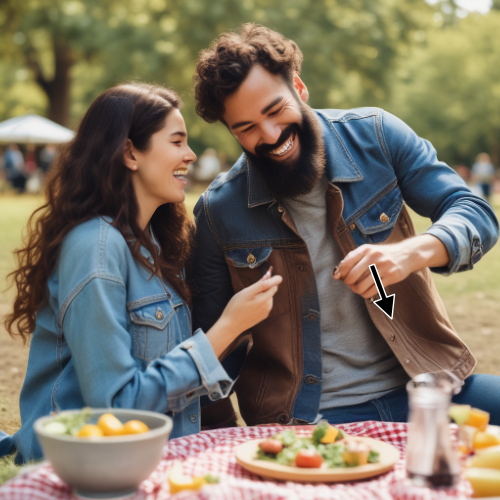} &
            \includegraphics[width=.3\columnwidth]{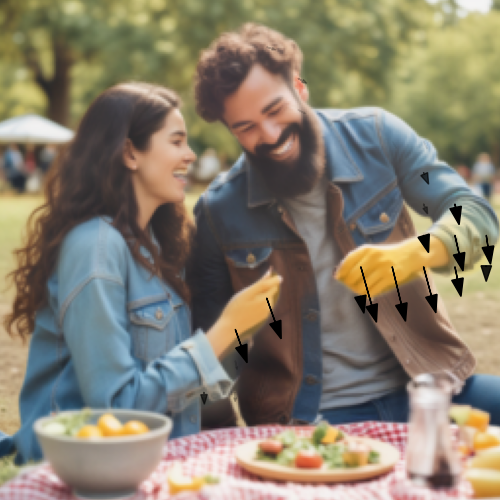} &
            \includegraphics[width=.3\columnwidth]{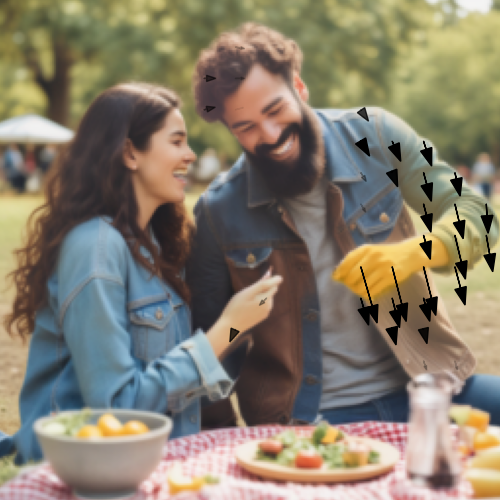}\\
        \end{tabular}
    \caption{Jointly training the vision encoder is important. We train a model with a frozen pretrained vision encoder. The model struggles with instance-specificity, predicting the same movement for the woman's hand as it does for the man's hand. When jointly training the vision encoder with the flow poke transformer, the man's hand's movement does not directly influence the woman's hand.}
    \label{fig:vision_feature_extractor_finetuning}
\end{figure}

First, we ablate whether to initialize the vision encoder with pre-trained weights and whether to continue training them. We find that initializing with pre-trained weights gives a performance boost in this setting. We hypothesize that, for very long trainings, both versions might end up performing similarly well. As noted in \suppvspreprint{Sec. 4.1}{\cref{sec:imp_details}}, freezing the feature extractor when initializing with DINOv2~\cite{oquab2023dinov2} empirically results in a model with reduced instance segmentation capabilities compared to the unlocked version. We show a qualitative example of this in \cref{fig:vision_feature_extractor_finetuning}. This behavior can also be observed in the quantitative evaluations, where, for low poke counts, the jointly trained model performs better than the one with a frozen feature extractor. At high poke counts, instance segmentation capabilities likely become less relevant, as the movement of all instances is likely already given explicitly via the conditioning.

When ablating the number of GMM components, we find that adding too many components reduces the model's performance when conditioned on low numbers of given pokes. Only predicting a single component results in better quantitative performance at low given poke counts, but, obviously prevents the model from predicting multimodal distributions (see, e.g., \suppvspreprint{Figures 1, 3, and \ref{fig:app_distribution}}{\cref{fig:teaser,fig:distributions_qualitative,fig:app_distribution}}), omitting a central property of our model. Parametrizing the covariance matrices as a pure diagonal matrix as done in GIVT~\cite{tschannen2024givt} results in slightly reduced performance on average. Qualitatively, it also prevents angled distributions that our model successfully uses to express directional uncertainty (see, e.g., \suppvspreprint{Figure 1}{\cref{fig:teaser}}).

\section{Extension to 3D Motion}
\label{sec:3d-extension}
\begin{figure}[t]
    \centering
    \begin{tabular}{c@{\hskip .5em}c@{\hskip .5em}c@{\hskip .5em}c}
        \footnotesize Input & \footnotesize \makecell[c]{Prediction $\mathbf{F}|_{xy}$} & \footnotesize \makecell[c]{Prediction $\mathbf{F}|_{z}$}\\
        \includegraphics[width=.25\columnwidth]{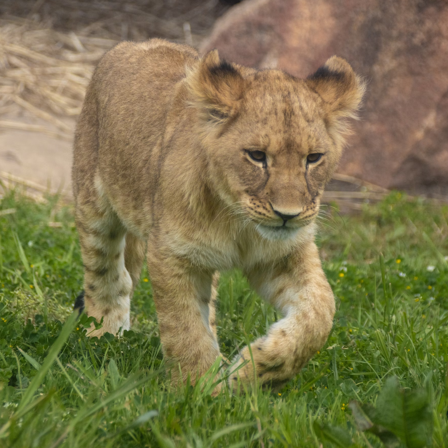} &
        \includegraphics[width=.25\columnwidth]{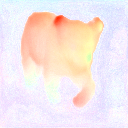} &
        \includegraphics[width=.25\columnwidth]{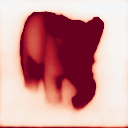} &
        \multirow{2}{*}[32pt]{\includegraphics[width=.1\columnwidth]{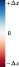}}\\
        \includegraphics[width=.25\columnwidth]{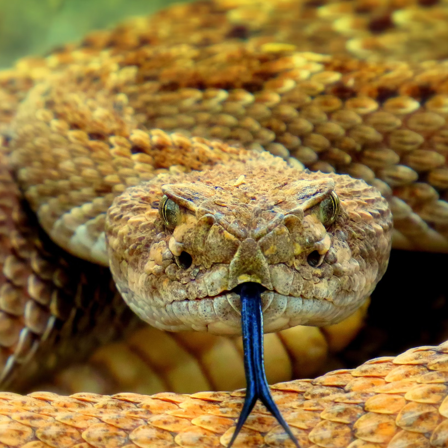} &
        \includegraphics[width=.25\columnwidth]{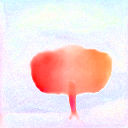} &
        \includegraphics[width=.25\columnwidth]{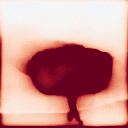}\\
    \end{tabular}
    \caption{\textbf{3D Motion Estimation}. We show unconditional 3D motion estimation samples from an FPT variant fine-tuned on 3D track data. The in-plane motion prediction $\mathbf{F}|_{xy}$ resembles that of a 2D FPT model, while this version can also successfully predict plausible out-of-plane motion $\mathbf{F}|_{z}$.
    }
    \label{fig:3d_samples}
\end{figure}

In this paper, we evaluated the Flow Poke Transformer in the two-dimensional setting, meaning that the model only reasons in the image plane. However, the architecture itself is not limited to this setting and can trivially be extended to higher dimensions if desired. We show qualitative motion prediction results from such a version in \cref{fig:3d_samples}, where the model also successfully predicts reasonable out-of-plane motion in full 3D. This model was obtained by continued training from a 2D FPT checkpoint with 3D trackers obtained using SpatialTrackerV2~\cite{xiao2025spatialtracker,xiao2024spatialtracker} on a subset of OpenVid1M~\cite{nan2024openvid}.
The model is capable of predicting movement towards and away from the camera without any pokes, as shown by the near-static background and the clearly segmented motion of the animals in all three dimensions. When combining the predicted flow in Z-direction with a depth estimation of the input image, it is possible to tell which parts in the image will be occluded in the future.

\section{Additional Qualitative Examples}
\begin{figure}[b]
    \centering
    \adjustbox{max width=\columnwidth}{
        \begin{minipage}{1.2\linewidth}\hspace{-1mm}
            \begin{tabularx}{\columnwidth}{@{}X@{}X@{}X@{}X@{}X@{}X@{}}
                 \makecell[c]{\footnotesize Input} & \makecell[c]{\footnotesize Motion-I2V} & \makecell[c]{\footnotesize DragAPart} & \makecell[c]{\footnotesize PuppetMaster} & \makecell[c]{\footnotesize Ours} & \makecell[c]{\footnotesize Ground Truth}
            \end{tabularx}
        \end{minipage}
    }
    \includegraphics[width=\columnwidth]{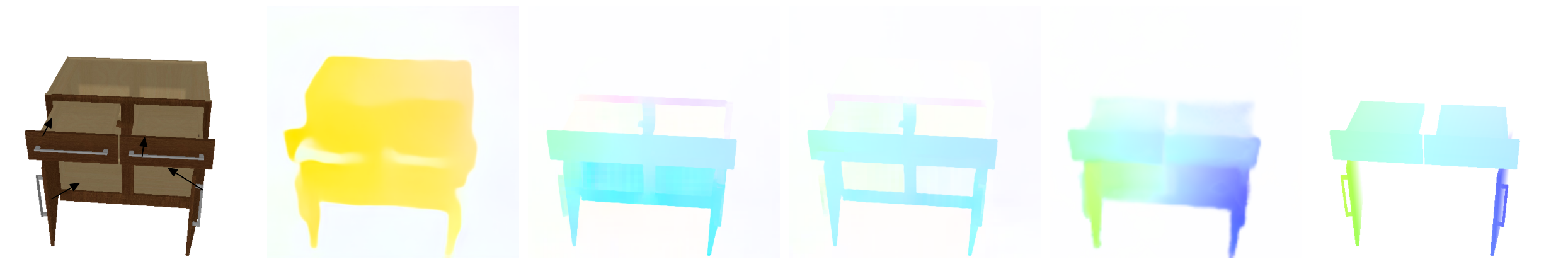}\\[-0.1em]
    \includegraphics[width=\columnwidth]{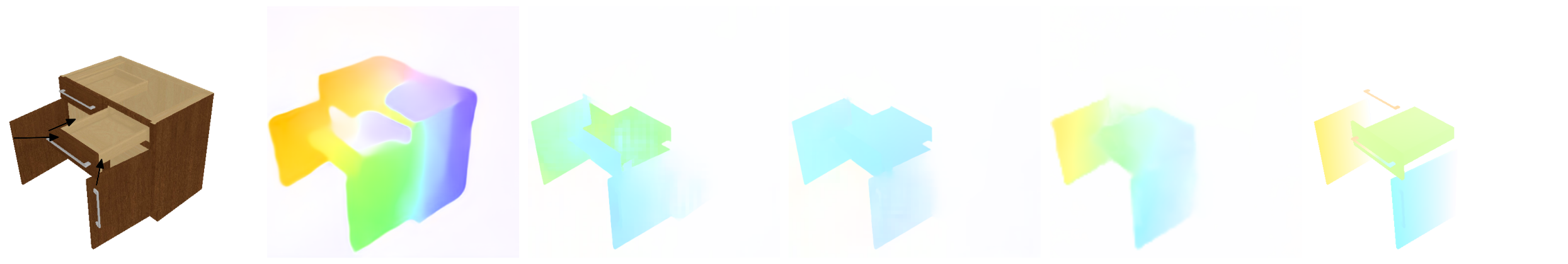}\\[-0.1em]
    \includegraphics[width=\columnwidth]{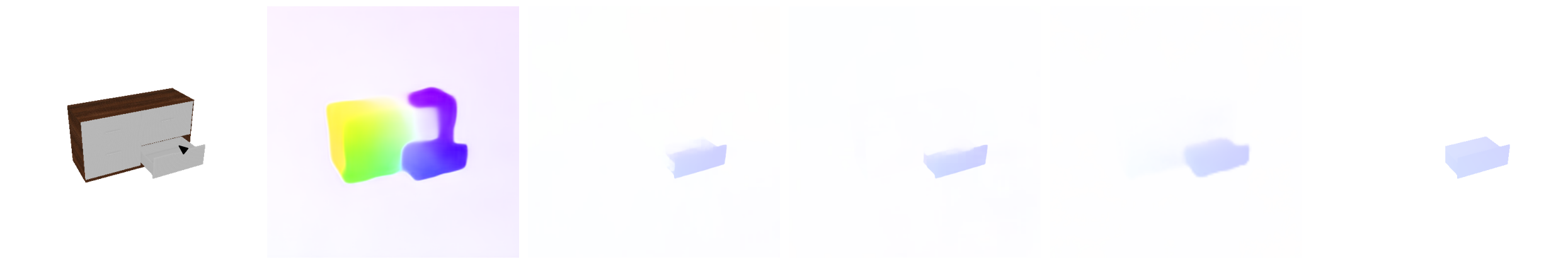}\\[-0.1em]
    \includegraphics[width=\columnwidth]{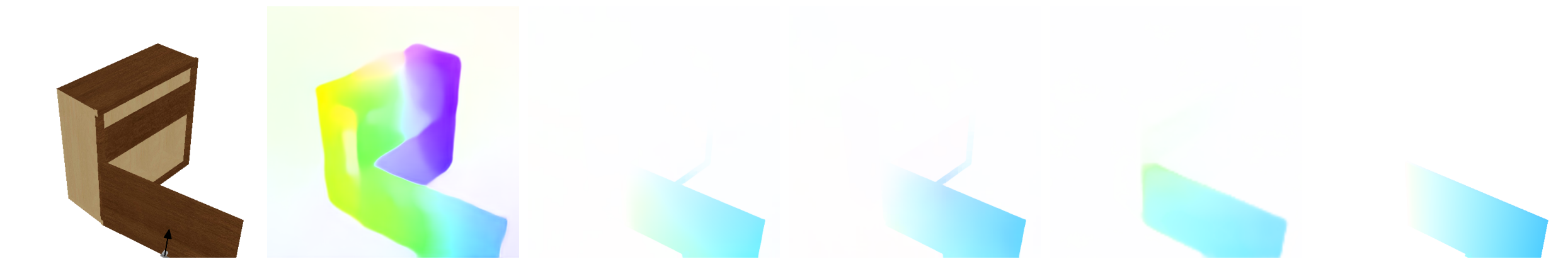}
    \caption{\textbf{Qualitative Results for Articulated Object Motion Estimation}. We compare on Drag-A-Move~\cite{li2024dragapart} with Motion-I2V~\cite{shi2024motion}, DragAPart~\cite{li2024dragapart}, and PuppetMaster~\cite{li2024puppet}. Our model is qualitatively more capable of capturing complex conditioning with multiple different pokes than DAP and PM in this setup. Motion-I2V often fails to accurately follow the conditioning locally.}
    \label{fig:dragamove_qualitative}
\end{figure}

For additional context for our quantitative results in \cref{tab:articulated_objects_quantitative}, we show visualizations of some samples from that experiment in \cref{fig:dragamove_qualitative}.
We also show additional qualitative examples for and pointwise motion predictions in \cref{fig:app_distribution} and samples for dense motion estimation in \cref{fig:app_dense}.

{
\setlength{\tabcolsep}{0.007\linewidth}
\begin{figure*}[htb]
    \centering
    \begin{subfigure}[b]{\linewidth}
    \centering
    \begin{tabular}{cc cc cc}
        {\footnotesize Input Image} & {\footnotesize Overlayed Distribution} &{\footnotesize Zoom Distribution} &{\footnotesize Input Image} &{\footnotesize Overlayed Distribution} &{\footnotesize Zoom Distribution} \\
        
        \includegraphics[width=0.15\linewidth]{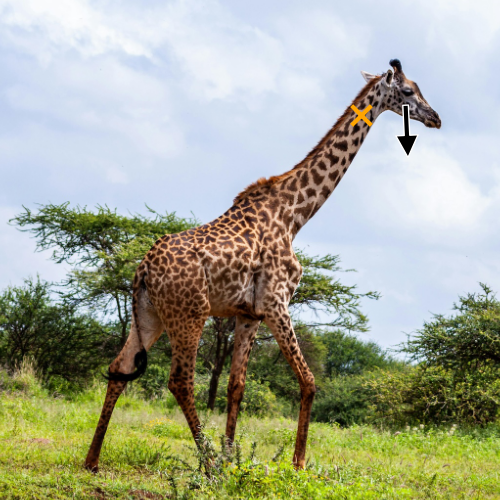} &
        \includegraphics[width=0.15\linewidth]{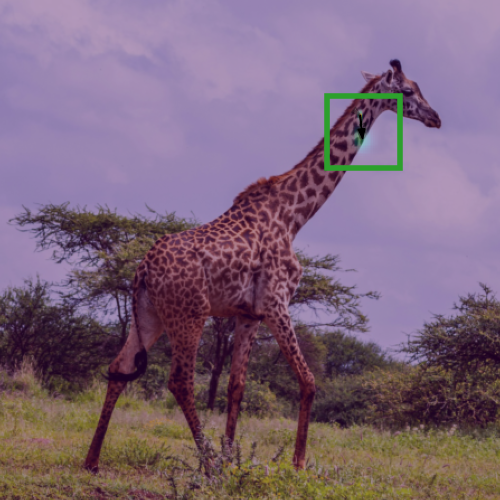} &
        \includegraphics[width=0.15\linewidth]{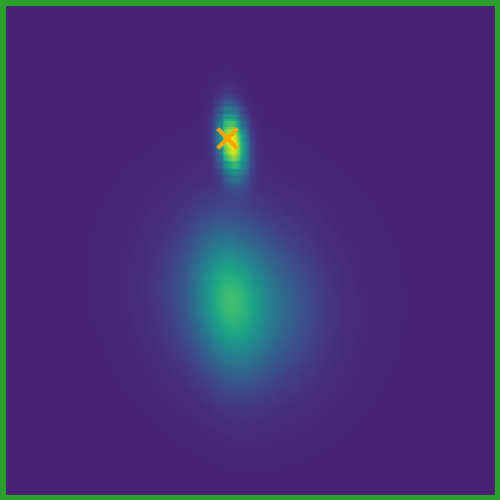} &
        \includegraphics[width=0.15\linewidth]{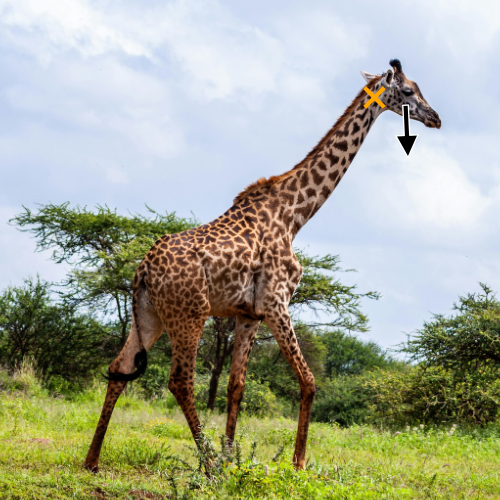} &
        \includegraphics[width=0.15\linewidth]{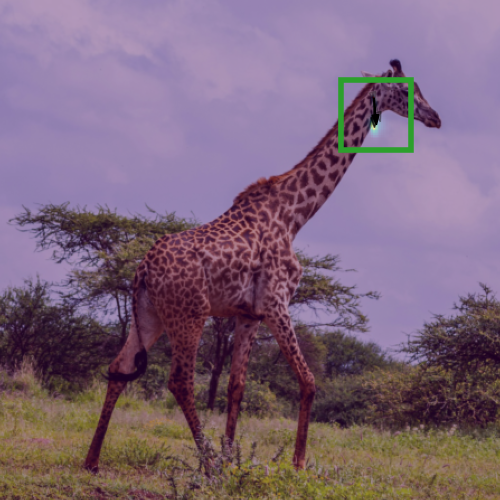} &
        \includegraphics[width=0.15\linewidth]{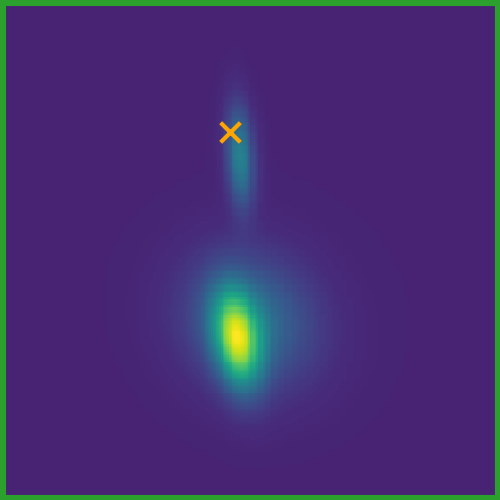} \\
    \end{tabular}
    \caption{When the head of the giraffe is moving down, we get different flow distributions depending on how close the query is to the head. Since the head can also move down without the neck following, we get distributions with more emphasis on no movement when the query is further away from the head (first example). When the query gets really close to the head (second example), the likelihood of movement at the query also increases which can be seen in the stronger bottom mode.}
    \end{subfigure}
    \hfill

    \begin{subfigure}[b]{\linewidth}
    \centering
    \begin{tabular}{cc cc cc}
        \includegraphics[width=0.15\linewidth]{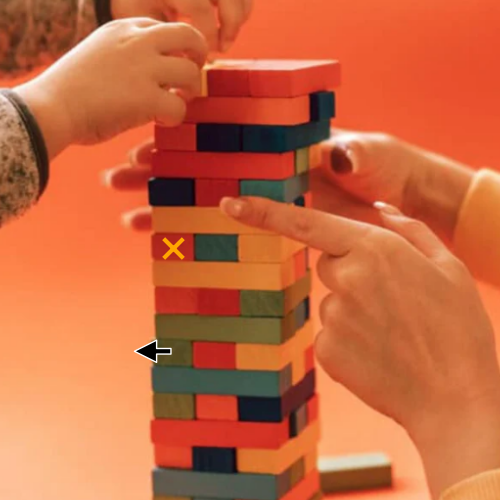} &
        \includegraphics[width=0.15\linewidth]{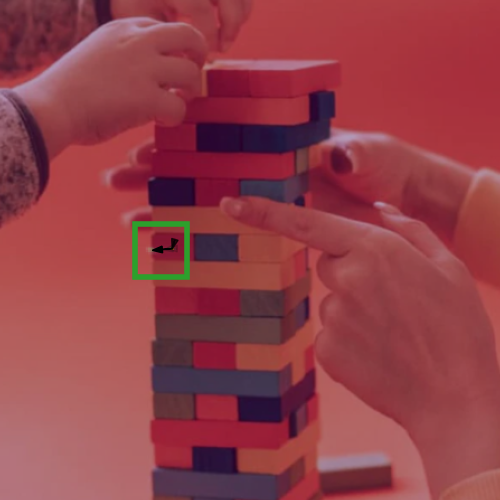} &
        \includegraphics[width=0.15\linewidth]{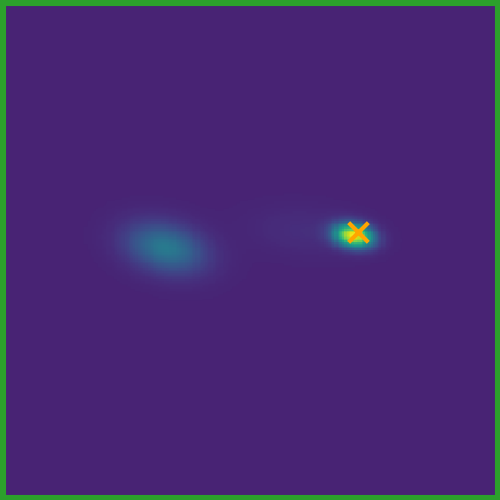} &
        \includegraphics[width=0.15\linewidth]{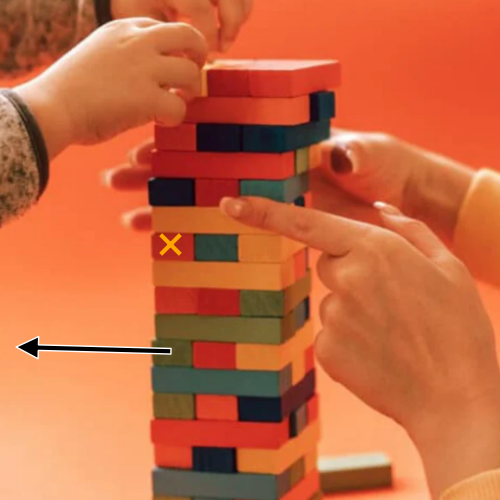} &
        \includegraphics[width=0.15\linewidth]{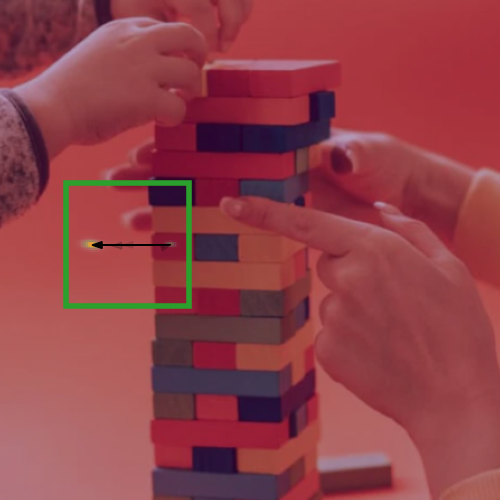} &
        \includegraphics[width=0.15\linewidth]{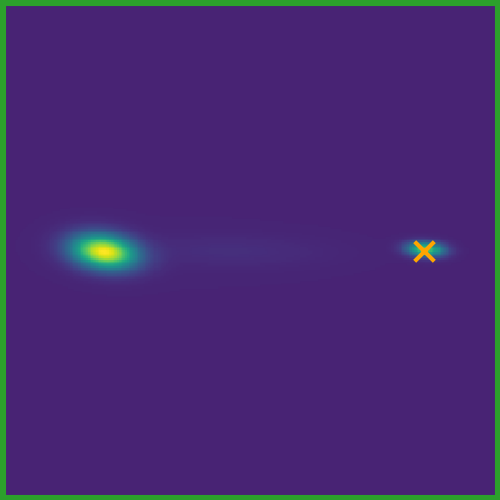} \\
    \end{tabular}
    \caption{The model accounts both for the possibility of the tower falling over with the brick's movement and with it staying stationary. The likelihood of the tower falling over depends on the velocity with which the brick is removed.}
    \end{subfigure}
    \hfill

    \begin{subfigure}[b]{\linewidth}
    \centering
    \begin{tabular}{cc cc cc}
        \includegraphics[width=0.15\linewidth]{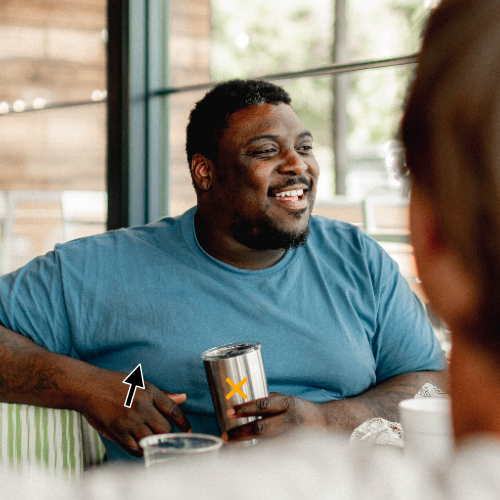} &
        \includegraphics[width=0.15\linewidth]{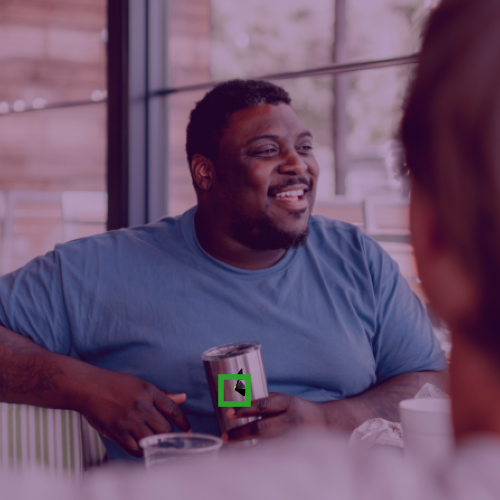} &
        \includegraphics[width=0.15\linewidth]{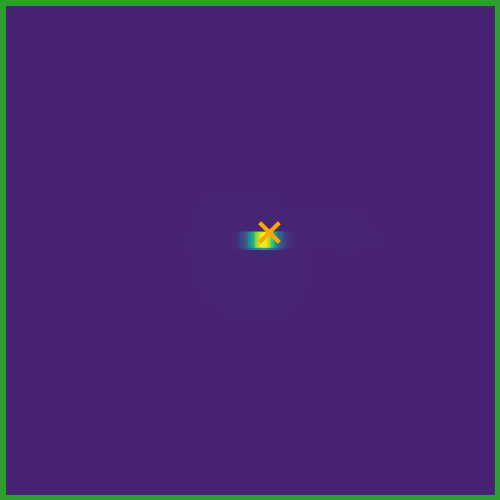} &
        \includegraphics[width=0.15\linewidth]{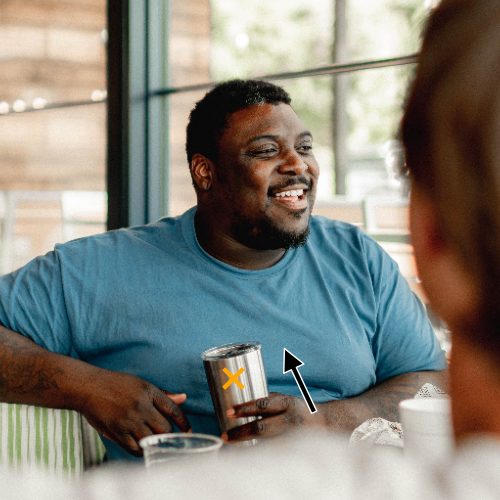} &
        \includegraphics[width=0.15\linewidth]{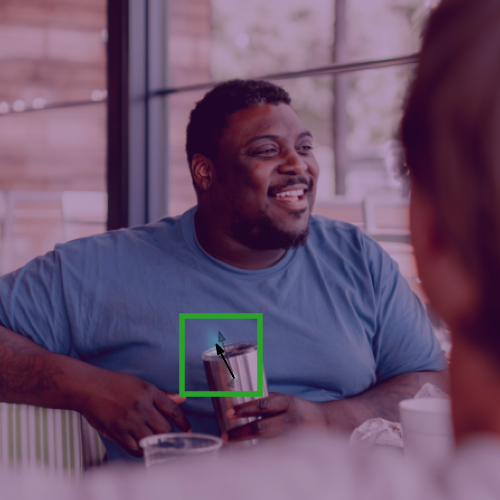} &
        \includegraphics[width=0.15\linewidth]{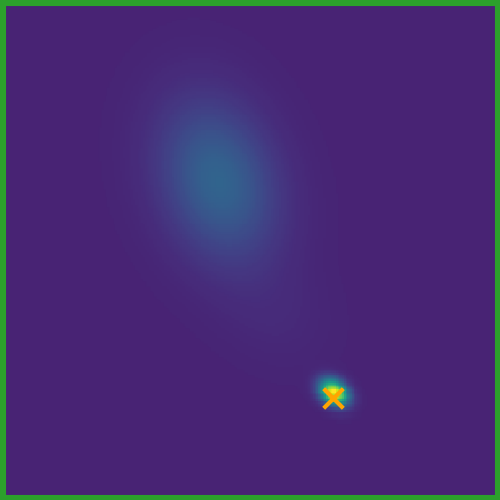} \\
    \end{tabular}
    \caption{Depending on which hand moves, the cup is predicted to be either stationary or \textit{potentially} moving together with the hand holding it. Note that the case of the cup not moving with the hand holding it is very improbable, as visualized by the arrow pointing to that mode having substantially less opacity.}
    \end{subfigure}
    \hfill

    \begin{subfigure}[b]{\linewidth}
    \centering
    \begin{tabular}{cc cc cc}
        \includegraphics[width=0.15\linewidth]{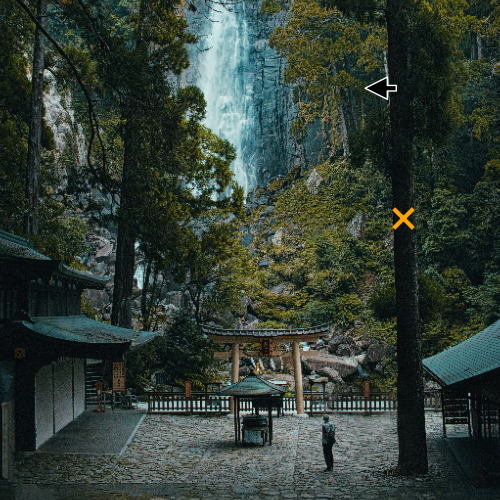} &
        \includegraphics[width=0.15\linewidth]{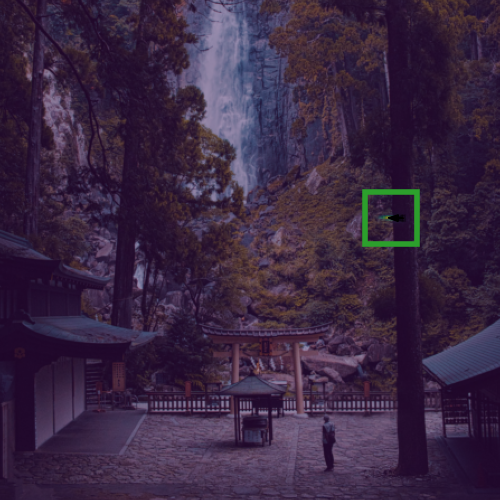} &
        \includegraphics[width=0.15\linewidth]{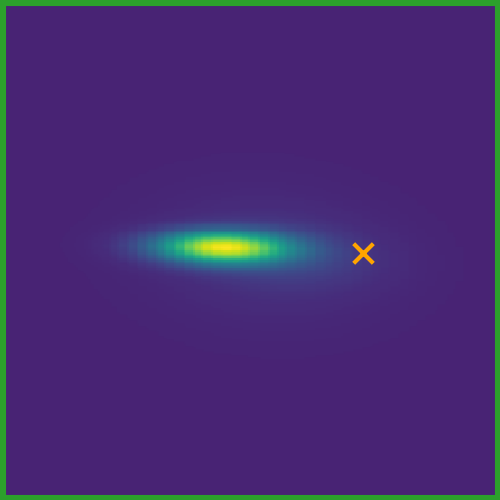} &
        \includegraphics[width=0.15\linewidth]{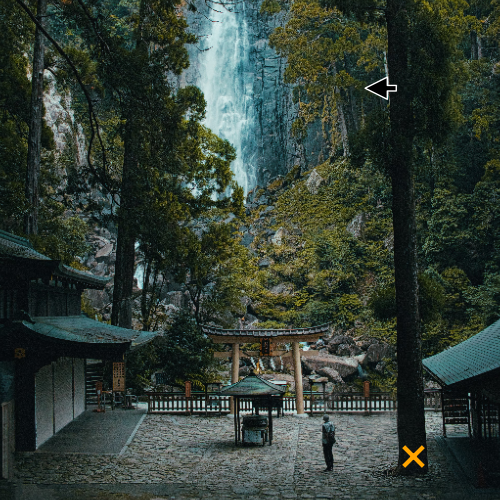} &
        \includegraphics[width=0.15\linewidth]{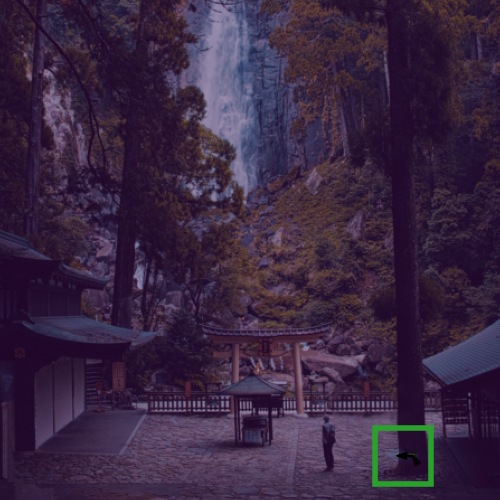} &
        \includegraphics[width=0.15\linewidth]{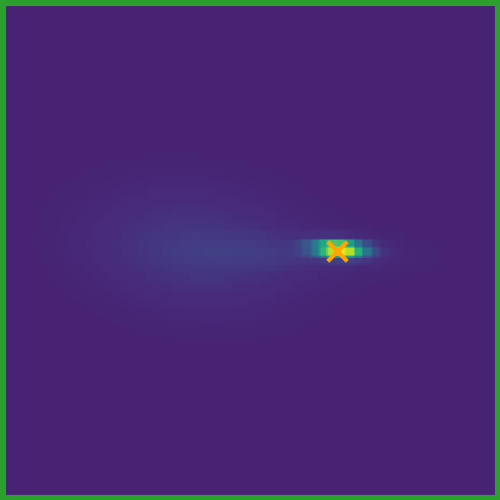} \\
    \end{tabular}
    \caption{Depending on the height of the position queried on the tree, the magnitude of the predicted movement changes, reflecting typical intuition as to how a tree moves.}
    \end{subfigure}
    \hfill

    \begin{subfigure}[b]{\linewidth}
    \centering
    \begin{tabular}{cc cc cc}
        \includegraphics[width=0.15\linewidth]{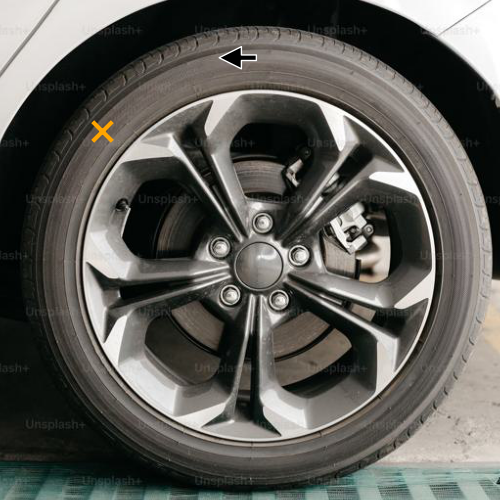} &
        \includegraphics[width=0.15\linewidth]{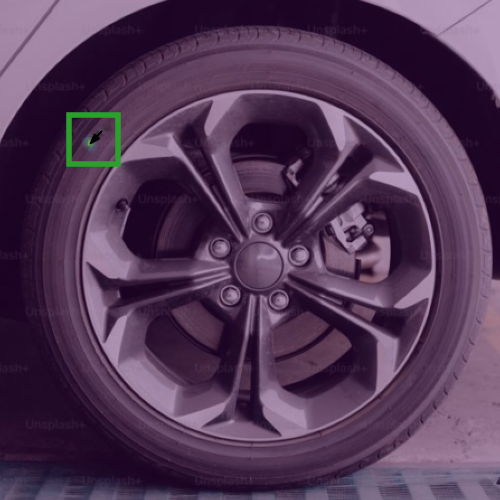} &
        \includegraphics[width=0.15\linewidth]{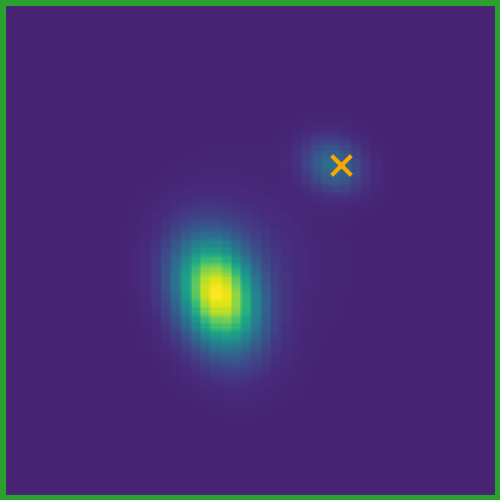} &
        \includegraphics[width=0.15\linewidth]{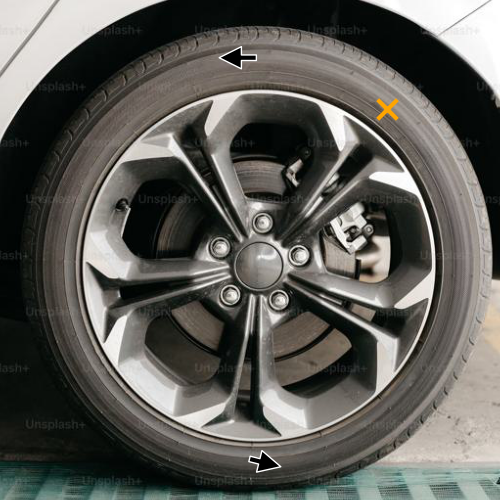} &
        \includegraphics[width=0.15\linewidth]{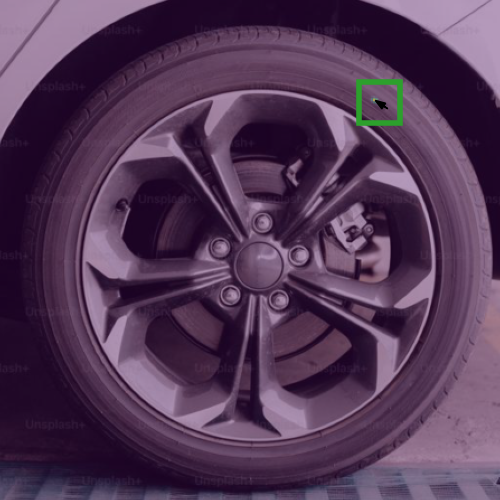} &
        \includegraphics[width=0.15\linewidth]{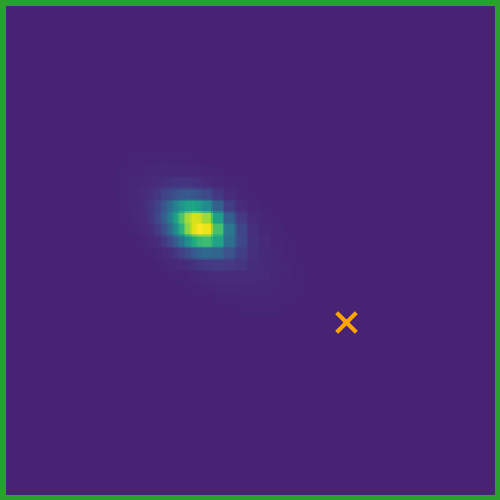} \\
    \end{tabular}
    \caption{The model is capable of understanding the effect of rotational movements.}
    \end{subfigure}
    \hfill
    
    \caption{Visualization of flow distribution for different pokes on the same image. The overlayed distribution visualizes the potential movement in the overall images, with the opacity of arrows denoting how likely each mode is (the more likely a mode, the less transparent the arrow).}
    \label{fig:app_distribution}
\end{figure*}
}

{
\setlength{\tabcolsep}{0.007\linewidth}
\begin{figure*}[htb]
    \centering
    \adjustbox{max width=\linewidth}{
    \begin{tabular}{cc cc cc}
        {\footnotesize Input Image} & {\footnotesize Flow Prediction} &{\footnotesize Input Image} &{\footnotesize Flow Prediction} &{\footnotesize Input Image} &{\footnotesize Flow Prediction} \\
        \includegraphics[width=0.15\linewidth]{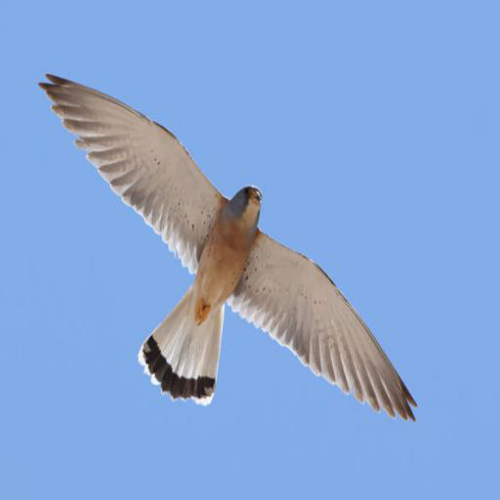} &
        \includegraphics[width=0.15\linewidth]{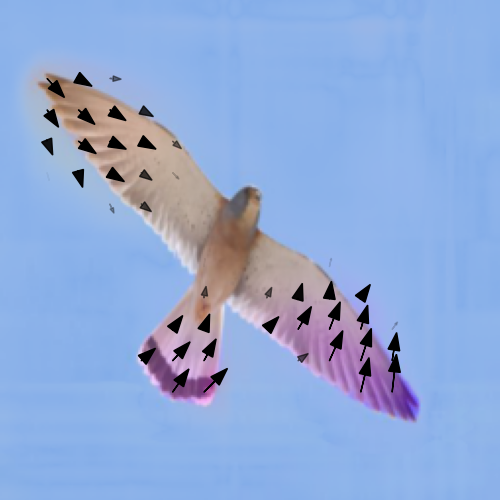} &
        \includegraphics[width=0.15\linewidth]{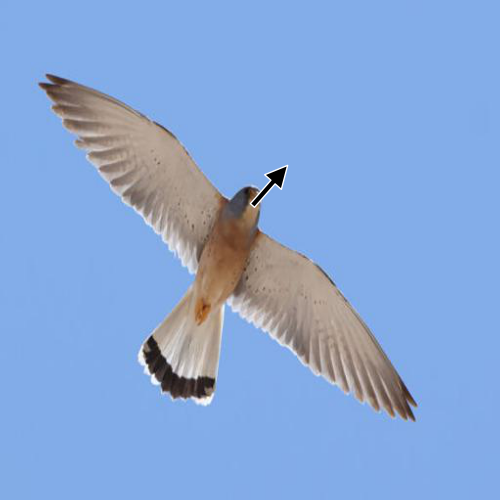} &
        \includegraphics[width=0.15\linewidth]{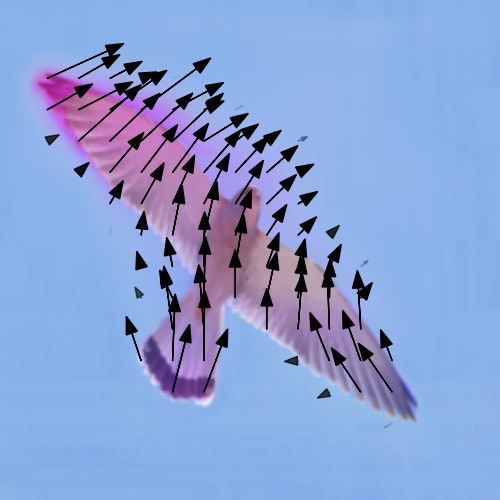} &
        \includegraphics[width=0.15\linewidth]{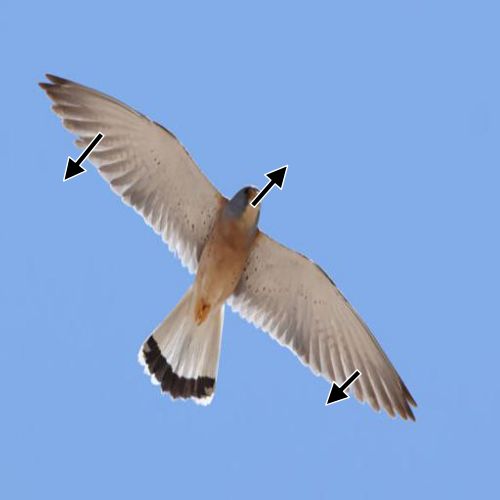} &
        \includegraphics[width=0.15\linewidth]{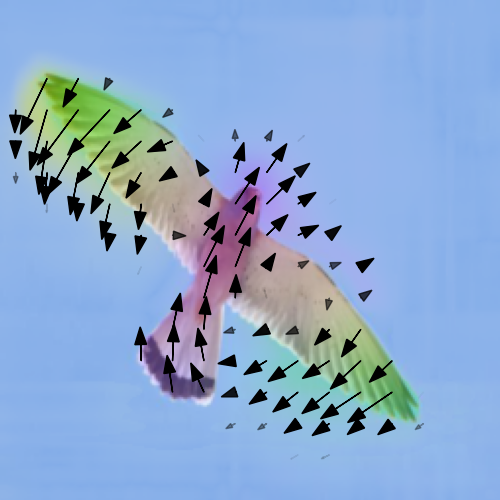} \\

        \includegraphics[width=0.15\linewidth]{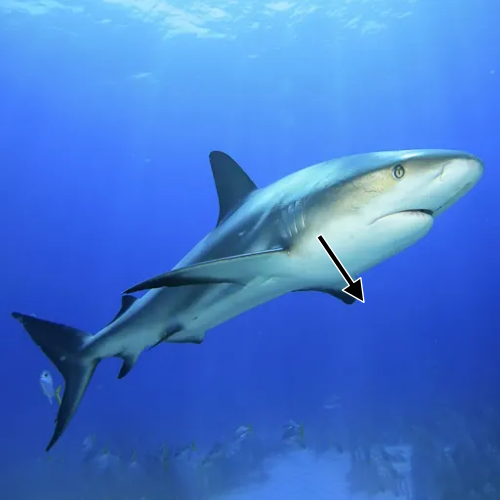} &
        \includegraphics[width=0.15\linewidth]{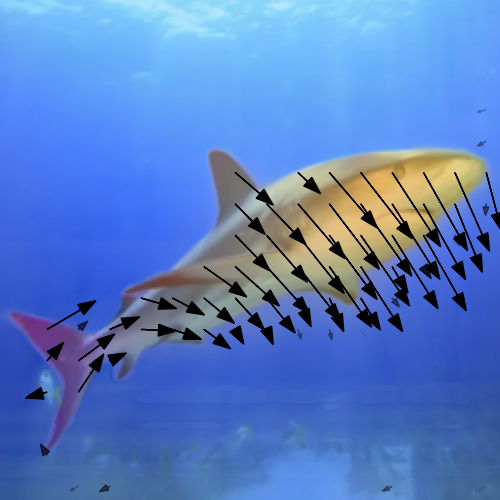} &
        \includegraphics[width=0.15\linewidth]{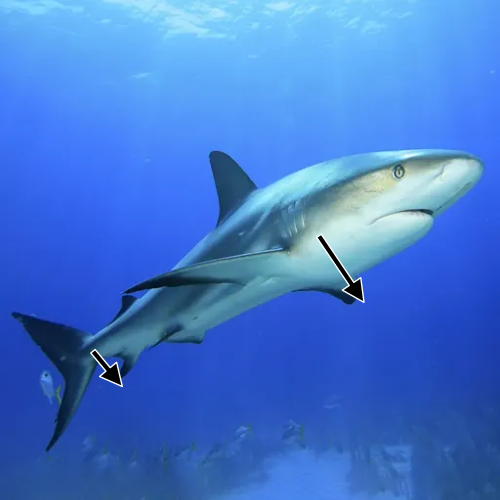} &
        \includegraphics[width=0.15\linewidth]{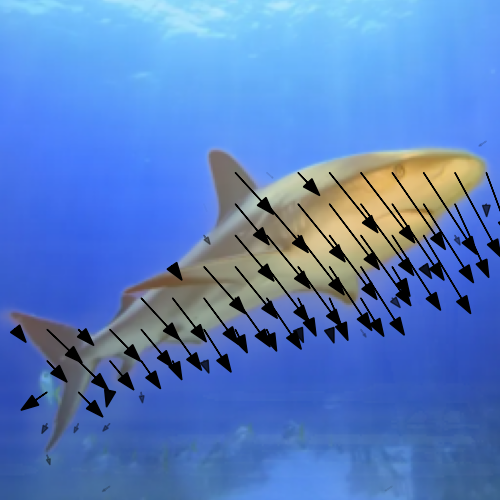} &
        \includegraphics[width=0.15\linewidth]{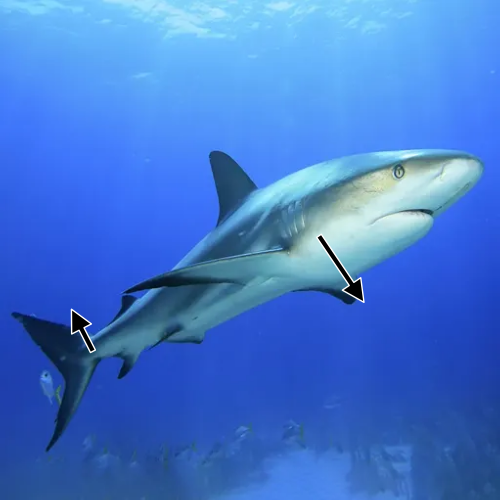} &
        \includegraphics[width=0.15\linewidth]{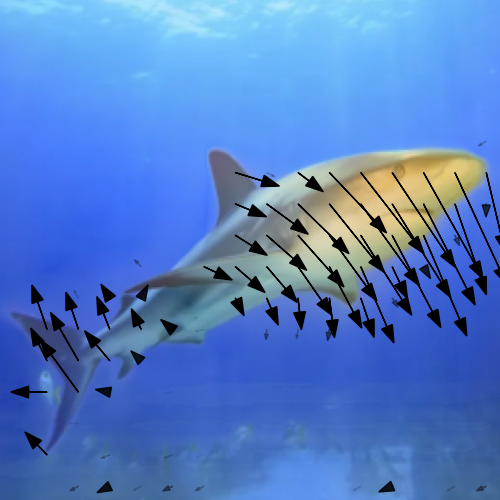} \\

        \includegraphics[width=0.15\linewidth]{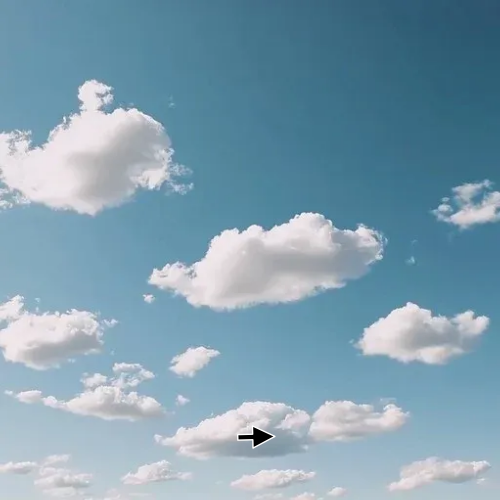} &
        \includegraphics[width=0.15\linewidth]{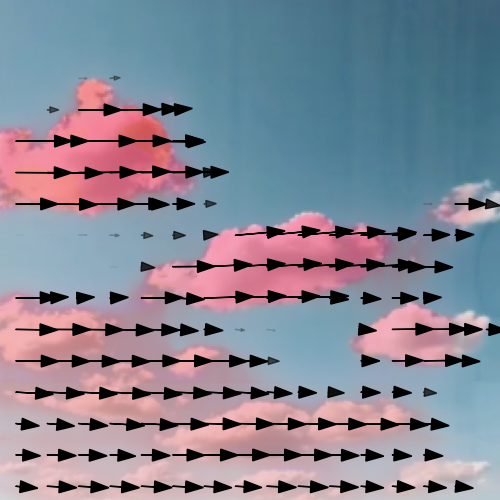} &
        \includegraphics[width=0.15\linewidth]{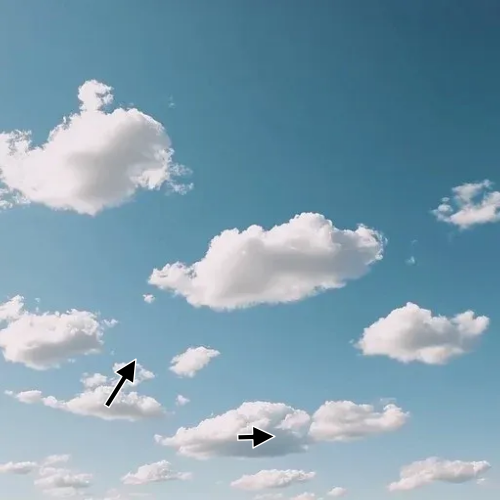} &
        \includegraphics[width=0.15\linewidth]{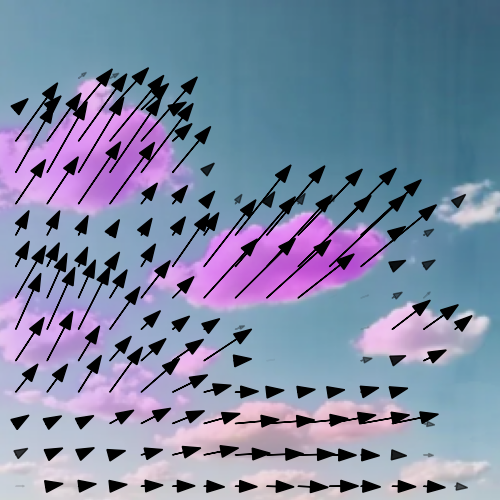} &
        \includegraphics[width=0.15\linewidth]{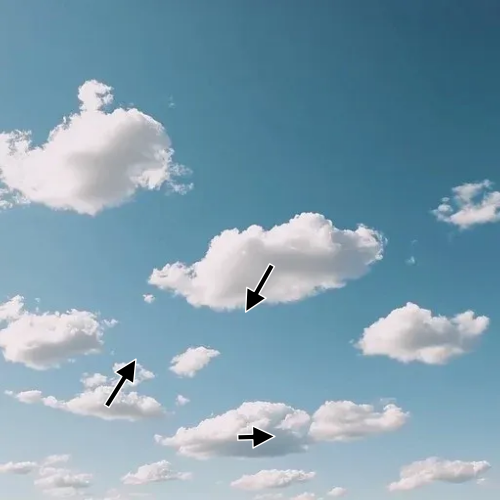} &
        \includegraphics[width=0.15\linewidth]{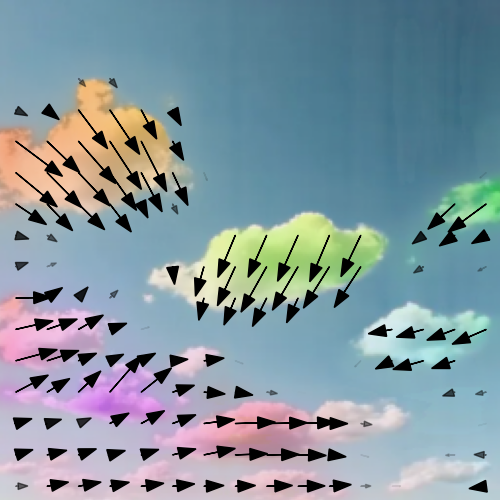} \\

        \includegraphics[width=0.15\linewidth]{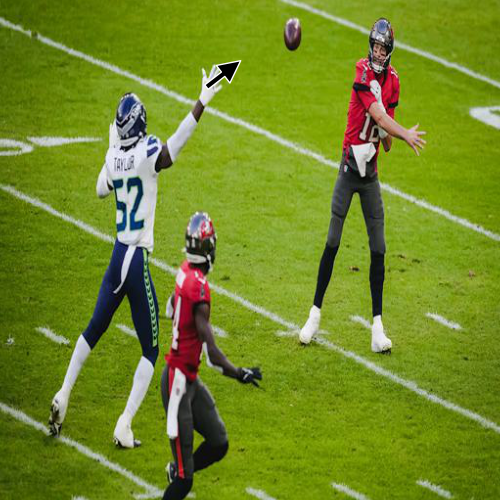} &
        \includegraphics[width=0.15\linewidth]{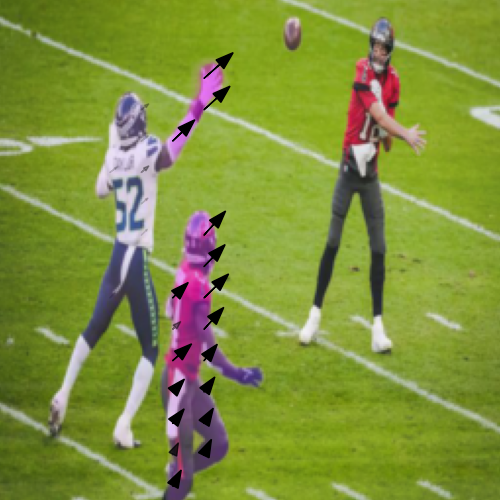} &
        \includegraphics[width=0.15\linewidth]{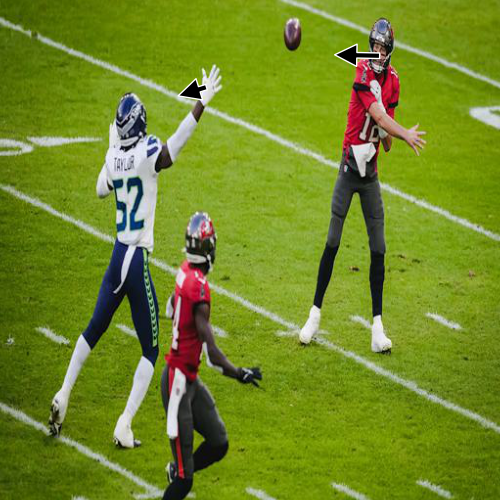} &
        \includegraphics[width=0.15\linewidth]{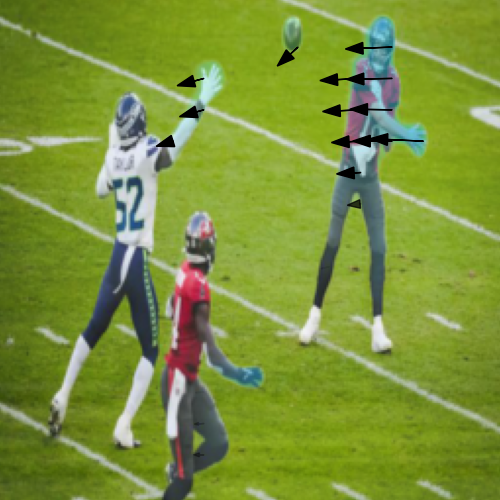} &
        \includegraphics[width=0.15\linewidth]{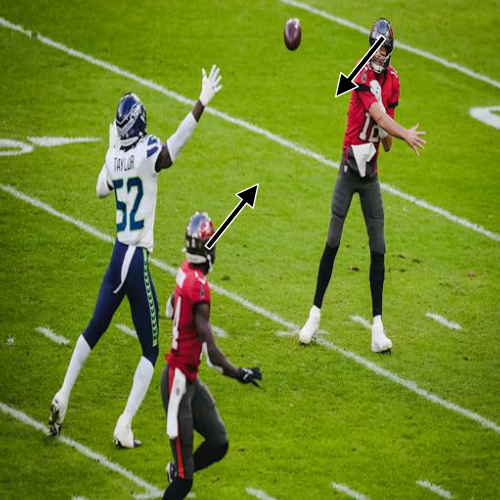} &
        \includegraphics[width=0.15\linewidth]{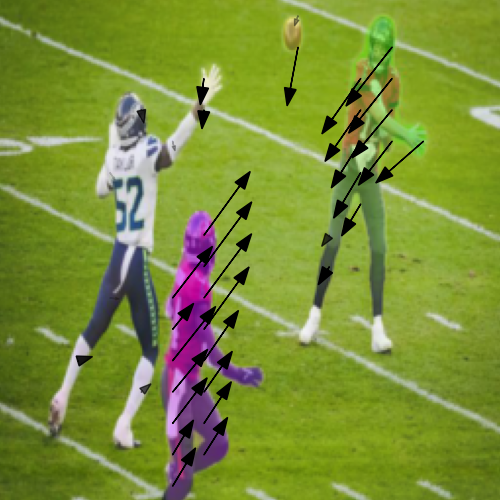} \\

        \includegraphics[width=0.15\linewidth]{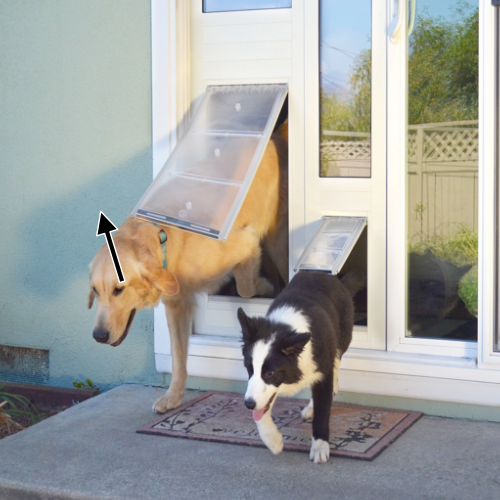} &
        \includegraphics[width=0.15\linewidth]{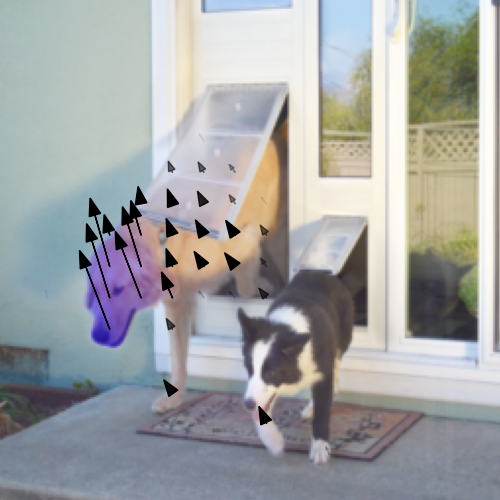} &
        \includegraphics[width=0.15\linewidth]{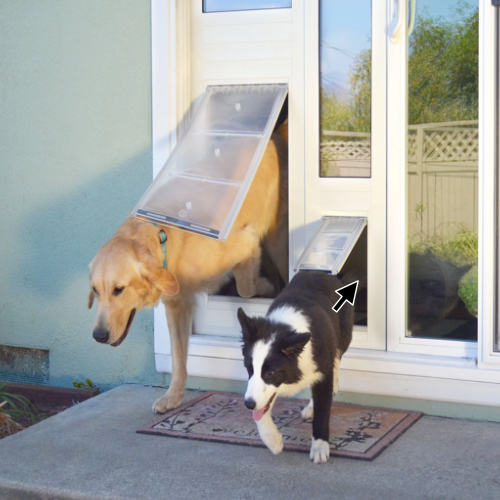} &
        \includegraphics[width=0.15\linewidth]{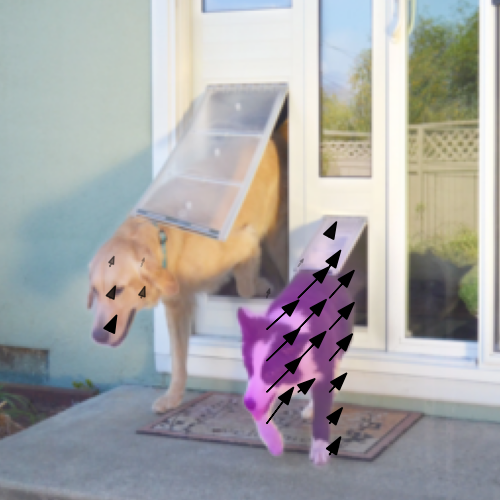} &
        \includegraphics[width=0.15\linewidth]{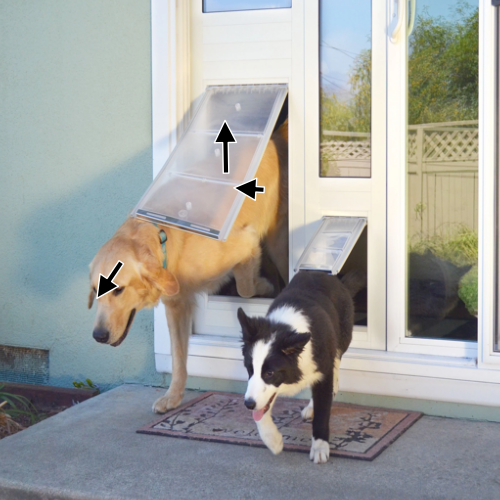} &
        \includegraphics[width=0.15\linewidth]{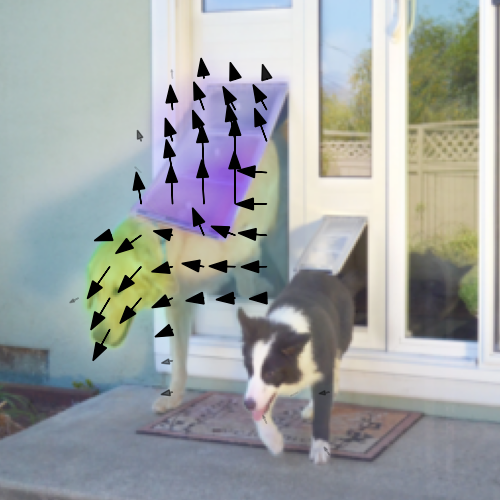} \\

        \includegraphics[width=0.15 \linewidth]{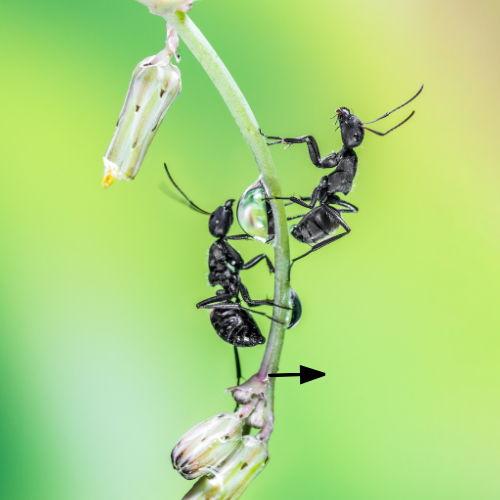} & 
        \includegraphics[width=0.15 \linewidth]{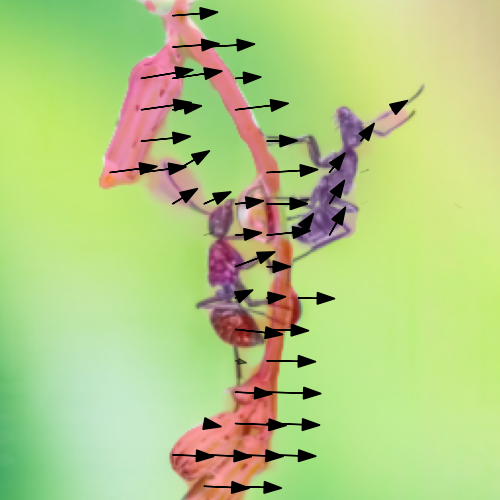} &
        \includegraphics[width=0.15 \linewidth]{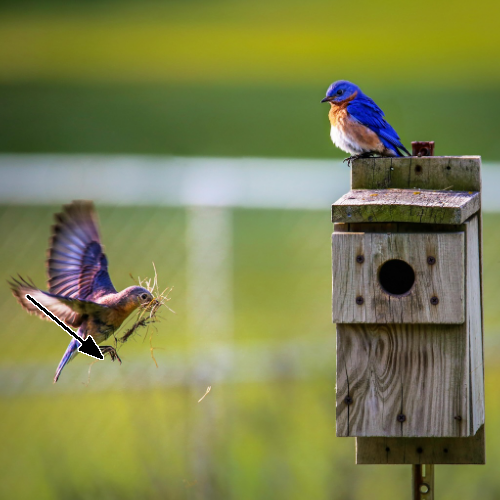} & 
        \includegraphics[width=0.15 \linewidth]{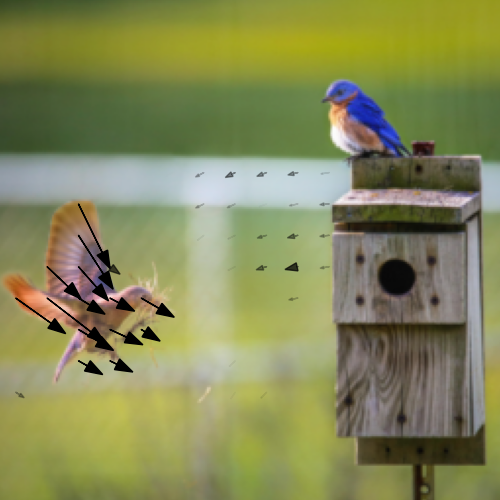} &
        \includegraphics[width=0.15 \linewidth]{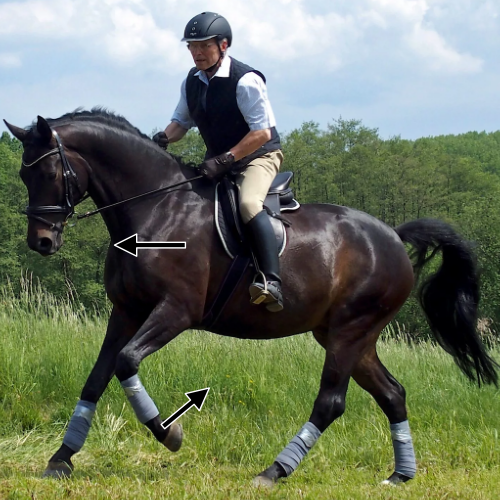} & 
        \includegraphics[width=0.15 \linewidth]{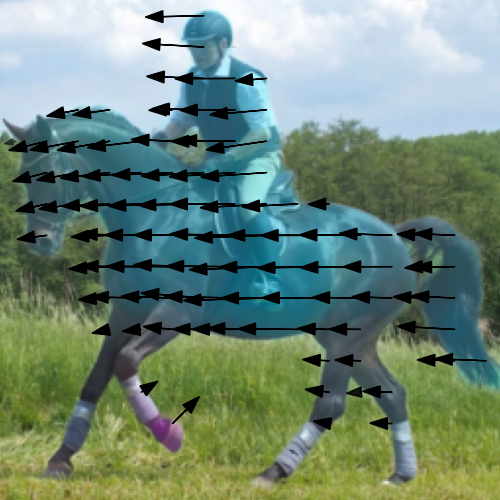} \\

        \includegraphics[width=0.15 \linewidth]{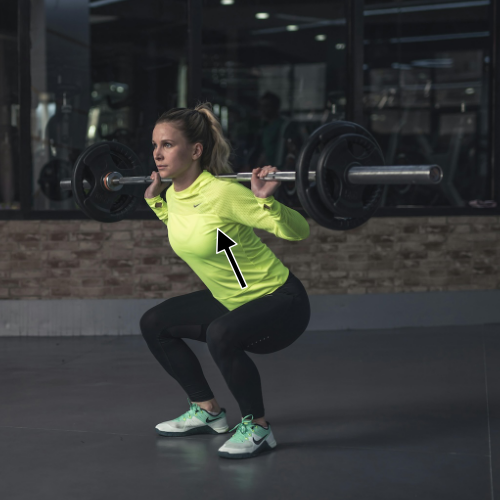} & 
        \includegraphics[width=0.15 \linewidth]{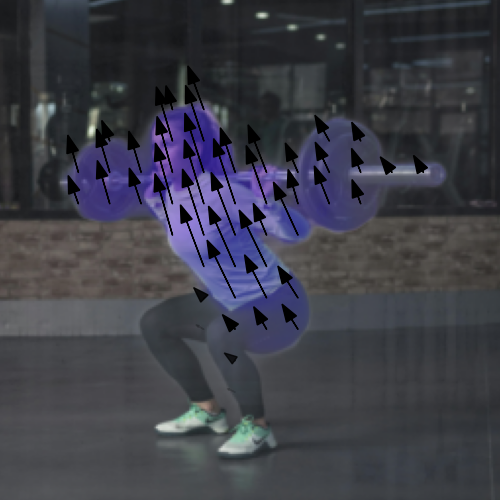} &
        \includegraphics[width=0.15 \linewidth]{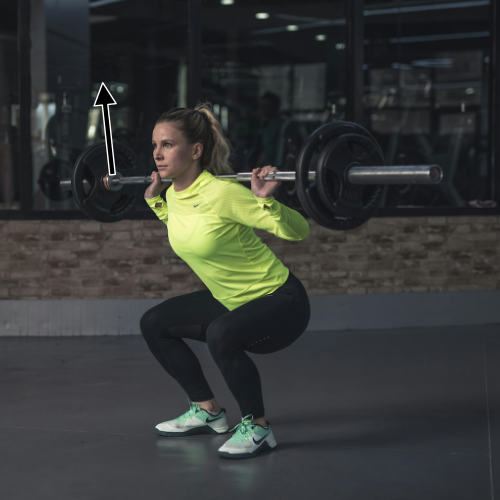} & 
        \includegraphics[width=0.15 \linewidth]{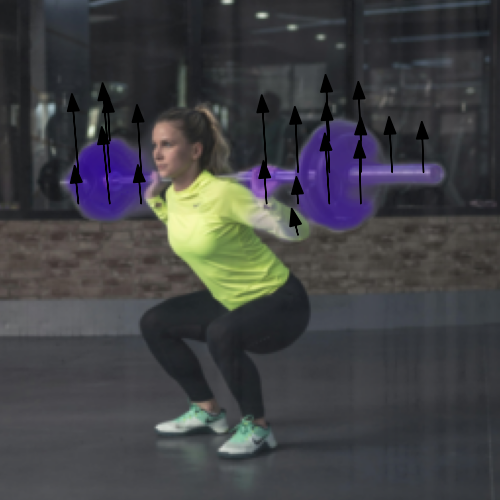} &
        \includegraphics[width=0.15 \linewidth]{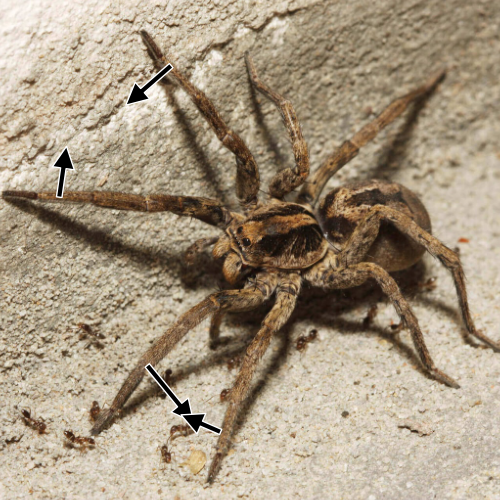} & 
        \includegraphics[width=0.15 \linewidth]{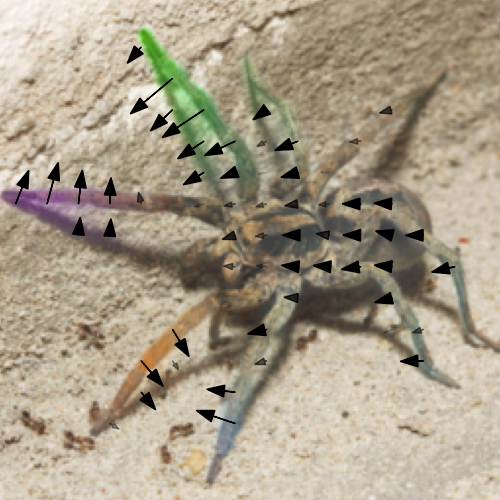} \\
        
    \end{tabular}
    }
    
    \caption{Qualitative samples visualizing motion predictions inferred from a single image and (optionally) pokes.}
    \label{fig:app_dense}
\end{figure*}
}

\end{document}